\documentclass[10pt,twocolumn,letterpaper]{article}
\pdfoutput=1

\usepackage[pagenumbers]{cvpr} %

\usepackage[table]{xcolor}
\usepackage{color}
\usepackage{soul}
\usepackage[export]{adjustbox}
\usepackage{colortbl}
\usepackage{tabularray}
\usepackage{graphicx}
\usepackage{amsmath}
\usepackage{amssymb}
\usepackage{enumitem}
\usepackage{booktabs}
\usepackage{array}
\usepackage[longtable]{multirow}
\usepackage{longtable}
\usepackage{booktabs}

\usepackage{todonotes}
\usepackage{lipsum}                     %
\usepackage{xargs}                      %
\usepackage{svg}

\usepackage{alltt}
\usepackage[most]{tcolorbox}
\tcbset{
  aibox/.style={
    width=474.18663pt,
    top=10pt,
    colback=white,
    colframe=black,
    colbacktitle=black,
    enhanced,
    center,
    attach boxed title to top left={yshift=-0.1in,xshift=0.15in},
    boxed title style={boxrule=0pt,colframe=white,},
  }
}
\newtcolorbox{AIbox}[2][]{aibox,title=#2,#1}

\newcommand{\fld}{\emph{FLD-5B}\xspace}
\newcommand{\davitb}{DaViT-B\xspace}
\newcommand{\flmodel}{\emph{Florence-2}\xspace}
\newcommand{\flmodelb}{\emph{Florence-2-B}\xspace}
\newcommand{\flmodell}{\emph{Florence-2-L}\xspace}

\renewcommand{\paragraph}[1]{\vspace{1.25mm}\noindent\textbf{#1}}

\definecolor{lightgray}{gray}{0.9}
\sethlcolor{lightgray}

\definecolor{cvprblue}{rgb}{0.21,0.49,0.74}
\usepackage[pagebackref,breaklinks,colorlinks,citecolor=cvprblue]{hyperref}
\usepackage[british,UKenglish,USenglish,english,american]{babel}

\newcommand{\hide}[1]{}

\usepackage[capitalize]{cleveref}
\crefname{section}{Section}{Sections.}
\Crefname{section}{Section}{Sections}
\Crefname{table}{Table}{Tables}
\crefname{table}{Table}{Tables}
\crefname{figure}{Figure}{Figures.}
\Crefname{figure}{Figure}{Figures.}

\renewcommand{\comment}[1]{}

\def\cvprPaperID{*****} %
\def\confName{CVPR}
\def\confYear{2024}

\begin{document}

\title{Florence-2: Advancing a Unified Representation for a Variety of Vision Tasks}
\author{Bin Xiao$^{\dagger}$ \quad Haiping Wu$^{*}$ \quad Weijian Xu$^{*}$ \quad Xiyang Dai \quad Houdong Hu \\ \quad Yumao Lu \quad Michael Zeng \quad Ce Liu$^{\ddagger}$ \quad Lu Yuan$^{\ddagger}$  \vspace{0.5em} \\
\small $^{\dagger}$project lead \qquad $^{*}$equal contribution \qquad $^{\ddagger}$direcional lead \\[3mm]
Azure AI, Microsoft
}

\maketitle

\begin{abstract}

We introduce \flmodel, a novel vision foundation model with a unified, prompt-based representation for a variety of computer vision and vision-language tasks. While existing large vision models excel in transfer learning, they struggle to perform a diversity of tasks with simple instructions, a capability that implies handling the complexity of various spatial hierarchy and semantic granularity. \flmodel was designed to take text-prompt as task instructions and generate desirable results in text forms, whether it be captioning, object detection, grounding or segmentation. This multi-task learning setup demands large-scale, high-quality annotated data. To this end, we co-developed \fld that consists of 5.4 billion comprehensive visual annotations on 126 million images, using an iterative strategy of automated image annotation and model refinement. We adopted a sequence-to-sequence structure to train \flmodel to perform versatile and comprehensive vision tasks. Extensive evaluations on numerous tasks demonstrated \flmodel to be a strong vision foundation model contender with unprecedented zero-shot and fine-tuning capabilities.

\end{abstract}

\section{Introduction}

\begin{figure}[t]
  \centering
   \includegraphics[width=1.0\linewidth]{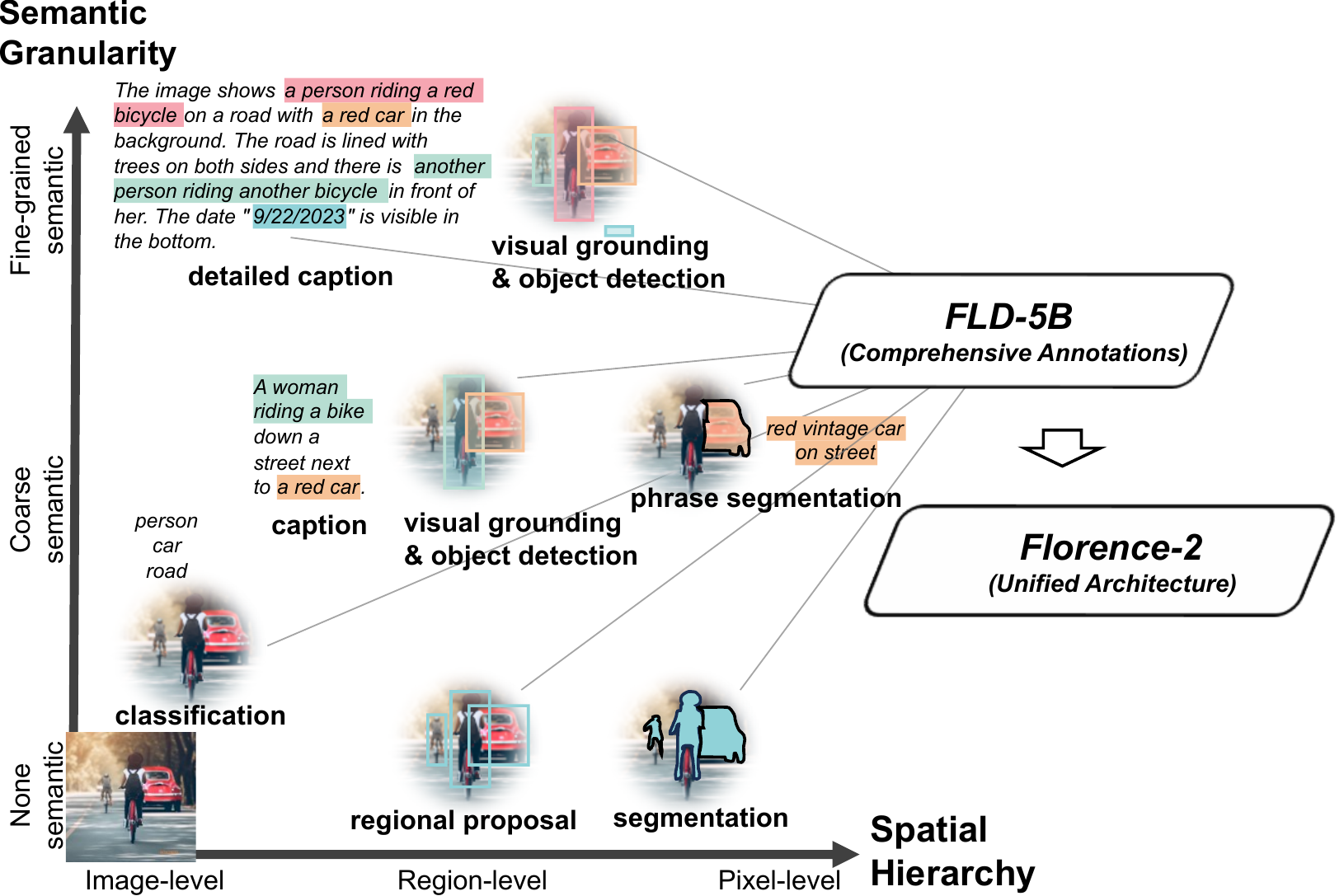}

   \caption{We aim to build a vision foundation model to enable extensive perception capabilities including spatial hierarchy and semantic granularity. To achieve this, a single unified model \textbf{\flmodel} is pre-trained on our \textbf{\fld} dataset encompassing a total of 5.4B comprehensive annotations across 126M images, which are collected by our Florence data engine. }
   \label{fig:overall}
\end{figure}

In the realm of Artificial General Intelligence (AGI) systems, there has been a notable shift towards utilizing pre-trained, versatile representations, acknowledged for task-agnostic benefits accross diverse applications. This trend is evident in natural language processing (NLP), where advanced models~\cite{bommasani2021opportunities,devlin2019bert,lewis2019bart,raffel2020exploring, NEURIPS2020_1457c0d6, radford2019language} show adaptability with comprehensive knowledge spanning various domains and tasks with simple instructions. The success of NLP motivates a parallel approach in computer vision.

Universal representation for diverse vision-related tasks presents unique challenges, notably the need for comprehensive perceptual abilities. Unlike NLP, which deals mainly with text, computer vision requires handling intricate visual data like object location, masked contours, and attributes. Attaining universal representation in computer vision demands adept management of a spectrum of complex tasks, organized two-dimensionally as illustrated in~\cref{fig:overall}: \vspace{-0.3em} 
\begin{itemize}
    \item  \textbf{Spatial Hierarchy}: The model must discern spatial details across varying scales, understanding image-level concepts and fine-grained pixel specifics. Accommodating the intricate spatial hierarchy within vision demands the model's proficiency in handling diverse levels of granularity. \vspace{-0.3em} 
    \item \textbf{Semantic Granularity}: Universal representation in computer vision should span a spectrum of semantic granularity. The model transitions from high-level captions to nuanced descriptions, enabling versatile understanding for diverse applications. \vspace{-0.3em} 
\end{itemize}

This pursuit is characterized by distinctiveness and substantial challenges. A key hurdle is the scarcity of \emph{comprehensive visual annotations}, hindering the development of a foundational model capable of capturing the intricate nuances of spatial hierarchy and semantic granularity. Existing datasets, such as ImageNet~\cite{deng2009imagenet}, COCO~\cite{lin2014microsoft}, and Flickr30k Entities~\cite{plummer2015flickr30k}, tailored for specialized applications, are extensively labeled by humans. To overcome this constraint, it is imperative to generate extensive annotations for each image on a larger scale.

Another challenge is the absence of a \emph{unified pre-training framework with a singular network architecture} that seamlessly integrates spatial hierarchy and semantic granularity in computer vision. Traditional models excel in tasks like object detection~\cite{he2017mask,zhang2022dino}, semantic segmentation~\cite{cheng2021mask2former,xiao2018unified}, and image captioning~\cite{li2022blip,wang2022git} with task-specific design. However, it is essential to develop a comprehensive, unified model that is capable of adapting across various vision tasks in a task-agnostic manner, even accommodating new tasks with minimal or no task-specific fine-tuning.

The model \emph{Florence}~\cite{yuan2021florence} pioneers the integration of spatial, temporal, and multi-modal aspects in computer vision through unified pre-training and network architecture. The first evolutionary version~\cite{yuan2021florence} excels in transfer learning via pre-training with noisy text-image pairs and task-specific fine-tuning using specialized adapters. However, it relies on large task-specific datasets and adapters, leaving gaps in addressing the above dual key challenges.

In this paper, we introduce \flmodel, a universal backbone achieved through multitask learning with extensive visual annotations. This results in a unified, prompt-based representation for diverse vision tasks, effectively addressing the challenges of limited comprehensive data and the absence of a unified architecture.

Multitask learning necessitates large-scale, high-quality annotated data. Our data engine, instead of relying on labor-intensive manual annotation, autonomously generates a comprehensive visual dataset called \fld, encompassing a total of 5.4B annotations for 126M images. This engine consists of two efficient processing modules. The first module uses specialized models to collaboratively and autonomously annotate images, moving away from the traditional single and manual annotation approach. Multiple models work together to reach a consensus, reminiscent of the wisdom of crowds concept~\cite{yi2012wisdom, weststrate2019wisdom, kittur2007power}, ensuring a more reliable and unbiased image understanding. The second module iteratively refines and filters these automated annotations using well-trained foundational models.

By utilizing this extensive dataset, our model employs a sequence-to-sequence (seq2seq) architecture~\cite{sutskever2014sequence, cho2014learning, raffel2020exploring, devlin2019bert}, which integrates an image encoder and a multi-modality encoder-decoder. This design accommodates a spectrum of vision tasks without the need for task-specific architectural modifications, aligning with the ethos of the NLP community for versatile model development with a consistent underlying structure. All annotations in the dataset \fld, are uniformly standardized into textual outputs, 
facilitating a unified multi-task learning approach with consistent optimization with the same loss function as the objective. The outcome is a versatile vision foundation model, \flmodel, capable of performing a variety of tasks, such as object detection, captioning, and grounding, all within a single model governed by a uniform set of parameters. Task activation is achieved through textual prompts, reflecting the approach used by Large Language Models (LLMs)~\cite{radford2019language}.

Our approach attains a universal representation, demonstrating broad applicability across various visual tasks. Key results include: \vspace{-0.3em} 

\begin{itemize}
\item As a versatile vision foundation model, \flmodel achieves new state-of-the-art zero-shot performance in tasks such as captioning on COCO~\cite{lin2014microsoft}, visual grounding on Flick30k~\cite{plummer2015flickr30k}, and referring expression comprehension on RefCOCO/+/g~\cite{kazemzadeh2014referitgame, yu2016modeling, mao2016generation}. \vspace{-0.3em} 

\item After fine-tuning with public human-annotated data, \flmodel, despite its compact size, competes with larger specialist models. Notably, the fine-tuned \flmodel establishes new state-of-the-art results on the benchmarks on RefCOCO/+/g. \vspace{-0.3em} 

\item The pre-trained \flmodel backbone enhances performance on downstream tasks, \eg COCO object detection and instance segmentation, and ADE20K semantic segmentation, surpassing both supervised and self-supervised models. Compared to pre-trained models on ImageNet, ours improves training efficiency by 4$\times$ and achieves substantial improvements of 6.9, 5.5, and 5.9 points on COCO~\cite{lin2014microsoft} and ADE20K~\cite{zhou2017scene} datasets, using Mask-RCNN~\cite{he2017mask}, DINO~\cite{zhang2022dino}, and UperNet~\cite{xiao2018unified} frameworks respectively.  \vspace{-0.3em} 
\end{itemize}

\section{Rethinking Vision Model Pre-training}
\label{sec:task}

\begin{figure*}[ht]
    \centering
    \includegraphics[width=1.0\textwidth]{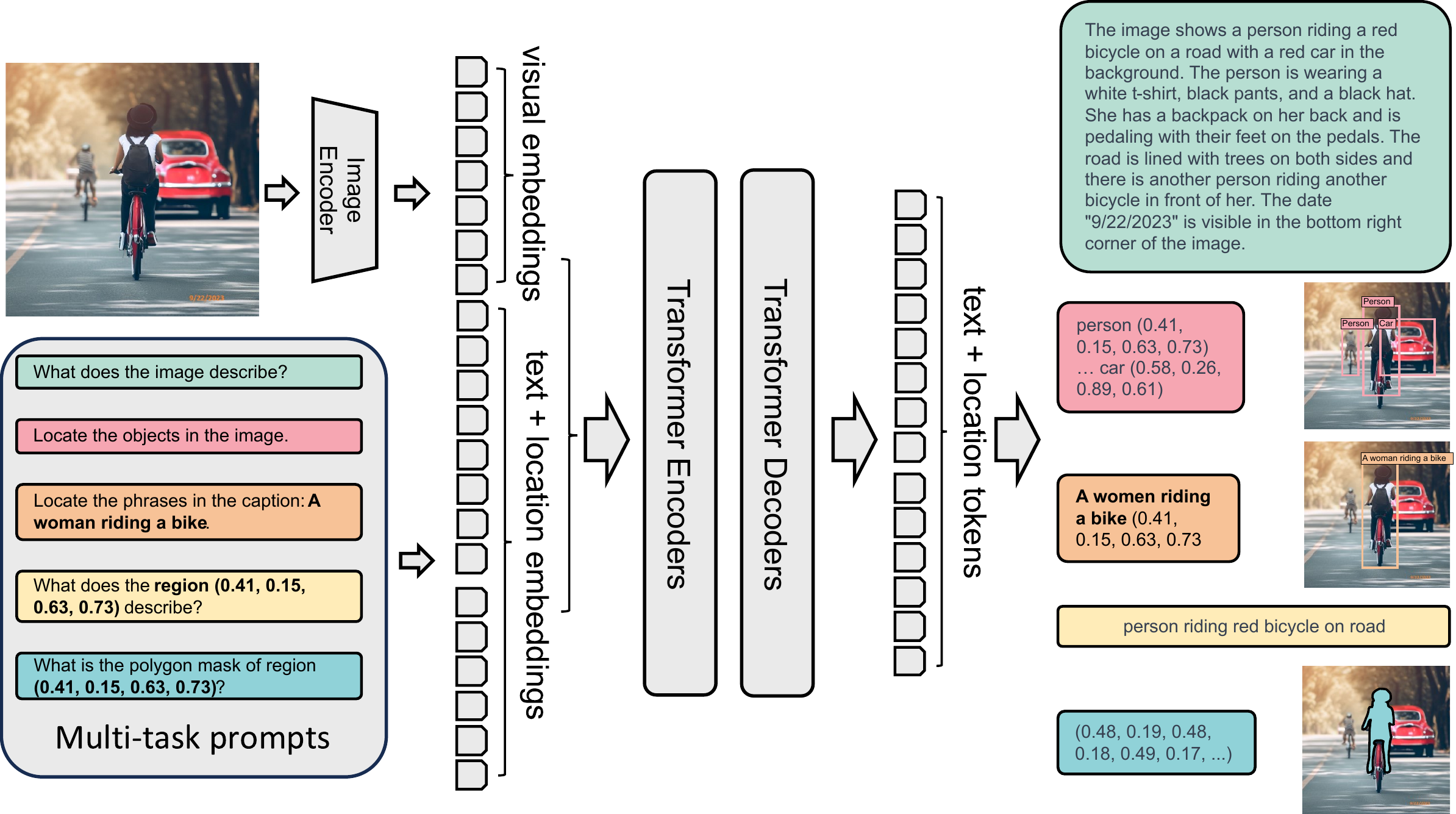}
    \caption{\textbf{\flmodel} consists of an image encoder and standard multi-modality encoder-decoder. We train \textbf{\flmodel} on our \textbf{\fld} data in a unified multitask learning paradigm, resulting in a generaslist vision foundation model, which can perform various vision tasks.}
    \label{fig:model_pipeline}
\end{figure*}

In pursuit of a versatile vision foundation model, we revisit three predominant pre-training paradigms: supervised (\eg, ImageNet classification \cite{deng2009imagenet}), self-supervised (\eg, SimCLR \cite{chen2020simple}, MoCo \cite{he2020momentum}, BEiT \cite{biao2021beit}, MAE~\cite{he2022masked}), and weakly supervised (\eg, CLIP~\cite{radford2021learning}, Florence~\cite{yuan2021florence}, SAM~\cite{kirillov2023segment}). Each paradigm captures unique aspects of visual data but is inherently limited by the constraints of single-task learning frameworks. Supervised pre-training excels in object recognition but lacks adaptability~\cite{krizhevsky2012imagenet}; self-supervised algorithms reveal intricate features but may overemphasize certain attributes~\cite{caron2020unsupervised}; weakly supervised methods leverage unstructured textual annotations but yield only image-level understanding~\cite{radford2021learning}. To build a unified vision foundation model suitable for various applications, we must explore innovative pre-training strategies that overcome single-task limitations and integrate both textual and visual semantics.

Image understanding necessitates capturing multiple levels of granularity, from global semantics to local details, and comprehending spatial relationships between objects and entities in their semantic context. To address these core aspects of image understanding, our approach incorporates a diverse set of annotations, effectively capturing visual understanding nuances and bridging the gap between vision and language understanding.

\subsection{Comprehensive Multitask Learning}

To develop a versatile vision foundation model, we formulate a range of multitask learning objectives, each tailored to address specific aspects of visual comprehension. These objectives align with our predefined criteria: spatial hierarchy and semantic granularity, inspired by recent research on multitask learning~\cite{alayrac2022flamingo, wang2022ofa, lu2022unifiedio, chen2022pali, chen2023pali, chen2023pali3}. Our multitask learning approach incorporates three distinct learning objectives, each addressing a different level of granularity and semantic understanding: \vspace{-0.3em}

\begin{itemize}
\item \textbf{Image-level understanding} tasks capture high-level semantics and foster a comprehensive understanding of images through linguistic descriptions~\cite{chen2015microsoft, young2014image, deng2009imagenet, krause2017hierarchical}.
They enable the model to comprehend the overall context of an image and grasp semantic relationships and contextual nuances in the language domain. Exemplar tasks include image classification, captioning, and visual question answering.\vspace{-0.3em}

\item \textbf{Region/pixel-level recognition} tasks facilitate detailed object and entity localization within images, capturing relationships between objects and their spatial context. Tasks include object detection, segmentation, and referring expression comprehension.\vspace{-0.3em}

\item \textbf{Fine-grained visual-semantic alignment} tasks require fine-grained understanding of both text and image. It involves locating the image regions that correspond to the text phrases, such as objects, attributes, or relations. These tasks challenge the ability to capture the local details of visual entities and their semantic contexts, as well as the interactions between textual and visual elements. \vspace{-0.3em}
\end{itemize}

By combining these three learning objectives in a multitask learning framework, our foundation model learns to handle different levels of detail and semantic understanding. This strategic alignment enables our model to deal with various spatial details, distinguish levels of detail in understanding, and go beyond surface-level recognition—ultimately learning a universal representation for vision understanding.

\section{Model}

We present the foundation model \flmodel, designed for universal representation learning, capable of handling various vision tasks with a single set of weights and a unified architecture. As depicted in~\cref{fig:model_pipeline}, \flmodel employs a sequence-to-sequence learning paradigm~\cite{vaswani2017attention}, integrating all tasks, described in~\cref{sec:task}, under a common language modeling objective. The model takes images coupled with task-prompt as task instructions, and generates the desirable results in text forms. It uses a vision encoder to convert images into visual token embeddings, which are then concatenated with text embeddings and processed by a transformer-based multi-modal encoder-decoder to generate the response. In the following sections, we will provide a detailed explanation of each model component.

\comment{
We subsequently introduce our \flmodel foundation model, specifically designed for comprehensive multitask learning that can handle various vision tasks with one model and one set of weights.
As shown in~\cref{fig:model_pipeline}, \flmodel enjoys a simple yet effective architecture by leveraging a sequence-to-sequence learning paradigm~\cite{vaswani2017attention} that integrates all the tasks described in~\cref{sec:task} under a common language modeling objective.
This is different from previous multitask learning methods~\cite{he2017mask}, which use separate task-specific heads.
 The model accepts an image and a task prompt as inputs and produces a response that is relevant to the task.
It consists of a vision encoder that transforms images into visual token embeddings, which are then concatenated with text embeddings and fed into a transformer-based multi-modal encoder-decoder to generate the response.
In the following sections, we will explain each component of our model in detail.}

\paragraph{Task formulation.} 
We adopt a sequence-to-sequence framework~\cite{vaswani2017attention,lu2022unifiedio,chen2022pali, chen2022pix2seq} to address various vision tasks in a unified manner. As shown in~\cref{tab:tasks}, we formulate each task as a translation problem: Given an input image and a task-specific prompt, we generate the corresponding output response. Depending on the task, the prompt and response can be either text or region:

\begin{itemize}
    \item \textbf{Text}: When the prompt or answer is plain text without special formatting, we maintain it in our final sequence-to-sequence format.
    \item \textbf{Region}: 
    For region-specific tasks, we add location tokens to the tokenizer's vocabulary list, representing quantized coordinates. We create $1,000$ bins, similar to~\cite{chen2022pix2seq, lu2022unifiedio, chen2022unified, wang2022ofa}, and represent regions using formats tailored to task requirements:

    \begin{itemize}
    \item \textbf{Box representation $(x_0, y_0, x_1, y_1)$}: Utilized in tasks such as object detection and dense region captioning, with location tokens corresponding to the box coordinates. The location tokens are the coordinates of the top-left and bottom-right corners of the box.
    \item \textbf{Quad box representation $(x_0, y_0, ..., x_3, y_3)$}: For text detection and recognition tasks, using location tokens for each coordinate of the quadrilateral enclosing the text. The location tokens are the coordinates of each corner of the quad box, starting from the top-left and going clockwise.
    \item \textbf{Polygon Representation $(x_0, y_0, ..., x_n, y_n)$}: For referring segmentation tasks, with location tokens representing the vertices of the polygon. The location tokens are the coordinates of the vertices of the polygon, in clockwise order. 
    \end{itemize}
\end{itemize}

By extending the tokenizer's vocabulary to include location tokens, we enable the model to process region-specific information in a unified learning format. This eliminates the need to design task-specific heads for different tasks and allows for a more data-centric approach.

\paragraph{Vision encoder.}
We employ DaViT~\cite{ding2022davit} as the vision encoder. It processes an input image $\mathbf{I} \in \mathbb{R}^{H \times W \times 3}$ (with $H$ and $W$ denoting height and width, respectively) into flattened visual token embeddings $\mathbf{V} \in \mathbb{R}^{N_v \times D_v}$, where $N_v$ and $D_v$ represent the number and dimensionality of vision tokens, respectively.

\paragraph{Multi-modality encoder decoder.}
We use a standard encoder-decoder transformer architecture to process visual and language token embeddings. 
We first obtain prompt text embeddings $\mathbf{T}_{prompt} \in \mathbf{R}^{N_t \times D}$ using our extended language tokenizer and word embedding layer~\cite{lewis2019bart}.
Then, we concatenate vision token embeddings with prompt embeddings to form the multi-modality encoder module input, $\mathbf{X} = [\mathbf{V}', \mathbf{T}_{prompt}]$, where $\mathbf{V}' \in \mathbb{R}^{N_v \times D}$ is obtained by applying a linear projection and LayerNorm layer~\cite{ba2016layer} to $\mathbf{V}$ for dimensionality alignment.

\paragraph{Optimization objective.}
Given the input $x$ combined from the image and the prompt, and the target $y$, we use the standard language modeling with cross-entropy loss for all the tasks. 

\begin{equation}
    \mathcal{L}  = -  \sum_{i=1}^{|y|}logP_{\theta}(y_i|y_{<i}, x),
\end{equation}
where $\theta$ are the network parameters, $|y|$ is the number of target tokens.

\begin{figure*}[t]
    \centering
    \includegraphics[width=1.0\textwidth]{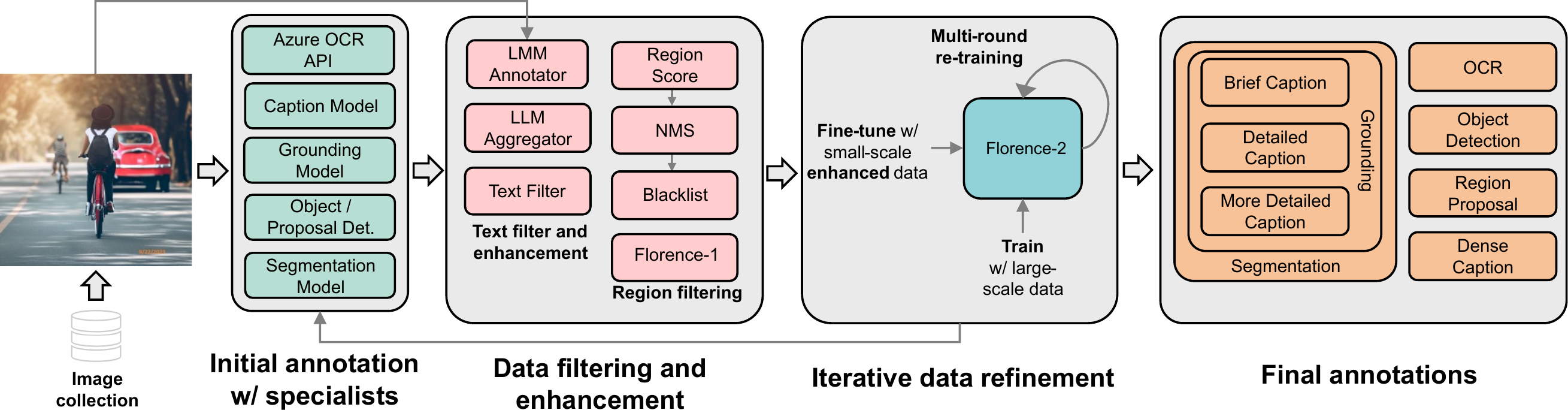}
    \caption{\textbf{\flmodel data engine} consists of three essential phrases: (1) initial annotation employing specialist models, (2) data filtering to correct errors and remove irrelevant annotations, and (3) an iterative process for data refinement. Our final dataset (\textbf{\fld}) of over \textbf{5B} annotations contains \textbf{126M} images, \textbf{500M} text annotations, \textbf{1.3B} region-text annotations, and \textbf{3.6B} text-phrase-region annotations.}
    \label{fig:data_pipeline}
\end{figure*}

\section{Data Engine}
\label{sec:data_engine}
To train our \flmodel model, we require a comprehensive, large-scale, high-quality multitask dataset encompassing various image data aspects. Given the scarcity of such data, we have developed a new multitask image dataset. This dataset \fld includes \textbf{126M} images, \textbf{500M} text annotations, and \textbf{1.3B} text-region annotations, and \textbf{3.6B} text-phrase-region annotations across different tasks. We extensively explain our data collection and annotation procedures, encompassing adaptations for various annotation types. The data engine pipeline, shown in ~\cref{fig:data_pipeline}, will be discussed in subsequent sections.

\subsection{Image Collection}

We construct our data by gathering a diverse collection of images from various sources.
We begin with the identification of three key tasks that act as primary sources for our image corpus: image classification, object detection, and image captioning.
Consequently, we curate and combine five distinct datasets originating from the aforementioned tasks: ImageNet-22k \cite{deng2009imagenet}, Object 365 \cite{shao2019objects365}, Open Images \cite{kuznetsova2020open}, Conceptual Captions \cite{sharma2018conceptual}, and LAION \cite{schuhmann2021laion} filtered by~\cite{li2022blip}. 
This combination results in a dataset of 126 million images in total.

\subsection{Data Annotation}
\label{subsec:data_annotation}
\begin{figure*}
    \centering
    \includegraphics[width=1.0\textwidth]{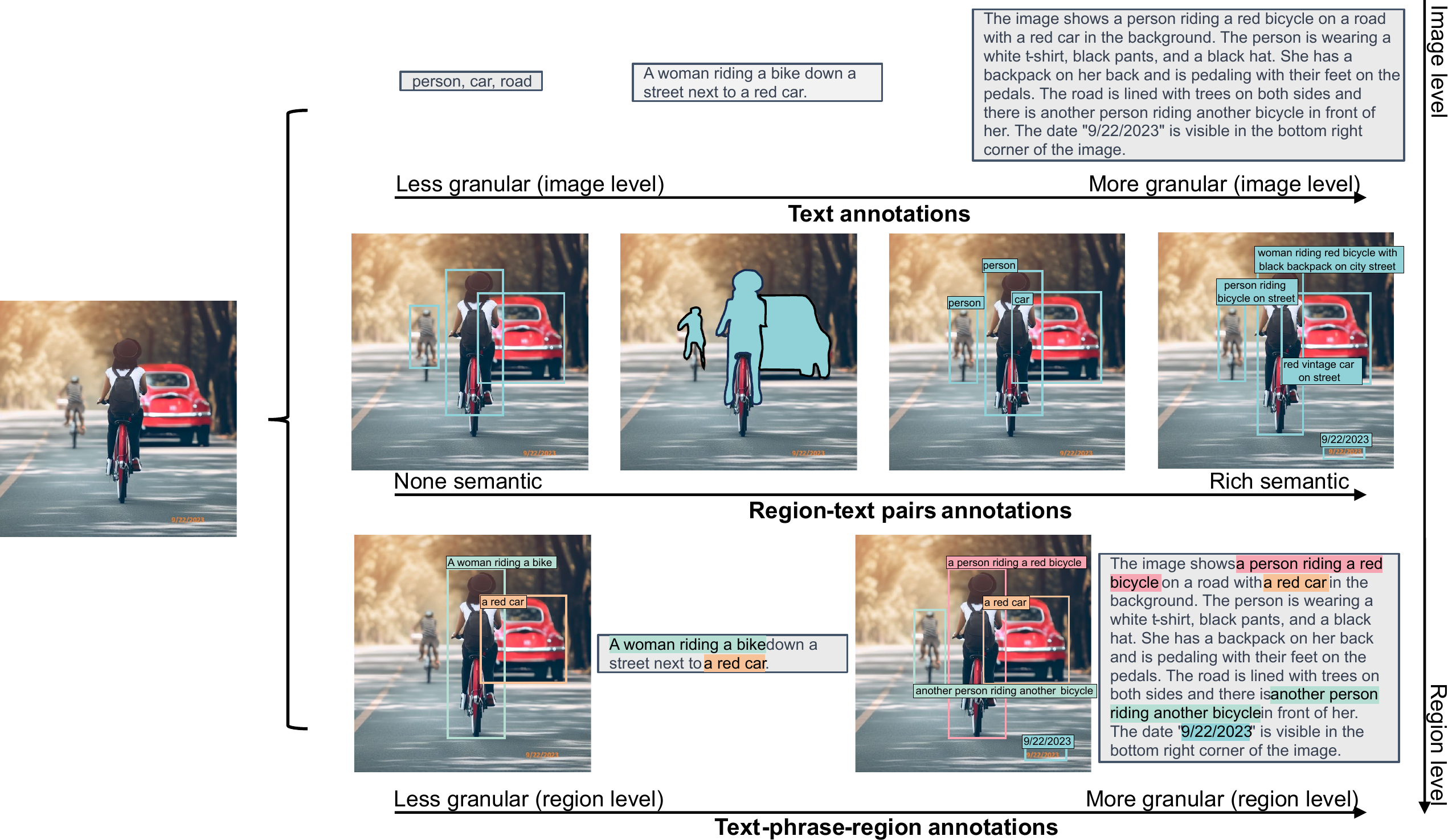}
    \caption{An illustrative example of an image and its corresponding annotations in \fld dataset. Each image in \fld is annotated with text, region-text pairs, and text-phrase-region triplets by Florence data engine, which covers multiple spatial hierarchies, brief-to-detailed progressive granularity, and a wide semantics spectrum, enabling more comprehensive visual understanding from diverse perspectives.}
    \label{fig:data_annotations}
\end{figure*}

Our primary objective is to generate comprehensive annotations that can support multitask learning effectively.
Accordingly, our annotation endeavors span a comprehensive range of tasks, encapsulated within three discrete annotation categories: \textit{text}, \textit{region-text} pairs, and \textit{text-phrase-region} triplets, which is illustrated in~\cref{fig:data_annotations}. 
The data annotation workflow consists of three essential phases, each of which ensures the accuracy and quality of the annotations: (1) initial annotation employing specialist models, (2) data filtering to correct errors and remove irrelevant annotations, and (3) an iterative process for data refinement.

\paragraph{Initial annotation with specialist models.}
To initiate the annotation process for each annotation type, we employ synthetic labels obtained from specialist models. 
These specialist models are a combination of offline models trained on a diverse range of publicly available datasets and online services hosted on cloud platforms. They are specifically tailored to excel in annotating their respective annotation types.

It is worth noting that certain image datasets may already contain partial annotations for some annotation types. For instance, the Object 365~\cite{shao2019objects365} dataset already includes human-annotated bounding boxes and corresponding categories as region-text annotations. In such cases, we merge the pre-existing annotations with the synthetic labels generated by the specialist models. This approach enhances the coverage and diversity of the annotations. 

Moreover, specific annotations, such as detailed descriptions in the text annotation type, are represented by datasets of a considerably small size. This inherently poses challenges in obtaining high-performance specialist models. Consequently, we opt to omit these tasks during the initial annotation phase. Annotations for these tasks are generated later during the iterative data refinement process.

In summation, through the rigorous initial annotation procedures, we ensure that the aggregated dataset of 126 million images is comprehensively labeled across the majority of annotation types.

\paragraph{Data filtering and enhancement.}
The initial annotations obtained from the specialist models, while comprehensive, are susceptible to noise and imprecision. In response to this challenge, we have implemented a multifaceted filtering process to refine and eliminate undesired annotations. Our general filtering protocol mainly focuses on two data types in the annotations: text and region data. 

First, pertaining to textual annotations, we are inspired by DiHT \cite{radenovic2023filtering} and develop a parsing tool based on SpaCy \cite{honnibal2020spacy} to extract objects, attributes, and actions. We filter out texts containing excessive objects, as they tend to introduce noise and may not accurately reflect the actual content in the corresponding images. Additionally, we assess the complexity of the actions and objects by measuring their degree of node in the dependency parsing tree. We retain texts with a certain minimum action and object complexity to ensure the richness of visual concepts in the images. 

Second, in relation to the region annotations, specifically bounding boxes, we remove the noisy boxes under a confidence score threshold. Complementing this, we also employ non-maximum suppression to reduce redundant or overlapping bounding boxes. 

\paragraph{Iterative data refinement.}
Using our filtered initial annotations, we trained a multitask model that processes sequences of data.
Upon evaluating this model against our training images, we discerned a marked enhancement in its predictions, particularly in instances where original labels were marred by inaccuracies or extraneous noise, such as in alt-texts. Motivated by these findings, we integrated these updated annotations with our original ones and subjected the model to another training iteration. This cyclical refinement process incrementally improves the quality of our training dataset.

In the case of tasks we initially bypassed due to insufficient data for the training of a robust specialist model, we leveraged the iteratively trained model for pre-training purposes. Subsequent fine-tuning of this pre-trained model with the sparse dataset showcased superior performance compared to a model trained from scratch on the same data. Thus, we harness the fine-tuned model as a specialist for annotating our expansive dataset comprising 126 million images, ensuring comprehensive annotation coverage.

\subsection{Annotation-specific Variations}

\label{subsec:task_var}
In ~\cref{subsec:data_annotation}, we introduce our general annotation workflow. This section delves into each annotation type and the corresponding variations of the annotation procedure.

\paragraph{Text.}  Text annotations categorize images using three types of granularities: brief, detailed, and more detailed. The brief text includes only one sentence that demonstrates the most salient objects and activities, which is similar to COCO caption~\cite{chen2015microsoft}. In contrast, the detailed text and more detailed text contain multiple sentences that describe the image with richer objects, attributes, and actions. 

For the brief text, a \flmodel model is trained as the specialist on publicly available image caption and image-text datasets, creating an image-to-text model for initial annotations. Iterative refinement is used to minimize noise in these texts. For the detailed text, prompts including existing image annotations like the brief text and region-text annotations, are fed to large language models (LLMs) or large multimodal models (LMMs) to generate comprehensive descriptions. Due to the high cost of the large models, only a small set of detailed text and more detailed text are generated. These are used to fine-tune the caption specialist, developing a detailed description specialist for further annotations.

\paragraph{Region-text pairs.} The region-text pairs provide descriptive textual annotation for semantic regions in the image. Semantic regions include regions of visual objects as well as text regions. The region is represented by a tight bounding box surrounds the region. Moreover, each region can be annotated with varying degrees of granularity, including phrases and sentences, that contribute to a richer understanding of the region.

Region-text pairs are annotated differently for text regions and visual object regions. Text regions are labeled using Azure AI Services' OCR API~\cite{azure_ai_services}, while visual objects are initially annotated with a DINO object detector~\cite{zhang2022dino} trained on public datasets. Data filtering, including confidence thresholding and non-maximum suppression, removes noisy boxes. Textual annotations for the visual object regions are further enriched by brief text generated from an image-to-text model with cropped image regions. Each region then receives three textual annotations: phrase from object category, brief text, and noun phrase chunks from the brief text. The Florence-1~\cite{yuan2021florence} model determines the most similar textual annotation to each image region.

 \paragraph{Text-phrase-region triplets.} Text-phrase-region triplets consist of a descriptive text of the image, noun phrases in this text related to image objects, and region annotations for these objects. The text includes brief, detailed, and more detailed text generated earlier. For each text, the Grounding DINO model~\cite{liu2023grounding} identifies noun phrases and creates bounding boxes for them. Additionally, the SAM model~\cite{kirillov2023segment} generates segmentation masks for each box, offering more precise object localization. During data filtering, a confidence score threshold is applied to both noun phrases and bounding boxes to ensure relevance. A blacklist is also used to exclude irrelevant noun phrases like pronouns and abstract concepts.

\begin{table*}[t]
\centering
\setlength{\tabcolsep}{7.5pt}%
\begin{tabular}{l|l|r|r|l|l}
\toprule
Dataset & Rep. Model & \#Images & \#Annotations & Spatial hierarchy &  Semantics granularity \\
\midrule

JFT300M~\cite{dosovitskiy2021image} & ViT & 300M & 300M & Image-level & Coarse  \\ %
WIT~\cite{radford2021learning} & CLIP & 400M & 400M & Image-level & Coarse         \\ %
SA-1B~\cite{kirillov2023segment} & SAM & 11M  & 1B & Region-level &Non-semantic \\ %
GrIT~\cite{peng2023kosmos} & Kosmos-2 & 91M  & 137M & Image \& Region-level & Fine-grained \\ %
M3W~\cite{alayrac2022flamingo} & Flamingo & 185M & 43.3M* & Multi-image-level & Fine-grained                     \\ %
\rowcolor{gray!20}
\fld (ours) & \flmodel (ours) & 126M & 5B & Image \& Region-level &  Coarse to fine-grained \\
\bottomrule
\end{tabular}
\caption{Comparison with datasets in vision foundation model training. *Flamingo's annotations are counted in the number of documents, where each document may have multiple images.}
\label{table:dataset_comparision}
\end{table*}

\section{Dataset}
This section introduces the statistics and analysis of \fld that we built using the data engine in ~\cref{sec:data_engine}. We begin with an overview of the dataset and compare it with the recent works. We then show further analyses of detailed annotation statistics, semantic coverage and spatial coverage in the established dataset.

\subsection{Overview}
Following the data engine, we build a large-scale training set (\fld) of 126M images, more than \textbf{500M} text annotations, \textbf{1.3B} region-text annotations, and \textbf{3.6B} text-phrase-region annotations. Each image is annotated with text, region-text pairs, and text-phrase-region triplets and each annotation type has multiple instances varying in diverse granularity. An illustrative example of an image and its corresponding annotations can be found in~\cref{fig:data_annotations}.

We provide a comparison between our data set and the existing data sets that are commonly used for training foundation models in~\cref{table:dataset_comparision}. Our data set has several advantages over the previous ones, such as having more annotations in total and per image. Moreover, the annotations in our data set span multiple levels of spatial and semantic granularity, which allows for more diverse and comprehensive visual understanding tasks.

\subsection{Data Analysis}

\paragraph{Annotation statistics.} 
The statistics for each annotation type within our dataset are presented in \cref{table:ann_stats}.

Firstly, we have around \textbf{500M} text annotations, including brief, detailed, and more detailed texts with different lengths. It is noteworthy that our detailed and more detailed text has 4x and 9x number of tokens compared with the brief text that is similar to COCO captions~\cite{chen2015microsoft}. These lengthy annotations provide much richer information for comphrensive visual understanding.

In addition, our dataset has around \textbf{1.3B} region-text annotations, which is more than 30x larger than the academic object detection datasets such as OpenImages~\cite{kuznetsova2020open} and Object 365~\cite{shao2019objects365}. On average, each image has around 5 regions, and each region is annotated with either a phrase or a relatively longer brief text. Note that the regional brief text (2.55 avg tokens) is shorter than typical brief text annotation (7.95 avg tokens), as the regional brief text annotation actually includes a mixture of phrase, noun chunks, and brief text based on the Florence-1 score. More details can be found from~\cref{subsec:task_var} - region-text pairs.

Moreover, we collect text-phrase-region annotations that include more than \textbf{3.6B} phrase-region pairs for the \textbf{500M} text annotations. Specifically, the brief text annotation has 4.27 average phrase-region pairs, while detailed and more detailed text annotation has more than 10 pairs, indicating that the richer text annotation covers more objects and their corresponding phrases in the text.

\begin{table*}[t]
\centering
\setlength{\tabcolsep}{5.5pt}%
\small
\renewcommand{\arraystretch}{1.35}

\begin{tabular}{l|l|r|r|r|r|r}
\toprule
Annotation Type & Text Type & \#Image Annotations & \#Avg Tokens & \#Regions & \#Avg Regions  & \#Avg Regional Tokens          \\ %

\midrule
Text                & Brief          & 235M  & 7.95  & -     & -      & -     \\ 
                    & Detailed       & 126M  & 31.65 & -     & -      & -     \\ 
                    & More detailed  & 126M  & 70.53 & -     & -      & -     \\ %
\midrule
Region-Text         & Phrase         & 126M  & -     & 681M  & 5.42   & 1.19  \\ 
                    & Brief          & 126M  & -     & 681M  & 5.42   & 2.55  \\ %
\midrule
Text-Phrase-Region  & Brief          & 235M  & 7.95  & 1007M & 4.27   & 1.93  \\ 
                    & Detailed       & 126M  & 31.65 & 1289M & 10.25  & 1.49  \\ 
                    & More detailed  & 126M  & 70.53 & 1278M & 10.17  & 1.35  \\ %
\bottomrule
\end{tabular}
\caption{Annotation statistics of \fld dataset.}
\label{table:ann_stats}
\end{table*}

\paragraph{Semantic coverage.} Our text annotations comprise various text types, addressing different levels of detail. To assess semantic coverage, we employ SpaCy~\cite{honnibal2020spacy} for tokenization and parsing, inspired by DiHT~\cite{radenovic2023filtering}. This process yields part-of-speech (POS) tags and the dependency parsing tree among tokens. We establish heuristic rules based on POS tags, categorizing tokens into semantic element types, \eg, objects, attributes, actions, and proper nouns. Additionally, we introduce the concept of \textit{token complexity}, measured by the total degrees of the token in the dependency parsing tree when treated as an undirected graph. This complexity reflects the richness of semantic connections. In our study, we focus on measuring the complexity of objects and actions. 

\Cref{table:semantic_coverage} presents the statistics on the average number of semantic elements and their corresponding complexity. The results indicate that all measurements increase with the inclusion of more details in text annotations. Notably, average actions experience the most significant boost, with detailed and more detailed text exhibiting 7$\times$ and 15$\times$ increases, respectively, compared to brief text. This highlights the limitations of traditional brief text annotations in describing image actions. Conversely, the increment in proper nouns is relatively low, potentially because specialists often describe objects more generally than using specific proper nouns. In terms of complexity measurements, both objects and actions show more semantic connections in detailed text annotations. The complexity of actions exhibits a higher improvement, aligning with our observation of the increasing number of actions.

\begin{table}[t]
\centering
\setlength{\tabcolsep}{5.1pt}
\small
\begin{tabular}{l|r|r|r}

\toprule
Text Type             & Brief & Detailed & More detailed \\ 
\midrule
\#Image Annotations   & 235M & 126M & 126M \\
\#Avg Tokens          & 7.95 & 31.65 & 70.53 \\
\#Avg Objects         & 3.23 & 13.31 & 28.06 \\
\#Avg Attributes      & 2.80 & 7.27 & 16.25 \\
\#Avg Actions         & 0.58 & 4.21 & 8.76 \\
\#Proper Nouns        & 1.10 & 2.40 & 2.41 \\
Avg Object Complexity & 2.80 & 4.00 & 4.02 \\
Avg Action Complexity & 1.14 & 3.63 & 4.38 \\ 
\bottomrule

\end{tabular}
\caption{Statistics of the average number of semantic elements
and corresponding complexity in \fld dataset.}
\label{table:semantic_coverage}
\end{table}

\begin{figure*}
    \centering
    \begin{subfigure}[b]{0.24\linewidth}
        \includegraphics[width=0.95\textwidth]{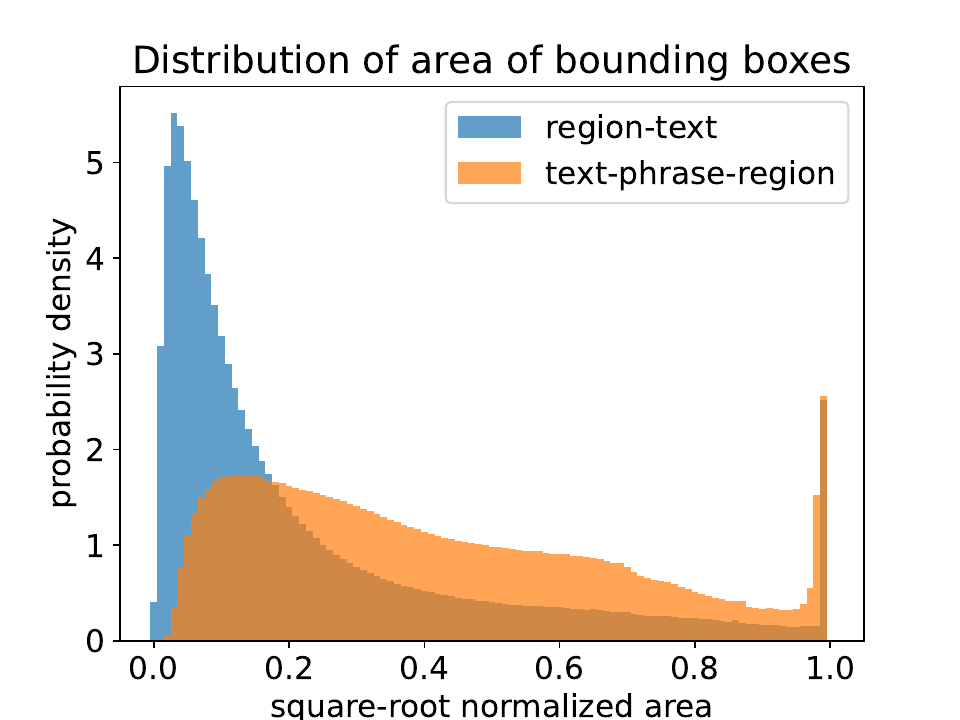}
        \caption{}
        \label{fig:sc-a}
    \end{subfigure}
    \begin{subfigure}[b]{0.24\linewidth}
        \includegraphics[width=0.95\textwidth]{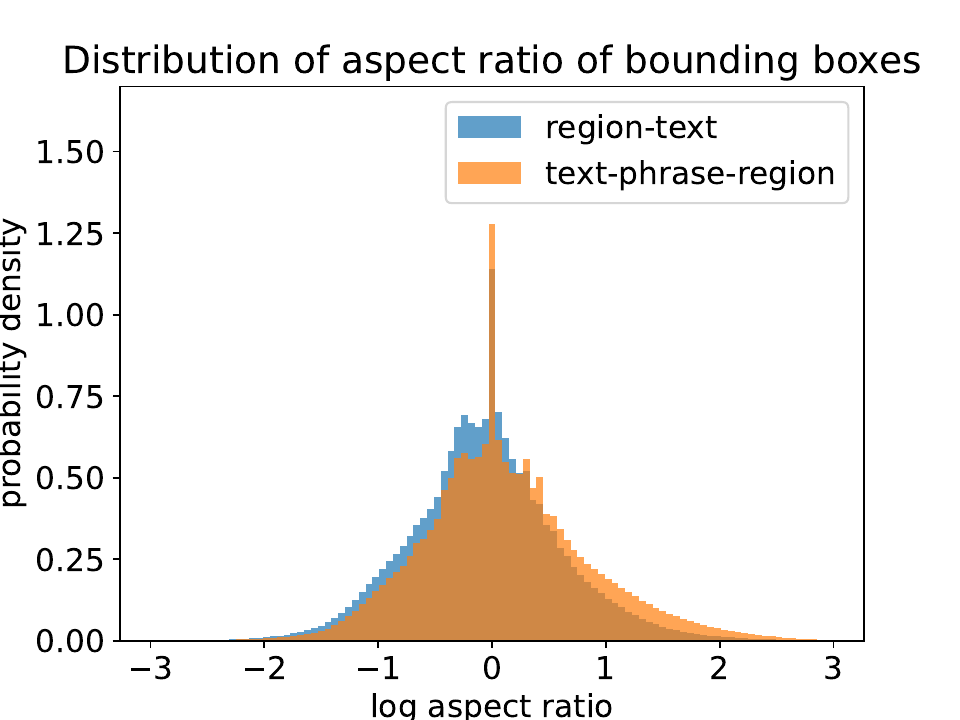}
        \caption{}
        \label{fig:sc-b}
    \end{subfigure}
    \begin{subfigure}[b]{0.24\linewidth}
        \includegraphics[width=0.95\textwidth]{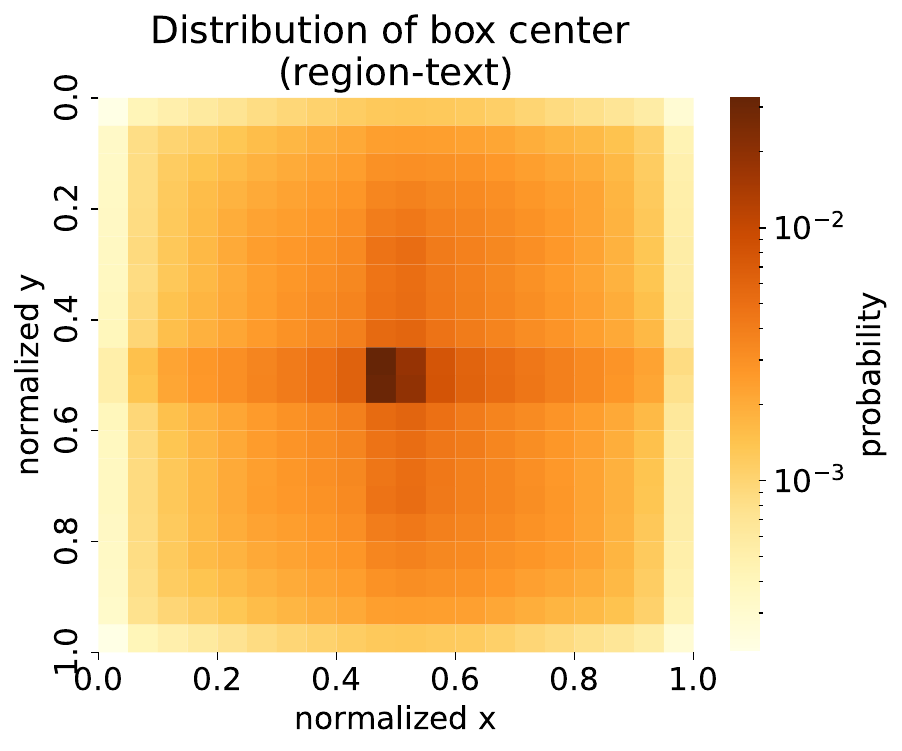}
        \caption{}
        \label{fig:sc-c}
    \end{subfigure}
    \begin{subfigure}[b]{0.24\linewidth}
        \includegraphics[width=0.95\textwidth]{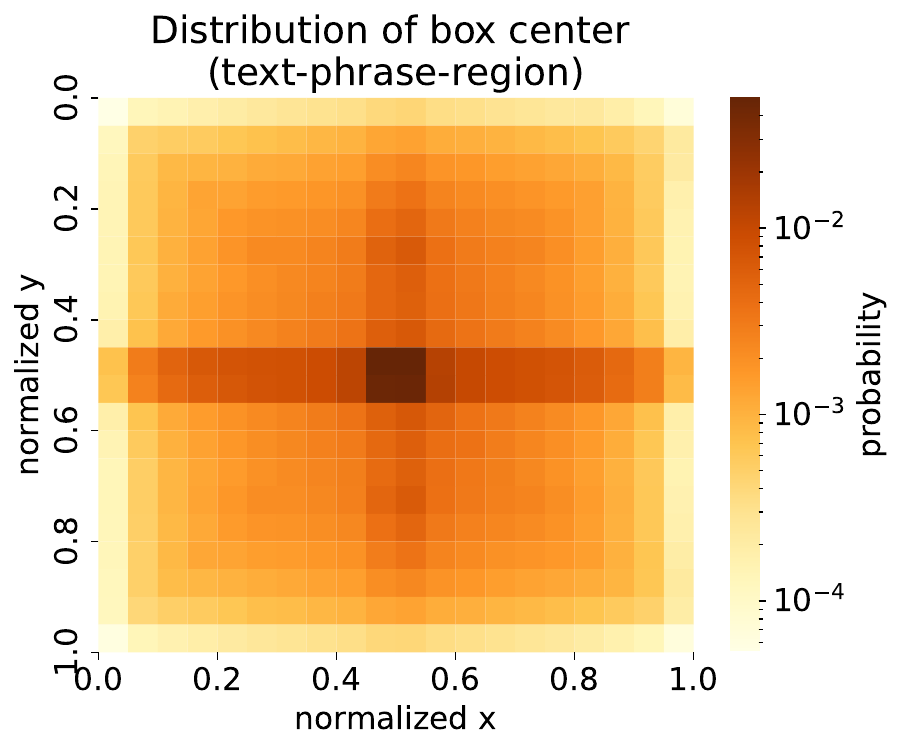}
        \caption{}
        \label{fig:sc-d}
    \end{subfigure}
    \caption{Distributions of bounding boxes in \fld dataset.}
    \label{fig:spatial_coverage}
\end{figure*}

\paragraph{Spatial coverage.} Our region-text and text-phrase-region annotations, represented by bounding boxes and masks, capture the location of visual concepts within images. The distribution of box areas, as shown in~\cref{fig:sc-a}, reveals more small boxes in region-text pairs and a uniform box size distribution in text-phrase-region triplets. This difference stems from the the divergent origins of these boxes: object detectors for region-text pairs and a grounding model for text-phrase-region triplets, which aligns boxes to textual phrases representing both localized and overarching image concepts. In~\cref{fig:sc-b}, the log-format distribution of aspect ratios is illustrated. Region-text pairs and text-phrase-region triplets exhibit similar symmetric distributions, covering a wide range of aspect ratios. Heatmaps of the box center for each annotation type, shown in~\cref{fig:sc-c,fig:sc-d}, indicate a center bias, with region-text pairs displaying a more uniform distribution than text-phrase-region triplets.

\section{Experiments}
Our \flmodel models are trained on~\fld to learn a universal image representation. We conduct our experiments in three main parts: 
(1) We evaluate the \textbf{\emph{zero-shot}} performance of our method on various tasks to show its inherent ability to handle multiple tasks without any extra fine-tuning on task-specific data using \textbf{\emph{one single generalist}} model.
(2) We show the adaptability of our method by further training \textbf{\emph{one single generalist}} model with additional supervised data on a wide range of tasks, achieving competitive state-of-the-art performance.
(3) We examine the performance of the learned visual representation on the downstream tasks as the backbone to show the superiority of our pre-training method over previous approaches.

\subsection{Setup}
We investigate two model variants with different sizes: \flmodelb model with 232 million parameters and \flmodell model with 771 million parameters. The detailed architectures of each model are given in~\cref{table:model_sizes}. We initialize the weights of the image encoder and multi-modality encoder-decoder from UniCL~\cite{yang2022unified} and BART~\cite{lewis2019bart}, respectively.

We adopt AdamW~\cite{loshchilov2019decoupled} with cosine learning rate decay~\cite{loshchilov2017sgdr} for training our models.
We leverage Deepspeed~\cite{rasley2020deepspeed} and mixed precision to improve the training efficiency. 
The maximum learning rate is set at $1e-4$ for the base model and $1e-5$ for the large model.
A linear warm-up to the maximum learning rate is applied during the first 5,000 optimization steps.

We train our models with a mini-batch size of 2048/3072 (base/large) and an image size of 384$\times$384 until reaching 3 billion effective training samples. Similar to~\cite{chen2022pali, radford2021learning, jia2021scaling, yuan2021florence, yu2022coca}, we further conduct high-resolution tuning with an image size of 768$\times$768 for 0.5 billion samples for the base model and 0.1 billion samples for the large model. 

\subsection{Zero-shot Evaluation Across Tasks}

\begin{table*}[ht]
\centering
\small
\setlength{\tabcolsep}{2.4pt}
\begin{tabular}{l|r|ccccccccccccccccccccccc} 
\toprule
\multirow{3}{*}{Method}                  & \multirow{3}{*}{\#params} && \rotatebox{90}{COCO Cap.} &  & \rotatebox{90}{NoCaps} &  & \rotatebox{90}{TextCaps} &  & \rotatebox{90}{COCO Det.} &  & \rotatebox{90}{Flickr30k} &  & \multicolumn{3}{c}{\rotatebox{90}{Refcoco}} &  & \multicolumn{3}{c}{\rotatebox{90}{Refcoco+}} &  & \multicolumn{2}{c}{\rotatebox{90}{Refcocog}} &  & \multicolumn{1}{c}{\rotatebox{90}{Refcoco RES}}  \\ 
\cline{4-4}\cline{6-6}\cline{8-8}\cline{10-10}\cline{12-12}\cline{14-16}\cline{18-20}\cline{22-23}\cline{25-25}
 & && test  && val && val && val2017 && test && val & test-A & test-B && val & test-A & test-B && val & test && val                      \\ 
 & && CIDEr && CIDEr && CIDEr && mAP && R@1  && \multicolumn{3}{c}{Accuracy} && \multicolumn{3}{c}{Accuracy}  && \multicolumn{2}{c}{Accuracy} && mIoU                      \\ 
\midrule
Flamingo~\cite{alayrac2022flamingo} & 80B && 84.3 &  & - &  & - &  & - &  & - &  & - & - & - &  & - & - & - &  & -  & - &  & -  \\
Kosmos-2~\cite{peng2023kosmos} & 1.6B  && - &  & - &  & - &  & - &  & 78.7 &  & 52.3 & 57.4  & 47.3 &  & 45.5 & 50.7  & 42.2 &  & 60.6 & 61.7 &  & -                         \\
\midrule
\flmodelb                     & 0.23B                      && 133.0             &  & 118.7     &  & 70.1         &  & 34.7    &  & 83.6      &  & 53.9 & 58.4  & 49.7      &  & 51.5 & 56.4  & 47.9       &  & 66.3 & 65.1                &  & 34.6                    \\
\flmodell                    & 0.77B                      & &135.6             &  & 120.8     &  & 72.8         &  & 37.5    &  & 84.4      &  & 56.3 & 61.6  & 51.4      &  & 53.6 & 57.9  & 49.9       &  & 68.0 & 67.0                &  & 35.8                    \\
\bottomrule
\end{tabular}
\caption{\textbf{Zero-shot} performance of generalist vision foundation models. The models do not see the training data of the evaluation tasks during training. \flmodel models are pre-trained on \fld dataset.
Karpathy test split is used for COCO caption evaluation.  }
\label{tab:table_sota}
\end{table*}
We present a powerful vision foundation model that does not require task-specific supervised annotations for fine-tuning.
The \textbf{zero-shot} performance of our model is shown in ~\cref{tab:table_sota}. 
For image-level tasks, \flmodell achieves a 135.6 CIDEr score on the COCO caption benchmark~\cite{lin2014microsoft}, utilizing less than 1\% of the parameters compared to the 80B Flamingo~\cite{alayrac2022flamingo} model (which has an 84.3 CIDEr score). For region-level grounding and referring expression comprehension tasks, \flmodell establishes a new record in zero-shot performance achieving a 5.7 improvement in Flickr30k~\cite{plummer2015flickr30k} Recall@1, and approximately 4\%, 8\%, and 8\% absolute improvements on Refcoco, Refcoco+, and Refcocog~\cite{10.1007/978-3-319-46475-6_5}, respectively, compared to the Kosmos-2~\cite{peng2023kosmos} model, which has 1.6B parameters. 
Additionally, our pre-trained model attains a 35.8\% mIOU in the Refcoco referring expression segmentation (RES)~\cite{10.1007/978-3-319-46475-6_5} task, a capability not supported by prior foundation models.
\subsection{Generalist Model with Public Supervised Data}

\begin{table*}[t]
    \centering
    \setlength{\tabcolsep}{9.4pt}%
    \small
    \begin{tabular}{l|r|cccccc}
    \toprule
    \multirow{3}{*}{Method} & \multirow{3}{*}{\#params} & COCO Caption & NoCaps & TextCaps & VQAv2 & TextVQA & VizWiz VQA \\
    &&Karpathy test & val & val & test-dev & test-dev & test-dev \\
    &&CIDEr & CIDEr & CIDEr & Acc & Acc & Acc \\
    \midrule
    \multicolumn{8}{c}{\textbf{\textit{Specialist Models}}} \\
    \midrule
    CoCa~\cite{yu2022coca} & 2.1B & 143.6 & 122.4  &-&  82.3   & - & -\\
    BLIP-2~\cite{li2023blip} & 7.8B & 144.5 & 121.6 &- &82.2&-&- \\
    GIT2~\cite{wang2022git} & 5.1B & 145 & 126.9 & 148.6 & 81.7 & 67.3 & 71.0 \\
    Flamingo~\cite{alayrac2022flamingo}  & 80B & 138.1 &-& -& 82.0 & 54.1 & 65.7 \\
    PaLI~\cite{chen2022pali}   & 17B & 149.1 & 127.0 & 160.0${^\bigtriangleup}$ & 84.3 & 58.8 / 73.1${^\bigtriangleup}$ & 71.6 / 74.4${^\bigtriangleup}$ \\
    PaLI-X~\cite{chen2023pali}  & 55B & 149.2 & 126.3 & 147 / 163.7${^\bigtriangleup}$ & 86.0 & 71.4 / 80.8${^\bigtriangleup}$ & 70.9 / 74.6${^\bigtriangleup}$ \\
    \midrule
    \multicolumn{8}{c}{\textbf{\textit{Generalist Models}}} \\
    \midrule
    Unified-IO~\cite{lu2022unifiedio} & 2.9B & - & 100 & -& 77.9 & -& 57.4 \\
    \flmodelb & 0.23B & 140.0 & 116.7 & 143.9 & 79.7 & 63.6 & 63.6 \\
    \flmodell & 0.77B & 143.3 & 124.9 & 151.1 & 81.7 & 73.5 & 72.6 \\
    \bottomrule
    \end{tabular}
    \caption{Performance of specialist and generalist models on captioning and VQA tasks. \textbf{\textit{Specialist Models}} refer to those that are fine-tuned specifically for each task, while \textbf{\textit{Generalist Models}} denote a single model fine-tuned in a task-agnostic manner, applicable across all tasks.  ${^\bigtriangleup}$ indicates usage of external OCR as input.}
    \label{tab:sota_ft_1}
\end{table*}
\begin{table*}[t]
    \centering
    \setlength{\tabcolsep}{3.3pt}
    \small
    \begin{tabular}{l|r|ccccccccccccccccc}
    \toprule
    \multirow{3}{*}{Method} & \multirow{3}{*}{\#params} && COCO Det. && Flickr30k && \multicolumn{3}{c}{Refcoco} && \multicolumn{3}{c}{Refcoco+}  && \multicolumn{2}{c}{Refcocog} && \multicolumn{1}{c}{Refcoco RES}  \\
    \cline{4-4}\cline{6-6}\cline{8-10}\cline{12-14}\cline{16-17}\cline{19-19}
    &&&val2017 && test && val & test-A & test-B && val & test-A & test-B && val & test && val  \\
    &&&mAP && R@1 && \multicolumn{3}{c}{Accuracy} && \multicolumn{3}{c}{Accuracy} && \multicolumn{2}{c}{Accuracy} && mIoU  \\
    \midrule
    \multicolumn{19}{c}{\textbf{\textit{Specialist Models}}} \\
    \midrule
    SeqTR~\cite{zhu2022seqtr}  &- &&- &&- && 83.7 & 86.5 & 81.2 && 71.5 & 76.3 & 64.9  && 74.9 & 74.2 && -  \\
    PolyFormer~\cite{liu2023polyformer} &- &&-&&-&& 90.4 & 92.9  & 87.2 &&85.0 & 89.8  & 78.0  && 85.8 & 85.9 && 76.9  \\
    UNINEXT~\cite{yan2023universal} & 0.74B && 60.6 &&- && 92.6 & 94.3  & 91.5  && 85.2 & 89.6 & 79.8 && 88.7 & 89.4 & & -  \\
    Ferret~\cite{you2023ferret} & 13B &&- && -&& 89.5 & 92.4 & 84.4 && 82.8 & 88.1  & 75.2 && 85.8 & 86.3 && - \\
    \midrule
    \multicolumn{19}{c}{\textbf{\textit{Generalist Models}}} \\
    \midrule
    UniTAB~\cite{yang2022unitab} &&&-&&-&& 88.6 & 91.1  & 83.8 && 81.0 & 85.4 & 71.6 && 84.6 & 84.7 && - \\
    \flmodelb & 0.23B && 41.4 && 84.0 && 92.6 & 94.8 & 91.5 && 86.8 & 91.7 & 82.2 && 89.8 & 82.2 && 78.0  \\
    \flmodell & 0.77B && 43.4 && 85.2 && 93.4 & 95.3 & 92.0 && 88.3 & 92.9 & 83.6 && 91.2 & 91.7 && 80.5 \\
    \bottomrule
    \end{tabular}
    \caption{Performance of specialist and generalist models on region-level tasks. \textbf{\textit{Specialist Models}} refer to those that are fine-tuned specifically for each task, while \textbf{\textit{Generalist Models}} denote a single model fine-tuned in a task-agnostic manner, applicable across all tasks.}
    \label{tab:sota_ft_2}
\end{table*}

We demonstrate the versatility and effectiveness of our model as a vision foundation that can be transferred to various downstream tasks. We fine-tune \flmodel models by adding a collection of public datasets that cover image-level, region-level, pixel-level tasks, yielding \textit{one} generalist model for various vision tasks. The details of the dataset collection are provided in ~\cref{table:datasets-by-task}.  ~\Cref{tab:sota_ft_1,tab:sota_ft_2} compare our model with other state-of-the-art models. Our key findings are:

\paragraph{Simple design for strong performance.} \flmodel demonstrates \emph{strong} performance with \emph{standard} multi-modality Transformer encoder-decoder without special designs, particularly for region-level and pixel-level tasks. 
For example, \flmodell outperforms PolyFormer~\cite{liu2023polyformer} on both RefCOCO REC task and RES task by 3.0 Accuracy@0.5 and 3.54 mIOU respectively, where PolyFormer~\cite{liu2023polyformer} adapts specifically designed regression-based prediction head for coordinates. \flmodell also outperforms previous SOTA method UNINEXT~\cite{yan2023universal} on RefCOCO by 0.8 Accuracy@0.5, where UNINEXT~\cite{yan2023universal} is based on advanced object detector Deformable DETR~\cite{zhu2020deformable} and DINO~\cite{zhang2022dino}. 

\paragraph{Competitive performance with fewer parameters.} \flmodell achieves competitive performance without the need for LLMs, showcasing efficiency in handling diverse tasks while maintaining a compact size. For instance, \flmodell attains a CIDEr score of 140.0 on the COCO Caption karpathy test split~\cite{Karpathy2014DeepVA}, outperforming models with significantly more parameters, such as Flamingo (80B parameters, 138.1 CIDEr score).

\paragraph{Adaptable generalization across task levels.} \flmodel demonstrates competitive performance across image-level, pixel-level, and region-level tasks, emphasizing its adaptability and effectiveness in addressing various challenges in computer vision and natural language processing. For example, in the TextVQA task, \flmodell sets a new state-of-the-art performance with an accuracy of 81.5 without any external OCR token input, surpassing previous SOTA methods~\cite{chen2022pali, chen2023pali}. 

These achievements emphasize \flmodel's efficiency in handling diverse tasks while maintaining a compact size, making it a unique and valuable asset in the ever-evolving landscape of AI research and applications.

\subsection{Downstream Tasks Fine-tuning}

In this section, we investigate the performance of our single model fine-tuning on downstream tasks. This experiment highlights the superiority of \flmodel pre-training over previous approaches, as it demonstrates the effectiveness of the learned universal image representation. We use the base size model with about 80M parameters in our experiments to ensure fair comparison with other methods. 

\begin{figure*}[t]
    \centering
    \begin{subfigure}[b]{0.33\linewidth}
        \includegraphics[width=1.0\textwidth]{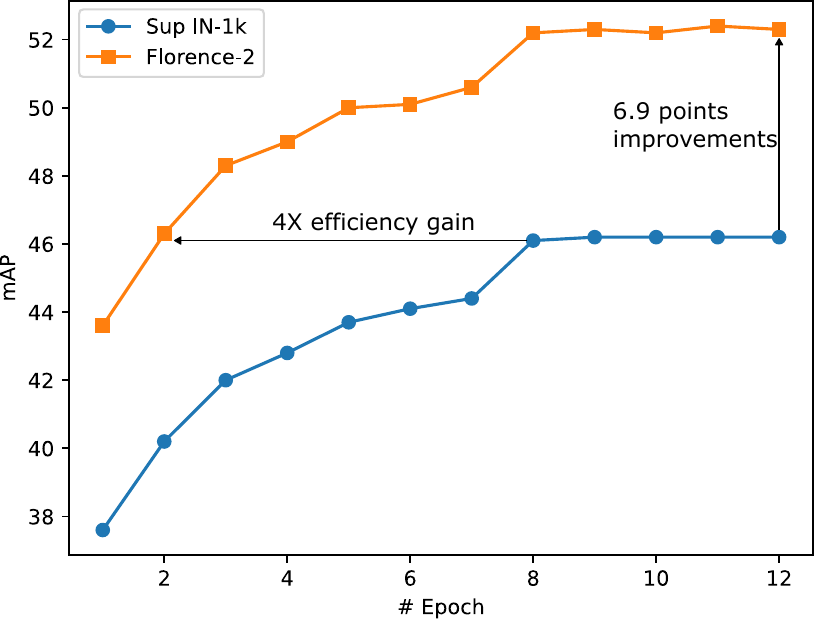}
        \caption{Mask-RCNN on COCO detection.}
        \label{fig:od1_curve}
    \end{subfigure}
    \begin{subfigure}[b]{0.33\linewidth}
        \includegraphics[width=1.0\textwidth]{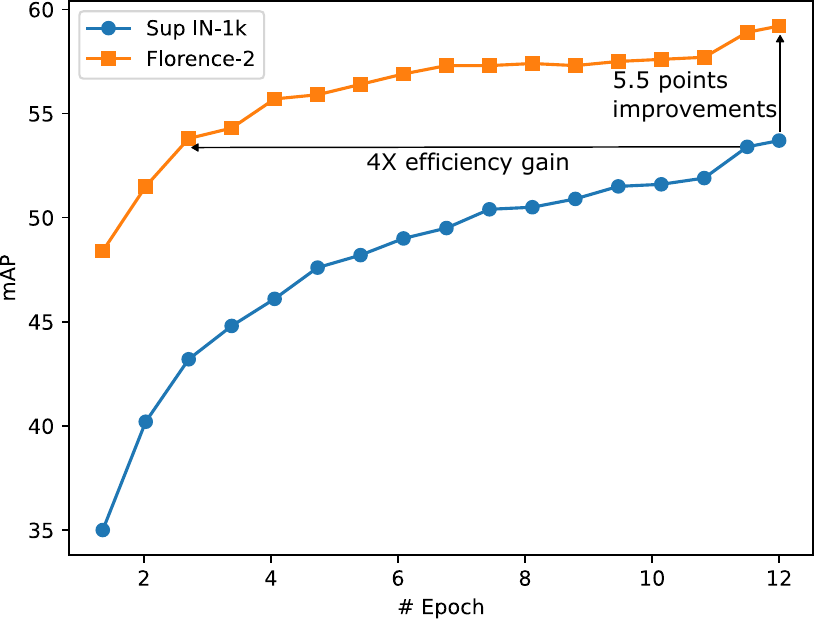}
        \caption{DINO on COCO detection.}
        \label{fig:od2_curve}
    \end{subfigure}
    \begin{subfigure}[b]{0.33\linewidth}
        \includegraphics[width=1.0\textwidth]{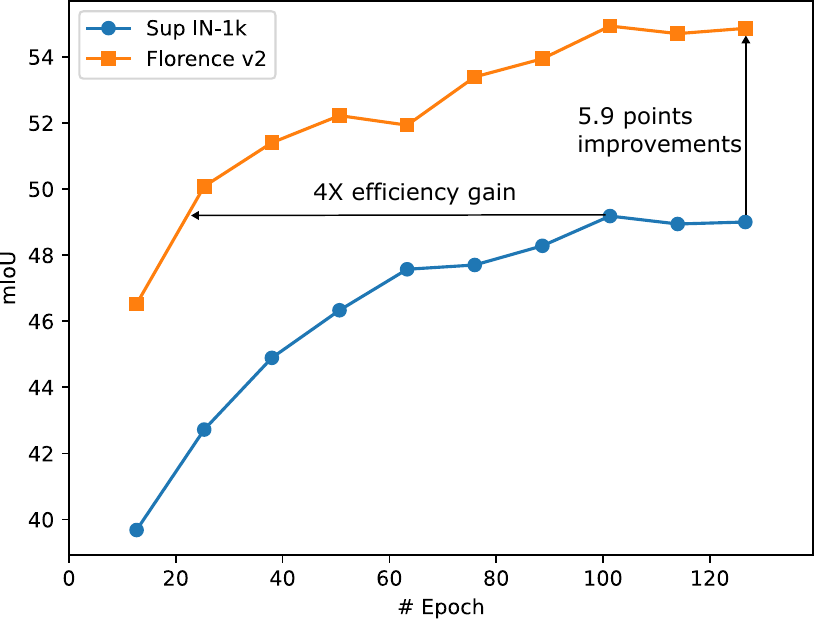}
        \caption{UpperNet on ADE20K.}
        \label{fig:seg_curve}
    \end{subfigure}
    \caption{Training efficiency on COCO object detection and segmentation, and ADE20K semantic segmentation tasks.}
    \label{fig:efficiency_curve}
\end{figure*}

\paragraph{Object detection and segmentation.}
We conduct COCO object detection and instance segmentation~\cite{lin2014microsoft} experiments with Mask R-CNN~\cite{he2017mask}, and COCO object detection~\cite{lin2014microsoft} experiments with DINO~\cite{zhang2022dino} to further demonstrate the effectiveness of \flmodel pre-training.
We train on the \emph{train2017} split and evaluate on the \emph{val2017} split.

For Mask R-CNN~\cite{he2017mask} experiments, we follow the common setup used in~\cite{liu2021swin,zhang2022dino}, we use the standard 1$\times$ (12 epochs) schedule with multi-scale training for all experiments. The learning rate is stepped down by a factor of 0.1 at the 67\% and 89\% of training epochs. 
We do not use any additional augmentation (such as random crop, mosaic, etc) or optimization techniques (such as EMA, weight normalization) during training to ensure a fair comparison. We do not use any test time augmentation (TTA) either. 
Thanks to the strong universal representation learned by \flmodel pre-training, we do not require longer training epochs, such as 36 epochs in~\cite{woo2023convnext, liu2021swin, yang2021focal, yang2022focal}, or 100 epochs in~\cite{li2022exploring}, to achieve better results.

For DINO~\cite{zhang2022dino} experiments, we train DINO-4scale~\cite{zhang2022dino} detector for 12 epochs (1$\times$) using the same data augmentation strategy as employed by~\cite{carion2020end}.

First, our base model achieves a strong performance improvement compared to other approaches. As shown in ~\cref{tab:od_sota_comparison}, our DaViT-B model pre-trained by \flmodel surpasses previous best base model (ConvNext v2-B), which is pre-trained by FCMAE~\cite{woo2023convnext}, by 0.7 $AP_b$ using Mask RCNN. Importantly, while ConvNeXt v2-B leverages a 3$\times$ schedule (36 epochs), our model efficiently employs a 1$\times$ schedule (12 epochs) thanks to our powerful pre-trained universal representation. For DINO framework, our model significantly outperforms the ViT-B, achieving a notable improvement of 4.2 AP.

Second, our pre-training demonstrates higher training efficiency. As shown in \cref{tab:downstram_tasks} and \cref{fig:efficiency_curve}, compared to the model with supervised ImageNet-1k pre-training, our model with \flmodel pre-training achieves 4x efficiency and a significant improvement of 6.9 AP and 5.5 AP with Mask-RCNN and DINO framework, respectively.

Third, our pre-training provides a good generic representation without extensive fine-tuning. ~\Cref{tab:downstram_tasks} indicates that the models with \flmodel pre-training maintains competitive performances when the first two stages are frozen with only 0.3 and 0.2 drops for Mask-RCNN and DINO, respectively. Moreover, our approach with completely frozen backbone can outperform the model with supervised ImageNet-1k pre-training by 1.6 and 2.4 for Mask-RCNN and DINO.

\begin{table}[t]
    \centering
    \setlength{\tabcolsep}{3.0pt}
    \small
        \begin{tabular}{l|c|ccccccc}
        \toprule
        & & & \multicolumn{2}{c}{Mask R-CNN} &&  DINO  \\
        \cline{4-5} \cline{7-7} 
        Backbone & Pretrain && AP$_b$& AP$_m$  && AP \\
        \midrule
        ViT-B~\cite{li2022exploring} & MAE, IN-1k && 51.6 & 45.9 && 55.0 \\ %
        Swin-B \cite{liu2021swin}   & Sup IN-1k   && 50.2   &  -    &&  53.4\\ %
        Swin-B \cite{liu2021swin}   & SimMIM~\cite{xie2022simmim}  && 52.3   &   -     && -\\ %
        FocalAtt-B~\cite{yang2021focal}      & Sup IN-1k   && 49.0   &  43.7   && -  \\ %
        FocalNet-B~\cite{yang2022focal}      & Sup IN-1k   && 49.8   &  44.1   && 54.4 \\ %
        ConvNeXt v1-B~\cite{liu2022convnet}  & Sup IN-1k   && 50.3   &   44.9   && 52.6 \\ %
        ConvNeXt v2-B~\cite{woo2023convnext} & Sup IN-1k   && 51.0   &   45.6  && - \\ %
        ConvNeXt v2-B~\cite{woo2023convnext} & FCMAE && 52.9   &  46.6      && -\\ %
        \rowcolor{gray!20}
        \davitb~\cite{ding2022davit}  & \flmodel && 53.6   &  46.4   && 59.2 \\
        \bottomrule
        \end{tabular}

    \caption{\textbf{COCO object detection and instance segmentation results} using Mask-RCNN framework, and \textbf{COCO object detection results} using DINO-4scale framework. All the entries use a base size model to ensure a fair comparison. For Mask-RCNN experiments, our method utilizes 1$\times$ schedule (12 epochs), ViT-B use 100 epochs, all others use 3$\times$ (36 epochs). For DINO experiments, all the entries use 1$\times$ schedule except for ViT-B which uses 50 epochs.}
    \label{tab:od_sota_comparison}
\end{table}

\begin{table}[t]
\centering
\setlength{\tabcolsep}{2.5pt}
\small

\begin{tabular}{lc|cccccccccc}
\toprule
\multirow{2}{*}{Pretrain} & \multirow{2}{*}{Frozen stages} && \multicolumn{2}{c}{Mask R-CNN} && DINO && UperNet \\
\cline{4-5} \cline{7-7} \cline{9-9}
  &  && AP$_b$& AP$_m$ && AP   && mIoU\\
\midrule
Sup IN1k & n/a && 46.7 & 42.0 &&  53.7 && 49\\ %
UniCL~\cite{yang2022unified} & n/a && 50.4 & 45.0 &&  57.3 && 53.6\\ %
\rowcolor{gray!20}
\flmodel & n/a && 53.6 & 46.4 &&  59.2 && 54.9 \\ %
\midrule
\flmodel & [1] && 53.6 & 46.3 &&  59.2 && 54.1\\ 
\flmodel & [1, 2] && 53.3 & 46.1 &&  59.0 && 54.4\\ 
\flmodel & [1, 2, 3] && 49.5 & 42.9 &&  56.7 && 49.6\\ 
\flmodel & [1, 2, 3, 4] && 48.3 & 44.5  && 56.1 && 45.9\\ 
\bottomrule
\end{tabular}
\caption{Downstream task fine-tuning on COCO and ADE20K dataset. \textbf{COCO object detection} using Mask R-CNN and DINO. \textbf{ADE20K semantic segmentation} using UperNet. All entries use \davitb with 80M parameters as the backbone and standard 1$\times$ schedule. }
\label{tab:downstram_tasks}
\end{table}

\paragraph{Semantic segmentation.}
We conduct semantic segmentation experiments with UperNet~\cite{xiao2018unified} framework on ADE20k~\cite{zhou2017scene} dataset.
We mostly follow the training and evaluation protocols from Swin~\cite{liu2021swin}. Specifically, we use input size 512$\times$512 and train the model for 40k iterations with a batch size of 64. We adopt the AdamW~\cite{loshchilov2019decoupled} optimizer with the optimal learning rate searched from \{8e-4,4e-4,2e-4,1e-4\}.

Our results show a similar trend to the object detection experiments. As illustrated in~\cref{tab:seg_sota_comparison}, our base model outperforms the previous SoTA model, which is BEiT pre-trained ViT-B~\cite{biao2021beit}, by 1.3 and 1.4 points in single-scale and multi-scale testing protocol, respectively. With the same backbone architecture of DaViT-B~\cite{ding2022davit}, \flmodel pre-trained model achieves a remarkable improvement of 4.9 points and 4$\times$ efficiency compared to the ImageNet-1k pre-trained counterpart as demonstrated in~\cref{tab:downstram_tasks,fig:efficiency_curve}. 

\begin{table}[t]
    \centering
    \setlength{\tabcolsep}{5.5pt}
    \small
        \begin{tabular}{l|c|ccc}
        \toprule
        Backbone  & Pretrain &  mIoU & ms-mIoU \\
        \midrule
        ViT-B~\cite{he2022masked} & Sup IN-1k & 47.4 & - \\  %
        ViT-B~\cite{he2022masked} & MAE IN-1k & 48.1 & - \\
        ViT-B~\cite{biao2021beit} & BEiT & 53.6 & 54.1 \\
        ViT-B~\cite{beitv2} & BEiTv2 IN-1k & 53.1 & - \\
        ViT-B~\cite{beitv2} & BEiTv2 IN-22k & 53.5 & - \\
        Swin-B \cite{liu2021swin}   & Sup IN-1k  &  48.1  &  49.7  \\ %
        Swin-B \cite{liu2021swin}   & Sup IN-22k &   -    &  51.8  \\ %
        Swin-B \cite{liu2021swin}   & SimMIM~\cite{xie2022simmim}   &   -    &  52.8  \\ %
        FocalAtt-B~\cite{yang2021focal}      & Sup IN-1k   &  49.0  &  50.5  \\ %
        FocalNet-B~\cite{yang2022focal}      & Sup IN-1k   &  50.5  &  51.4  \\ %
        ConvNeXt v1-B~\cite{liu2022convnet}  & Sup IN-1k   &   -    &  49.9  \\ %
        ConvNeXt v2-B~\cite{woo2023convnext} & Sup IN-1k   &   -    &  50.5  \\ %
        ConvNeXt v2-B~\cite{woo2023convnext} & FCMAE       &   -    &  52.1  \\ %
        \rowcolor{gray!20}
        \davitb~\cite{ding2022davit}  & \flmodel   &  54.9  &  55.5  \\ %
        \bottomrule
        \end{tabular}
    \caption{\textbf{ADE20K semantic segmentation results} using UperNet. The input size is $512\times512$ for all the entries, except for models with BEiT pre-trained, which use the input size of $640\times640$.}
    \label{tab:seg_sota_comparison}
\end{table}

\subsection{Ablation Studies}

\begin{figure*}[t]
    \centering
    \includegraphics[width=1.0\textwidth]{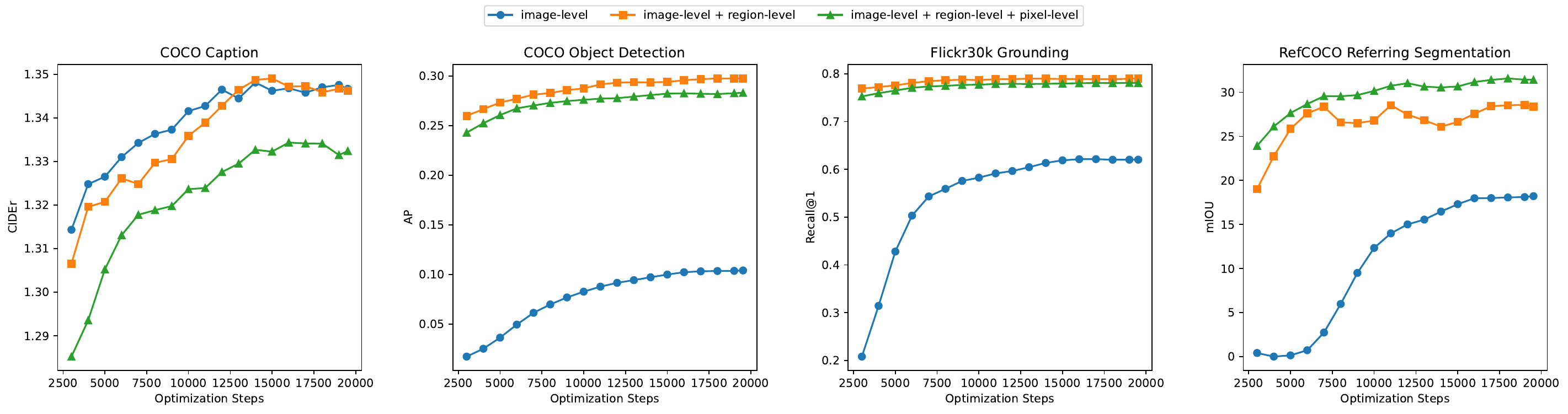}
    \caption{\textbf{Multitask transfer.} We conduct experiments with three different versions of \flmodel models, each trained on a different level of image annotation: image level, image and region level, and image, region, and pixel level. We then evaluate the transfer learning performance of these models on four downstream tasks: COCO caption, COCO object detection, Flickr30k grounding, and Refcoco referring segmentation.}
    \label{fig:multi_task_transfer}
\end{figure*}

\paragraph{Multitask transfer.}
In this study, we aimed to identify the most effective pre-trained model for transfer learning across various downstream tasks in computer vision. We compared three different models, each pre-trained on a different combination of tasks:
\begin{itemize}
    \item Image-level Model: pre-trained on image-level tasks only
    \item Image-Region Model: pre-trained on image-level and region-level tasks
    \item Image-Region-Pixel Model: pre-trained on image-level, region-level, and pixel-level tasks
\end{itemize}

For pre-training, we optimize all models for the same number of effective samples (72M) on a subset of our \fld dataset.

These models are then transferred to a combined dataset with four downstream tasks, each representing a different level of task granularity: COCO caption (image-level task), COCO object detection (region-level task), Flickr30k grounding (region-level task), RefCOCO referring segmentation (pixel-level task). 

The results are shown in ~\cref{fig:multi_task_transfer}. The results demonstrate that Image-Region-Pixel Model, pre-trained on all three levels of tasks, consistently demonstrated competitive performance across the four downstream tasks.

For the COCO caption task, Image-Region-Pixel Model initially performs worse than Image-level Model and Image-Region Model but eventually achieve a final performance (133.4 CIDEr) that is only slightly worse than the other models (134.6 CIDEr).

For the COCO object detection task, Image-Region-Pixel Model outperforms Image-level Model by a significant margin (28.3 vs. 0.1) and was only slightly worse than Image-Region Model (29.7).

For the Flickr30k grounding task, Image-Region-Pixel Model shows strong performance (78.1 recall@1), comparable to Image-Region Model (79.1 recall@1) and significantly better than Image-level Model (62.0 recall@1).

For the RefCOCO referring segmentation task, Image-Region-Pixel Model clearly outperforms both Image-level Model and Image-Region Model, achieving the highest performance (31.6 mIoU) compared to the other models (28.4 and 18.2 mIoU).

Our findings suggest that the Image-Region-Pixel Model, which is pre-trained on tasks at the image, region, and pixel levels, is the most effective base model for transfer learning across various computer vision tasks. This model shows strong performance on all four downstream tasks we evaluated, and consistently outperforms the Image-level Model and matches or exceeds the Image-Region Model in performance. By pre-training a model on tasks at different levels of granularity, we can ensure that the base model is better prepared to handle a diverse range of downstream tasks, offering a versatile and robust solution for transfer learning in computer vision.

\paragraph{Model scaling.}
We aimed to investigate the impact of increasing model capacity on zero-shot performance on various downstream tasks in computer vision. We compared two models:
\flmodelb and \flmodell, which have 232M and 771M parameters, respectively. The model architectures are described in~\cref{table:model_sizes}.
We show the zero-shot performance on four downstream tasks in~\cref{tab:ablation_model_scaling}. The large model clearly outperforms the base model across various downstream tasks.

\begin{table}[t]
\centering
\setlength{\tabcolsep}{3.3pt}
\small

\begin{tabular}{l|ccccccccc}
\toprule
\multirow{2}{*}{Model}&& Caption && Detection && Grounding && \multicolumn{2}{c}{RES} \\
\cline{3-3} \cline{5-5} \cline{7-7} \cline{9-10}
&&  CIDEr && AP && Recall@1 && mIOU & oIOU \\
\midrule
Base &&  118.7 && 19.7 && 76.3 && 18.6  & 17.8 \\ 
Large && \textbf{124.4} && \textbf{22.6} && \textbf{78.2} &&   \textbf{21.5} & \textbf{19.1} \\ 
\bottomrule
\end{tabular}
\caption{\textbf{Model scaling.} Zero-shot performance on COCO caption and COCO object detection, Flickr30k grounding, RefCOCO referring expression segmentation(RES).}
\label{tab:ablation_model_scaling}
\end{table}

\paragraph{Data scaling.}
We conducted experiments to study how zero-shot performance on various computer vision tasks is affected by the scale of pre-training data. 
We used four different data sizes for pre-training: 0.12M, 0.36M, 1.2M, and 12M images. All models were trained with the same effective sample size (72M) on a subset of \fld data.

\begin{table}[t]
\centering
\setlength{\tabcolsep}{3.1pt}
\small

\begin{tabular}{r|ccccccccc}
\toprule
Data&& Caption && Detection && Grounding && \multicolumn{2}{c}{RES} \\
\cline{3-3} \cline{5-5} \cline{7-7} \cline{9-10}
size && CIDEr && AP && Recall@1 && mIOU & oIOU \\
\midrule
0.12M &&  102.8 && 16.1 && 74.0 && 15.9 & 16.6 \\ 
0.36M &&  114.3 && 18.7 && 75.8 && 16.6 & 16.4 \\ 
1.2M &&   118.1 && 18.9 && 76.3 && \textbf{19.3} & \textbf{18.4} \\ 
12M &&   \textbf{118.7} && \textbf{19.7} && \textbf{76.3} && 18.6 & 17.8 \\ 
\bottomrule
\end{tabular}

\caption{\textbf{Data scaling.} Zero-shot performance on COCO caption, COCO object detection, Flickr30k grounding, COCORef referring segmentation.}
\label{tab:ablation_data_scaling}
\end{table}

~\Cref{tab:ablation_data_scaling} presents the zero-shot performance results on COCO caption, COCO object detection, Flickr30k grounding, and RefCoco referring segmentation (RES) tasks. 
We can observe a trend of improved zero-shot performance on the downstream tasks as the pre-training data size increases (except for RES, 1.2M data has slightly better performance compared to 12M).

Our experiments on data scaling demonstrate that larger pre-training data sizes generally lead to improved zero-shot performance across a variety of downstream tasks in computer vision. This finding suggests that investing in larger pre-training datasets can provide a more effective and versatile foundation for handling a wide range of downstream tasks. 

Our approach to scaling data is significantly more efficient than relying solely on human annotations, as most of the annotation generation is performed using model inference. By leveraging specialist models to generate annotations, we can substantially reduce the time and cost associated with manual annotation efforts, which often involve labor-intensive processes and may be subject to human errors or inconsistencies.

Furthermore, utilizing model-generated annotations enables us to scale the pre-training datasets more rapidly and efficiently, allowing us to explore the impact of larger data sizes on model performance across various downstream tasks in computer vision. This not only facilitates the development of more effective and versatile foundation models but also ensures that the annotation process remains sustainable and scalable as the need for high-quality labeled data continues to grow.

In summary, our data scaling approach offers a more efficient alternative to traditional human annotation methods by harnessing the power of specialist models for annotation generation. This strategy enables us to accelerate the pre-training process, optimize model performance, and effectively manage the ever-increasing demand for labeled data in the field of computer vision.

\paragraph{Training settings.}
We analyze the basic model training settings for the two primary components of our model, namely the vision encoder and the multi-modality encoder-decoder. The experiment results are presented in ~\cref{tab:ablation_basic_components}

\begin{table}[t]
\centering
\setlength{\tabcolsep}{2.1pt}
\small

\begin{tabular}{cc|cccccccccc}
\toprule
& && Caption && Detection && Grounding && \multicolumn{2}{c}{RES} \\
\cline{4-4} \cline{6-6} \cline{8-8} \cline{10-11}
V Pre & L Pre &&  CIDEr && AP && Recall@1 && mIOU & oIOU  \\
\midrule
\multicolumn{11}{c}{\textbf{\textit{Freeze Vision Encoder}}} \\
\midrule
\checkmark & \checkmark  &&  120.0 && 6.9 && 66.3 && 9.9 & 13.6 \\ 
\midrule
\multicolumn{11}{c}{\textbf{\textit{Unfreeze Vision Encoder}}} \\
\midrule
 & \checkmark  &&  81.3 && 4.9 && 69.0 && 15.3  & 15.6 \\ 
\checkmark &  &&  117.4  && 19.6 && 75.2 && \textbf{21.5} & \textbf{19.3}  \\ 
\checkmark & \checkmark  &&  \textbf{118.7} && \textbf{19.7} && \textbf{76.3} && 18.6 & 17.8 \\ 
\bottomrule
\end{tabular}
\caption{\textbf{Basic components.} Zero-shot performance on COCO caption, COCO object detection, Flickr30k grounding, and COCORef referring segmentation. V Pre and L Pre indicate that using vision and language pre-training initialization, respectively.}
\label{tab:ablation_basic_components}
\end{table}

We observe that freezing the vision encoders does not affect the performance on tasks that require image-level understanding, but it significantly degrades the performance on tasks that require region-level or pixel-level understanding (e.g., AP on COCO object detection drops from 19.7 to 6.9). Previous methods for pre-training vision foundation models mainly focus on image-level tasks (e.g., image classification~\cite{krizhevsky2012imagenet,he2016deep}, image-text contrastive learning~\cite{radford2021learning,yuan2021florence}), which may not provide them with sufficient region-level and pixel-level skills for downstream tasks. Therefore, it is important to unfreeze the vision backbone, enabling it to learn region-level and pixel-level features for various downstream tasks.

The effect of language pre-training weights on multi-modal encoder-decoder tasks varies depending on the task. Tasks that require more text understanding, such as captioning and grounding, benefit slightly from using language pre-training weights (e.g., COCO caption, Flickr30k grounding). Tasks that are mostly vision-focused, such as object detection and region segmentation, do not gain much from using language pre-training weights (for COCO object detection, the gain is only 0.1; for RES tasks, which use only localization tokens, the drop is 2.91 mIOU).

We investigate the effects of different training configurations on the performance of a foundation model in region-level and pixel-level tasks. We find that unfreezing the vision backbone is crucial for enhancing the model's ability to learn from regions and pixels, which is beneficial for transferring to various downstream tasks. Moreover, we observe that using language pre-training weights can help the model in tasks that require text understanding, but have less impact on tasks that are purely vision-based. These results offer useful guidance for choosing the best training settings for different computer vision tasks.

\section{Related Works}
\label{sec:related_works}

\subsection{Vision-Language Foundation Models}

Recent vision-language pre-training models~\cite{radford2021learning, jia2021scaling, yuan2021florence} have demonstrated impressive zero-shot transfer abilities to vision-language alignment and image classification tasks, thanks to the alignment of vision and text embeddings extracted from respective encoders through contrastive learning objectives~\cite{sohn2016improved, oord2018representation}. These models (\eg,~\cite{yuan2021florence}), trained on weakly large-scale image-text data, have been further extended to more downstream tasks such as object detection, achieving state-of-the-art performance with task-specific adaptation heads.

In contrast, other studies~\cite{yu2022coca, li2022blip, wang2022git, alayrac2022flamingo} propose using a multi-modality decoder to predict text in an autoregressive manner with language modeling pre-training objectives. Techniques for fusing vision and language embeddings vary: GIT~\cite{wang2022git} concatenates vision and text tokens as decoder input and designs a casual attention mask, CoCa~\cite{yu2022coca} uses attentional poolers with learnable queries to select task-specific vision representations which are then cross-attended via the decoder, and Flamingo~\cite{alayrac2022flamingo} pools a fixed number of vision tokens with a Perceiver Resampler and adds new learnable cross-attention layers to the decoder while freezing the pre-trained vision encoder and text decoder.

Beyond image captioning pre-training task, some research~\cite{lu2022unifiedio, chen2022pali, wang2022ofa} attempts to formulate more vision tasks in a unified sequence-to-sequence learning paradigm, including object detection and image segmentation. Customized special tokens accommodate representations beyond pure text, such as bounding boxes~\cite{lu2022unifiedio, chen2022pix2seq, wang2022ofa}. This approach uses the same architecture for pre-training and downstream tasks, potentially using the same set of weights for all tasks. Our method, which falls into this category, aims to obtain foundation models that understand dense information beyond simple image-level captions. It shares the same encoder-decoder design as other multi-modality encoder-decoder models~\cite{chen2022pali, lu2022unifiedio} adapted for sequence-to-sequence learning, but uses our built large-scale comprehensive annotation data instead of combining existing sparse annotated data.

\subsection{Vision Datasets}

\paragraph{Comprehensive annotations.} The quest for comprehensive understanding of visual scenes, the holy grail of computer vision~\cite{krishna2017visual}, has evolved from focusing on individual datasets each targeting a single perspective, \eg, image classification~\cite{deng2009imagenet}, to providing multi-perspective~\cite{lin2014microsoft,krishna2017visual,kuznetsova2020open}, comprehensive annotations for every visual data point. Notable datasets like MS-COCO~\cite{lin2014microsoft,chen2015microsoft} and Visual Genome~\cite{krishna2017visual} integrate various types of annotations, enabling richer understanding in spatial and semantic granularities and better model interactions across annotations. However, due to the high cost of human verification, these annotations are limited in size. Our datasets, while large-scale, maintain comprehensive annotations covering text, region-text pairs, and text-phrase-region triplets, with reduced human involvement.

\paragraph{Scalable annotations.}: Over the past decade, vision datasets have rapidly scaled up from thousands \cite{lecun2010mnist,krizhevsky2009learning} to billion examples \cite{jia2021scaling,zhai2022scaling} to encompass more visual concepts for better generalization. This shift is evident in recent foundation models that employ massive quantities of data~\cite{bommasani2021opportunities}. These large datasets typically collect images from the web and parse noisy annotations from the corresponding metadata, such as category label from query~\cite{sun2017revisiting,zhai2022scaling}, short description from alt-text~\cite{radford2021learning,jia2021scaling}, as well as detailed description from interleaved text~\cite{alayrac2022flamingo,laurenccon2023obelisc}. Despite their diversity, these annotations suffer from randomness and limited types (\ie, texts only). Some works~\cite{kirillov2023segment,li2022blip} attempt to scale up annotations using pseudo-label generation with iteratively trained models, which offer higher quality without significant diversity loss. Our data pipeline extends these large-scale, web-crawled noisy annotations with higher-quality, autonomous annotations generated from multiple specialist models. The pipeline iteratively refines labels and completes missing pieces, resulting in a scalable and comprehensive dataset for learning a unified visual representation.

\section{Conclusion}
The Florence Project endeavors to develop a foundational vision model endowed with a diverse array of perceptual capabilities, encompassing spatial hierarchy and semantic granularity. To this end, we construct \fld dataset containing an extensive collection of 126M images paired with 5B comprehensive annotations, which are collected by the Florence data engine. Subsequently, we pre-train \flmodel on this rich dataset through comprehensive multitask learning in a unified manner. \flmodel has exhibited remarkable zero-shot capabilities that extend across a wide spectrum of visual tasks, such as captioning, object detection, visual grounding, and referring segmentation, among others. The experimental findings underscore the potency of the universal representation pre-trained by \flmodel, revealing its substantial contributions to the enhancement of a multitude of downstream tasks.

\vspace{1.0em}
\paragraph{Acknowledgment.}
We would like to express our heartfelt gratitude to all the contributors from the Azure AI team who worked on the Florence project. We sincerely appreciate Misha Bilenko for the invaluable guidance and support. Our thanks are extended to Yi-Ling Chen, Mengchen Liu, Yen-Chun Chen and Dongdong Chen for engaging in helpful discussions and to Yunsheng Li for their assistance with segmentation annotations. Deep appreciation is also expressed to Qingfen Lin, Ryan Menezes, Kuan Lu, Gabe Blanco, Shohei Ono, Ping Jin, Jiahe Zhou, Xiong Qiao, Tong Bai, Xingchao Peng, Pei Guo, Lihang Li for providing valuable feedback in downstream applications discussions. Special thanks to Cha Zhang, Jinyu Li, Min Gao, Christina Sun, Oliver Ernst, Kevin Pan, Mei Gao for their work on data annotation support and insightful discussions in data pipeline. Furthermore, we would like to thank Thomas Soemo, Nguyen Bach for their constructive feedback.

{\small
\bibliographystyle{ieee_fullname}
\bibliography{main}
}

\clearpage
\onecolumn
\appendix    
\section{Supported Tasks and Annotations in Florence-2}

\begin{table*}[!htbp]
\centering
\setlength{\tabcolsep}{18pt}

\begin{tabular}{l|l|l|l}
\toprule
Task & Annotation Type & Prompt Input & Output \\ \midrule
Caption & Text & Image, text & Text  \\ %
Detailed caption & Text & Image, text & Text \\ %
More detailed caption & Text & Image, text & Text \\ %
Region proposal & Region & Image, text & Region\\ %
Object detection & Region-Text & Image, text & Text, region \\ %
Dense region caption & Region-Text & Image, text & Text, region\\ %
Phrase grounding & Text-Phrase-Region & Image, text & Text, region \\ %
Referring expression comprehension & Region-Text & Image, text & Text, region  \\ %
Open vocabulary detection & Region-Text & Image, text & Text, region \\ %
Referring segmentation & Region-Text & Image, text & Text, region \\ %
Region to text & Region-Text & Image, text, region & Text\\ %
Text detection and recognition & Region-Text & Image, text & Text, region \\ \bottomrule
\end{tabular}
\caption{Supported Tasks and annotations used for \flmodel pretraining.}
\label{tab:tasks}

\end{table*}

\section{Supervised Data Collection for Generalist Model Fine-tuning}
\begin{table*}[!htbp]
    \centering
    \setlength{\tabcolsep}{20pt}%
    \small
    \begin{tabular}{l|l}
    \toprule
    \label{table:datasets-by-task}
    Task   & Dataset                                          \\
    \midrule
    Caption                           & COCO~\cite{chen2015microsoft} \\
    Text Caption                      & TextCaps~\cite{sidorov2020textcaps}                                         \\
    Paragraph caption                 & Standford Paragraph Caption~\cite{krause2016paragraphs}\\
    Detailed caption                  & Localized Narratives~\cite{PontTuset_eccv2020}                      \\
    Detection                         & COCO~\cite{lin2015microsoft}, Object365$^{*}$~\cite{shao2019objects365}, Open Images$^{*}$~\cite{Kuznetsova_2020} \\
    Phrase Grounding                  & Flickr30k, Object365$^{*}$~\cite{shao2019objects365}, Open Images$^{*}$~\cite{Kuznetsova_2020} \\
    Referring expression              & RefCOCO-mix (RefCOCO, RefCOCO+, RefCOCOg)~\cite{kazemzadeh2014referitgame, yu2016modeling, mao2016generation}                      \\
    Referring expression segmentation & RefCOCO-mix (RefCOCO, RefCOCO+, RefCOCOg)~\cite{kazemzadeh2014referitgame, yu2016modeling, mao2016generation} \\
    Region to category                & COCO~\cite{lin2015microsoft}, Object365$^{*}$~\cite{shao2019objects365}, Open Images$^{*}$~\cite{Kuznetsova_2020}                 \\
    Region to polygon                 &  COCO~\cite{lin2015microsoft} (after deduplicating RefCOCO-mix val) \\
    VQA                               & VQAv2~\cite{balanced_vqa_v2}, OKVQA~\cite{marino2019okvqa}, AOKVQA~\cite{schwenk2022aokvqa}, TextVQA~\cite{singh2019towards}, ViZWiz VQA~\cite{gurari2018vizwiz}        \\
    OCR       & Subset from \fld OCR (2 millon samples)\\
    \bottomrule
    \end{tabular}
    \caption{Collection of dataset for finetuning one single generalist model for downstream tasks evaluation. $^{*}$ indicates using the annotations from \fld, which merges original annotations with ours.}
\end{table*}

\section{Model Configuration}

\begin{table*}[!htbp]
\centering
\setlength{\tabcolsep}{3.7pt}
\small
\begin{tabular}{c|cccc|cccc}
\toprule
\multirow{2}{*}{Model} & \multicolumn{4}{c|}{Image Encoder (DaViT)} & \multicolumn{4}{c}{Encoder-Decoder (Transformer)} \\
& dimensions & blocks & heads/groups & \#params & encoder layers & decoder layers & dimensions & \#params\\
\midrule
\flmodelb  & [128, 256, 512, 1024]   & [1, 1, 9, 1] & [4, 8, 16, 32]  &       90M   & 6 &  6 & 768 &  140M \\
\flmodell & [256, 512, 1024, 2048] & [1, 1, 9, 1] & [8, 16, 32, 64] &         360M & 12   & 12 & 1024     &  410M \\
\bottomrule
\end{tabular}

\caption{Model configuration of different size.}
\label{table:model_sizes}
\end{table*}

\section{More Examples of Annotations in \fld}
\vspace{-0.8em}
\begin{minipage}{\linewidth}
    \centering
    \includegraphics[height=0.97\textheight]{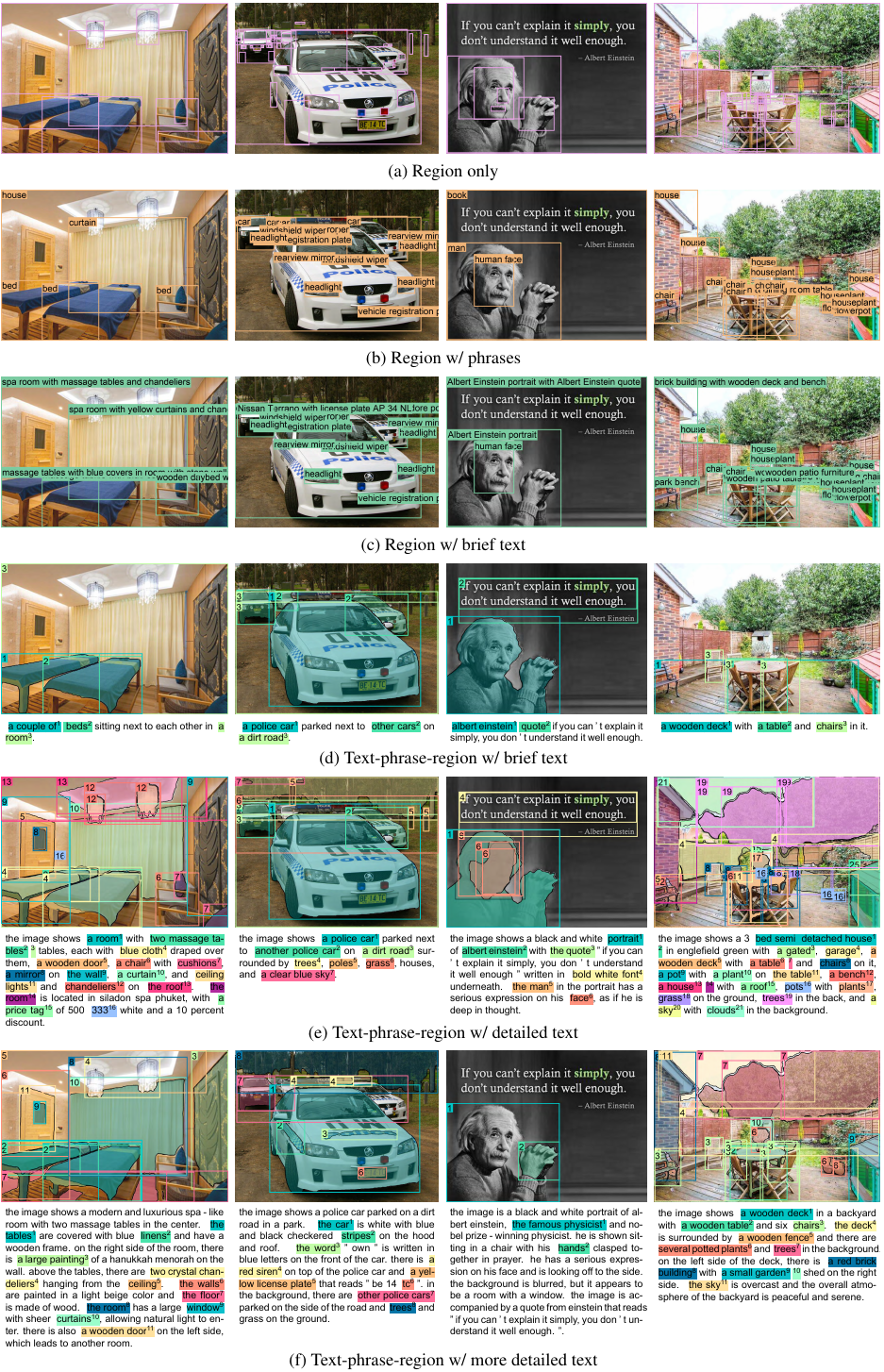}
    \vspace{-1em}
     \captionof{figure}{Examples of annotations in \fld.}
\end{minipage}

\newpage
\begin{minipage}{\linewidth}
    \centering
    \includegraphics[height=0.96\textheight]{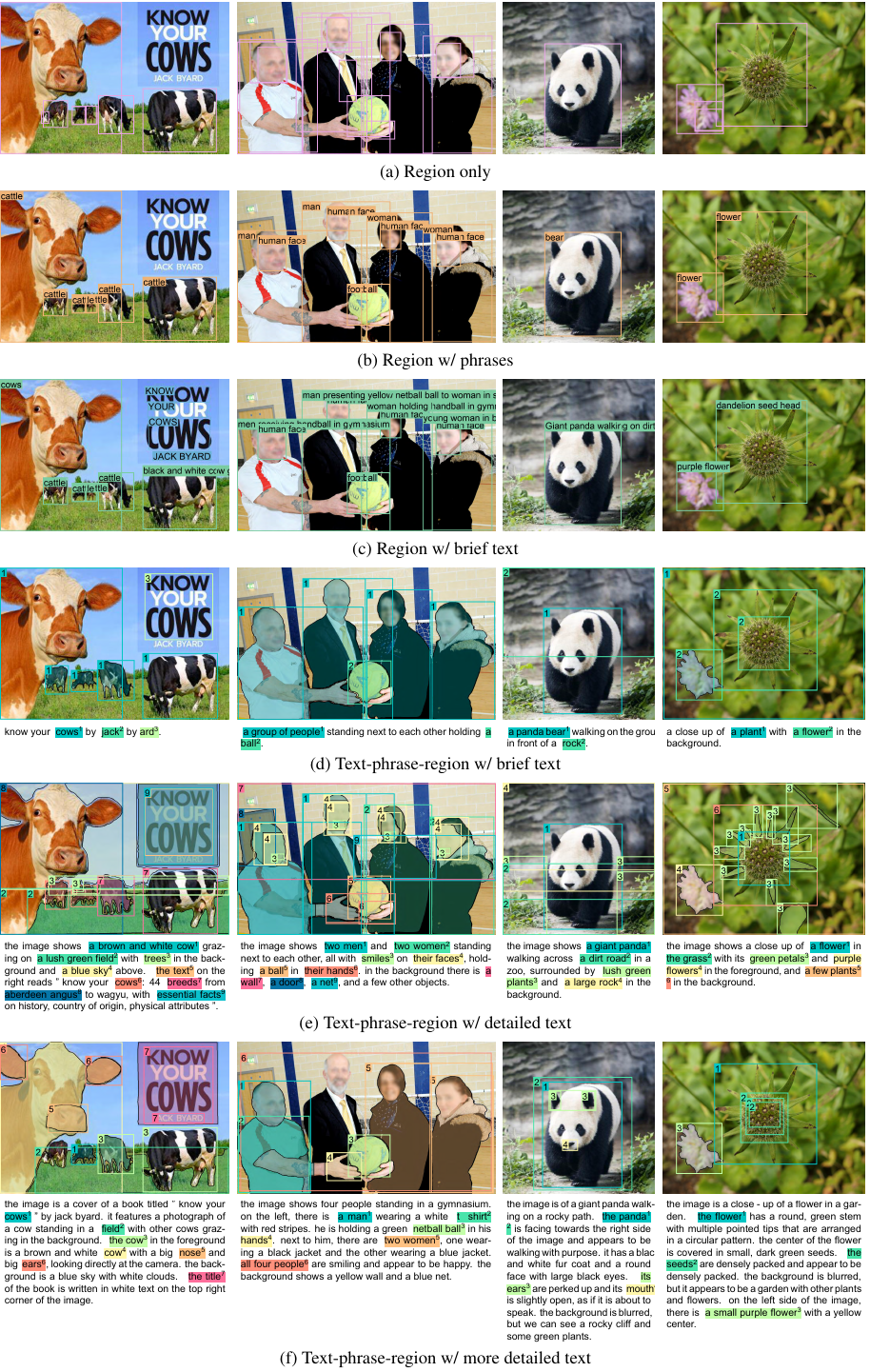}
    \captionof{figure}{Examples of annotations in \fld (continued).}
\end{minipage}

\section{Qualitative Evaluation and Visualization Results}

\subsection{Detailed Image Caption}
\begin{minipage}{1.0\linewidth}
\centering
\begin{AIbox}{Detailed Image Caption}

{\bf Prompt}: Describe with a paragraph what is shown in the image. 

\parbox[h]{0.45\textwidth}{
    \centering
    \includegraphics[width=0.25\textwidth]{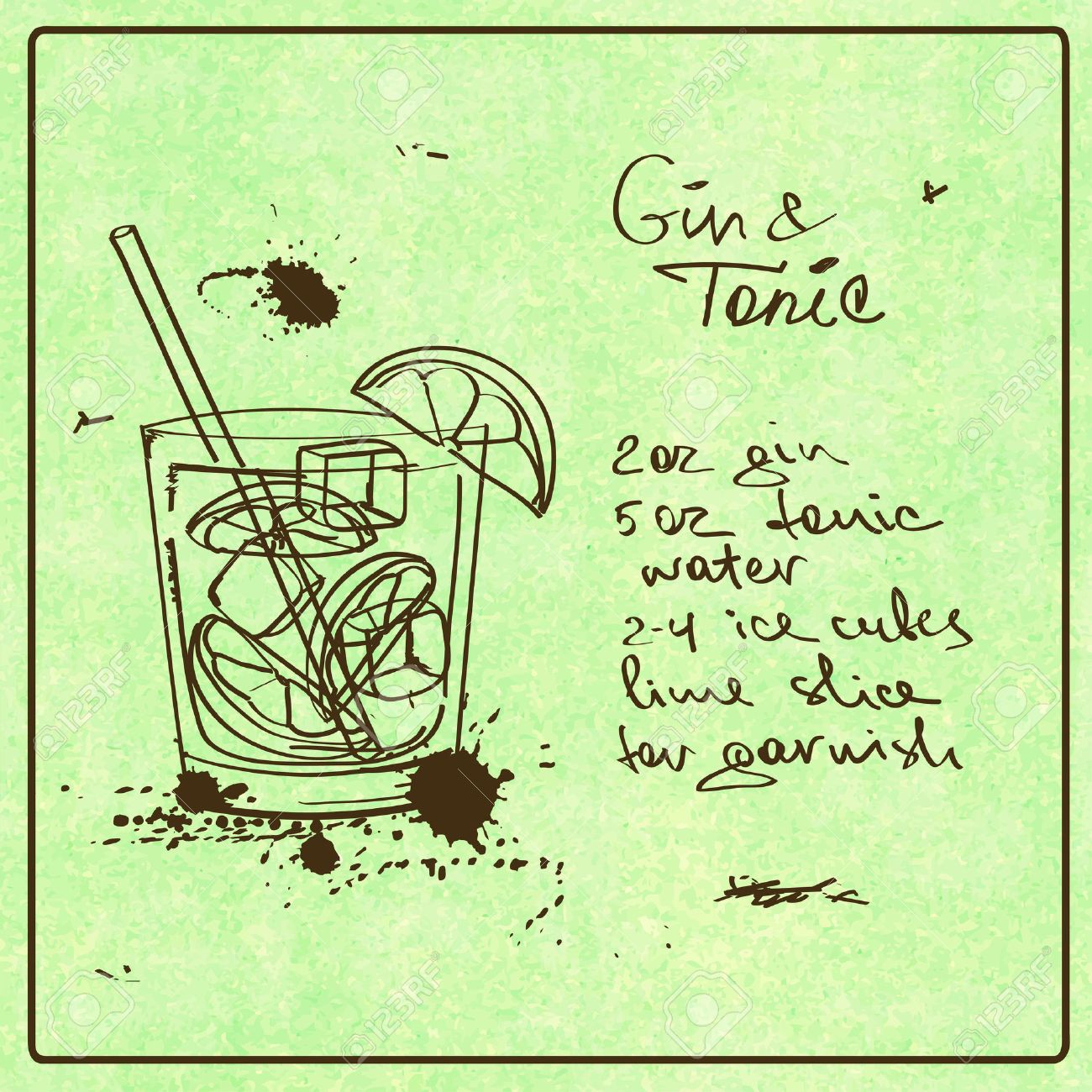}
}
\hspace{10px}
\parbox[h]{0.50\textwidth}{
\scriptsize 
\begin{alltt}
The image is a hand-drawn illustration of a glass of gin and tonic. The glass is filled with ice cubes, lime slices, and a slice of lemon. There is a straw sticking out of the top of the glass. The background is a light green color with a grungy texture. On the right side of the image, there is a list of ingredients written in black ink. The text reads "Gin \& Tonic" and "2oz gin, 5oz tonic water, 2-4 ice cubes lime slice for garnish".
\end{alltt}
}

\tcbline
\parbox[h]{0.45\textwidth}{
    \centering
    \includegraphics[width=0.25\textwidth]{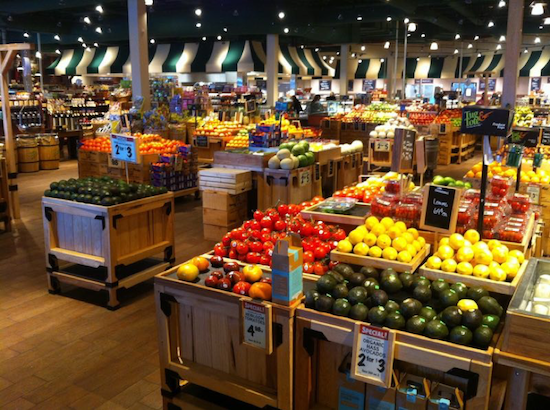}
}
\hspace{10px}
\parbox[h]{0.50\textwidth}{
\scriptsize \begin{alltt}
The image shows the interior of a grocery store with a large variety of fruits and vegetables on display. The store has a high ceiling with green and white striped awnings, and the floor is made of wood. There are rows of wooden crates filled with different types of fruits, including apples, oranges, lemons, limes, and limes. The crates are arranged in neat rows, and there are price tags attached to each crate. In the background, there are shelves stocked with various items such as fruits, vegetables, and other grocery items. The overall atmosphere of the store is bright and colorful, with a sense of freshness and abundance.
\end{alltt}
}
\tcbline

\parbox[h]{0.45\textwidth}{
    \centering
    \includegraphics[width=0.25\textwidth]{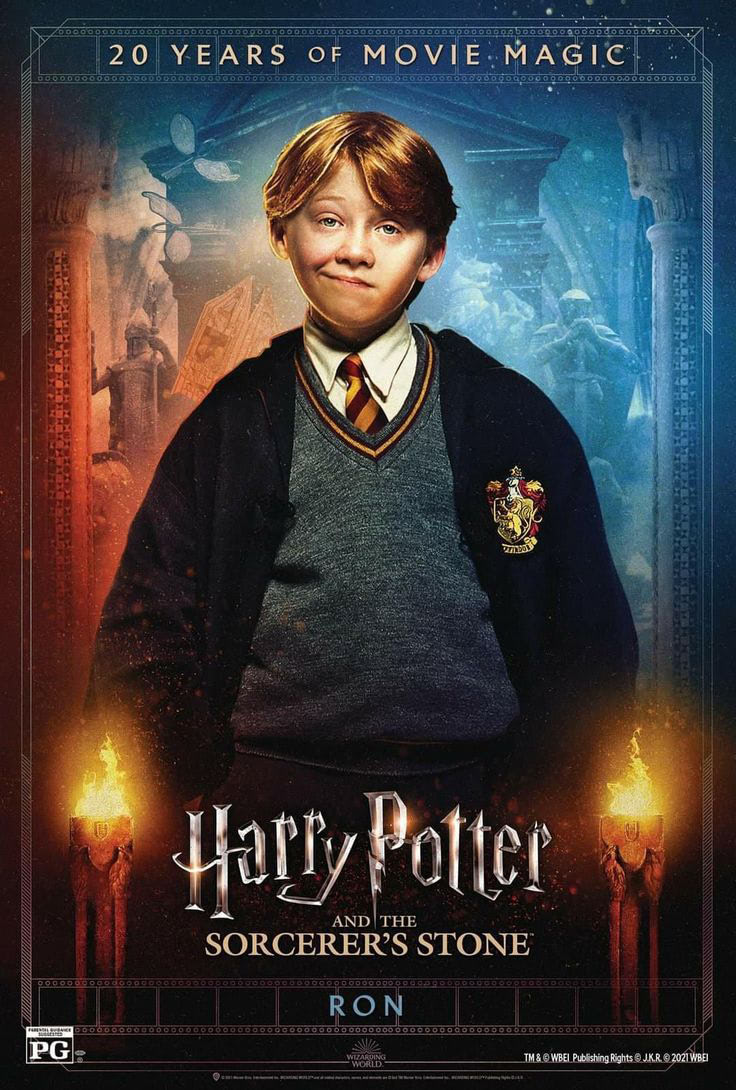}
}
\hspace{10px}
\parbox[h]{0.50\textwidth}{
\scriptsize \begin{alltt}
The image is a movie poster for the film "Harry Potter and the Sorcerer's Stone" starring Ron Weasley. The poster features a young boy, Ron Weasley, standing in front of a dark background with a castle-like structure in the background. He is wearing a school uniform with a Gryffindor crest on his jacket and a red tie. He has blonde hair and is looking directly at the camera with a serious expression on his face. The title of the film is written in white text at the top of the poster, with the tagline "20 years of movie magic" written in smaller text below.
\end{alltt}
}

\tcbline
\parbox[h]{0.45\textwidth}{
    \centering
    \includegraphics[width=0.25\textwidth]{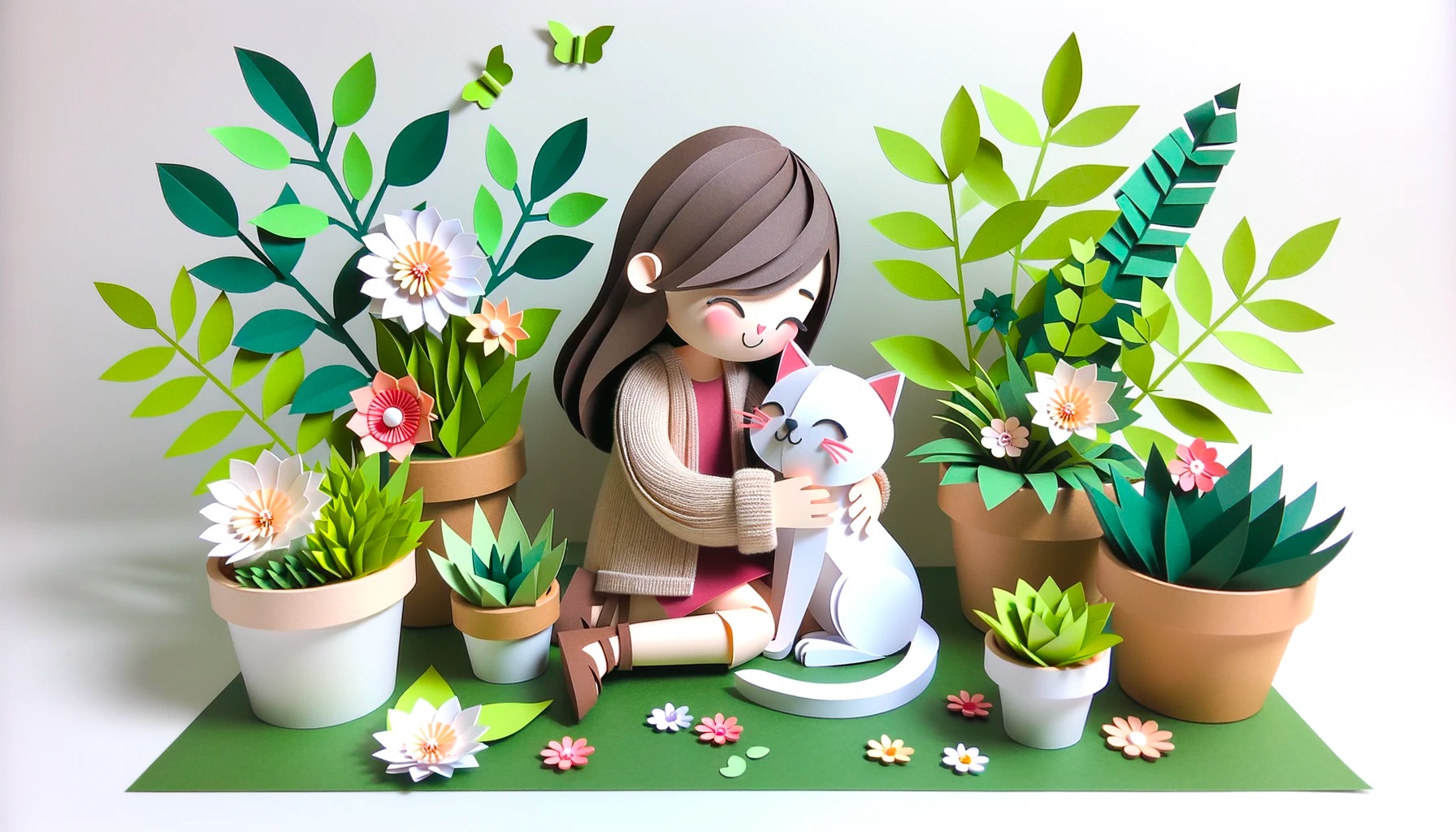}
}
\hspace{10px}
\parbox[h]{0.50\textwidth}{
\scriptsize \begin{alltt}
The image is a digital illustration of a girl hugging a white cat. The girl is wearing a pink sweater and has long brown hair. She is sitting on a green surface with several potted plants and flowers around her. The plants have green leaves and pink and white flowers. There are also two butterflies fluttering around the scene. The background is white. The overall style of the illustration is cartoon-like and playful.
\end{alltt}
}
\end{AIbox}
\captionof{figure}{Detailed captioning prediction results.}
\end{minipage}

\subsection{Visual Grounding}
\begin{minipage}{1.0\linewidth}
\begin{AIbox}{Visual Grounding}
{\bf Prompt}: Locate the phrases in the caption: \{caption\}

\parbox[h]{0.5\textwidth}{
    \scriptsize \begin{alltt}
    \centering
        \includegraphics[width=0.45\textwidth]{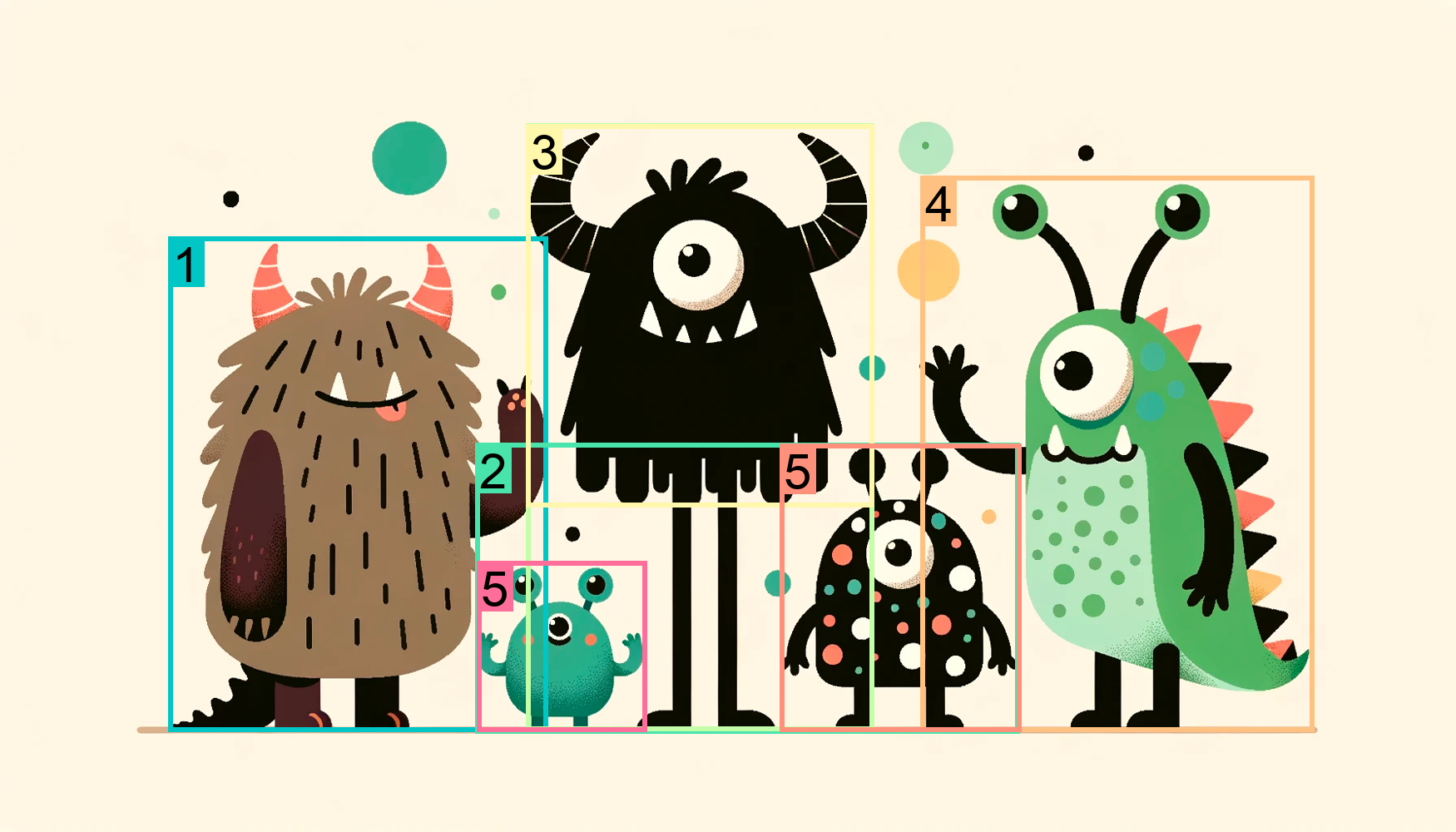}
    \end{alltt}
}
\hspace{10px}
\parbox[h]{0.50\textwidth}{ 
    \scriptsize \begin{alltt}
              \noindent The image shows a group of five cartoon monsters. On the left side, there is \definecolor{customcolor}{rgb}{0.0,0.7764705882352941,0.7725490196078432}{\sethlcolor{customcolor}\textcolor{black}{\hl{a brown monster\textsuperscript{1}}}} with horns and a big smile on its face. Next to it, there are two \definecolor{customcolor}{rgb}{0.28627450980392155,0.8941176470588236,0.6705882352941176}{\sethlcolor{customcolor}\textcolor{black}{\hl{smaller monsters\textsuperscript{2}}}}, one black and one green. \definecolor{customcolor}{rgb}{0.7647058823529411,0.996078431372549,0.6588235294117647}{\sethlcolor{customcolor}\textcolor{black}{\hl{The black monster\textsuperscript{3}}}} has two large horns on its head and is standing in the center of the group. \definecolor{customcolor}{rgb}{1.0,0.7568627450980392,0.5098039215686274}{\sethlcolor{customcolor}\textcolor{black}{\hl{The green monster\textsuperscript{4}}}} on the right side is a green monster with big eyes and a long antennae. It is standing on its hind legs with its arms stretched out to the sides. In the middle of the image, there appears to be \definecolor{customcolor}{rgb}{1.0,0.5686274509803921,0.48627450980392156}{\sethlcolor{customcolor}\textcolor{black}{\hl{a small blue monster\textsuperscript{5}}}} with a round head and two antennae on its back. The background is light beige with small green circles scattered around.
\end{alltt}
}
\tcbline
\parbox[h]{0.5\textwidth}{
    \scriptsize \begin{alltt}
        \centering
        \includegraphics[width=0.42\textwidth]{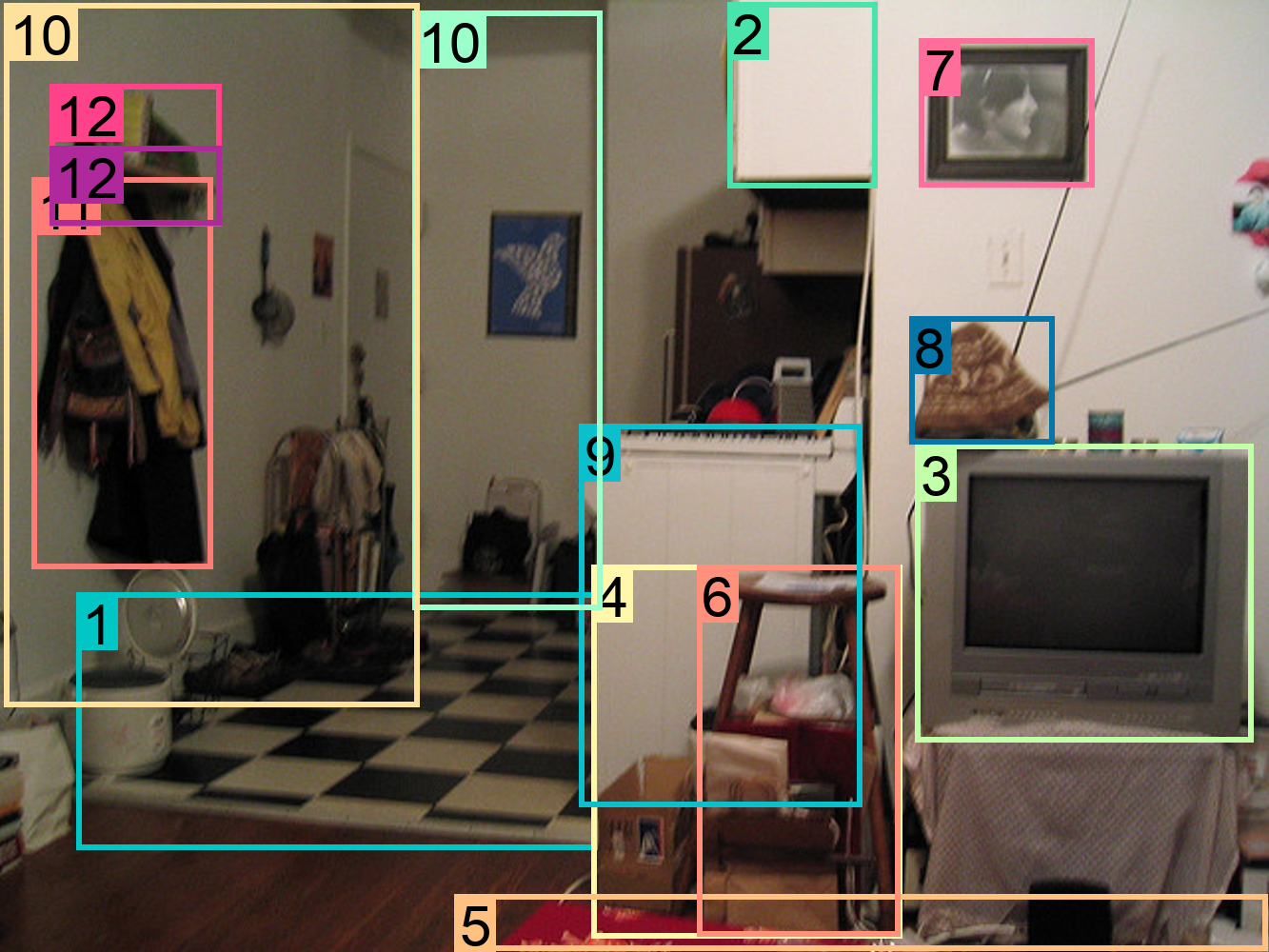}
    \end{alltt}
}
\hspace{10px}
\parbox[h]{0.50\textwidth}{ 
    \scriptsize \begin{alltt}
                \noindent The image shows a cluttered room with a black and white checkered \definecolor{customcolor}{rgb}{0.0,0.7764705882352941,0.7725490196078432}{\sethlcolor{customcolor}\textcolor{black}{\hl{floor\textsuperscript{1}}}}. On the right side of the image, there is \definecolor{customcolor}{rgb}{0.28627450980392155,0.8941176470588236,0.6705882352941176}{\sethlcolor{customcolor}\textcolor{black}{\hl{a small white cabinet\textsuperscript{2}}}} with a \definecolor{customcolor}{rgb}{0.7647058823529411,0.996078431372549,0.6588235294117647}{\sethlcolor{customcolor}\textcolor{black}{\hl{television\textsuperscript{3}}}} on top of it. Next to the cabinet, there are \definecolor{customcolor}{rgb}{1.0,0.9686274509803922,0.6705882352941176}{\sethlcolor{customcolor}\textcolor{black}{\hl{several items\textsuperscript{4}}}} scattered on the floor, including a red \definecolor{customcolor}{rgb}{1.0,0.7568627450980392,0.5098039215686274}{\sethlcolor{customcolor}\textcolor{black}{\hl{blanket\textsuperscript{5}}}}, \definecolor{customcolor}{rgb}{1.0,0.5686274509803921,0.48627450980392156}{\sethlcolor{customcolor}\textcolor{black}{\hl{a wooden stool\textsuperscript{6}}}}, and a pile of trash. On top of the cabinet is \definecolor{customcolor}{rgb}{1.0,0.43137254901960786,0.6078431372549019}{\sethlcolor{customcolor}\textcolor{black}{\hl{a picture frame\textsuperscript{7}}}} and a \definecolor{customcolor}{rgb}{0.023529411764705882,0.4627450980392157,0.6588235294117647}{\sethlcolor{customcolor}\textcolor{black}{\hl{hat\textsuperscript{8}}}}. In the center of the room is \definecolor{customcolor}{rgb}{0.043137254901960784,0.7529411764705882,0.8}{\sethlcolor{customcolor}\textcolor{black}{\hl{a white refrigerator\textsuperscript{9}}}} with a few items on top. \definecolor{customcolor}{rgb}{0.6078431372549019,0.996078431372549,0.796078431372549}{\sethlcolor{customcolor}\textcolor{black}{\hl{The walls\textsuperscript{10}}}} are painted white and there are \definecolor{customcolor}{rgb}{1.0,0.4823529411764706,0.4588235294117647}{\sethlcolor{customcolor}\textcolor{black}{\hl{a few clothes\textsuperscript{11}}}} hanging on a \definecolor{customcolor}{rgb}{1.0,0.25882352941176473,0.5411764705882353}{\sethlcolor{customcolor}\textcolor{black}{\hl{rack\textsuperscript{12}}}} on the left wall. The room appears to be in disarray, with some items strewn about and others scattered around.
\end{alltt}
}
\tcbline
\parbox[h]{0.5\textwidth}{
    \scriptsize \begin{alltt}
    \centering
        \includegraphics[width=0.45\textwidth]{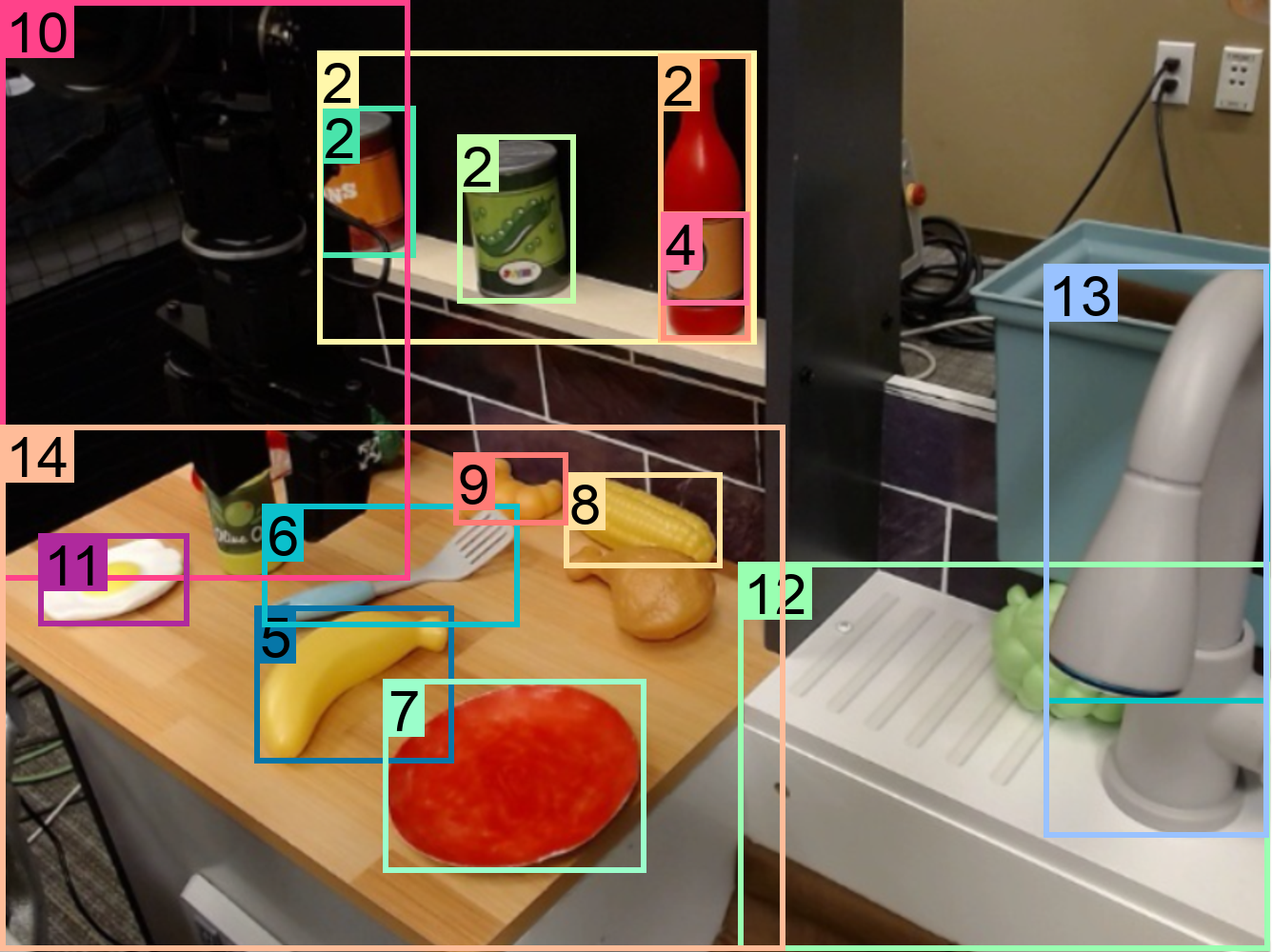}
    \end{alltt}
}
\hspace{10px}
\parbox[h]{0.50\textwidth}{ 
    \scriptsize \begin{alltt}
          \noindent The image shows a kitchen countertop with various kitchen items on it. On the left side of the countertop, there is a microscope with a black body and a white \definecolor{customcolor}{rgb}{0.0,0.7764705882352941,0.7725490196078432}{\sethlcolor{customcolor}\textcolor{black}{\hl{lens\textsuperscript{1}}}}. Next to the microscope, there are two bottles of \definecolor{customcolor}{rgb}{0.28627450980392155,0.8941176470588236,0.6705882352941176}{\sethlcolor{customcolor}\textcolor{black}{\hl{condiments\textsuperscript{2}}}} {-} one with \definecolor{customcolor}{rgb}{1.0,0.5686274509803921,0.48627450980392156}{\sethlcolor{customcolor}\textcolor{black}{\hl{a red label\textsuperscript{3}}}}\definecolor{customcolor}{rgb}{1.0,0.43137254901960786,0.6078431372549019}{\sethlcolor{customcolor}\textcolor{black}{\hl{\textsuperscript{4}}}} and the other with green. On top of the microscope is \definecolor{customcolor}{rgb}{0.023529411764705882,0.4627450980392157,0.6588235294117647}{\sethlcolor{customcolor}\textcolor{black}{\hl{a yellow banana\textsuperscript{5}}}}, \definecolor{customcolor}{rgb}{0.043137254901960784,0.7529411764705882,0.8}{\sethlcolor{customcolor}\textcolor{black}{\hl{a blue spatula\textsuperscript{6}}}}, \definecolor{customcolor}{rgb}{0.6078431372549019,0.996078431372549,0.796078431372549}{\sethlcolor{customcolor}\textcolor{black}{\hl{a red plate\textsuperscript{7}}}}, and \definecolor{customcolor}{rgb}{1.0,0.8823529411764706,0.6196078431372549}{\sethlcolor{customcolor}\textcolor{black}{\hl{a yellow corn\textsuperscript{8}}}}\definecolor{customcolor}{rgb}{1.0,0.4823529411764706,0.4588235294117647}{\sethlcolor{customcolor}\textcolor{black}{\hl{\textsuperscript{9}}}} on the cob. In the center of the image, there appears to be \definecolor{customcolor}{rgb}{1.0,0.25882352941176473,0.5411764705882353}{\sethlcolor{customcolor}\textcolor{black}{\hl{a frying pan\textsuperscript{10}}}} with a \definecolor{customcolor}{rgb}{0.6862745098039216,0.1607843137254902,0.615686274509804}{\sethlcolor{customcolor}\textcolor{black}{\hl{fried egg\textsuperscript{11}}}} on it, and on the right side is \definecolor{customcolor}{rgb}{0.6,1.0,0.6941176470588235}{\sethlcolor{customcolor}\textcolor{black}{\hl{a white sink\textsuperscript{12}}}} with a white \definecolor{customcolor}{rgb}{0.6,0.7607843137254902,1.0}{\sethlcolor{customcolor}\textcolor{black}{\hl{faucet\textsuperscript{13}}}}. \definecolor{customcolor}{rgb}{1.0,0.7372549019607844,0.6}{\sethlcolor{customcolor}\textcolor{black}{\hl{The countertop\textsuperscript{14}}}} is made of wood and has a gray tile backsplash.
\end{alltt}
}
\end{AIbox}
\captionof{figure}{Visual grounding prediction results.}

\end{minipage}

\begin{minipage}{1.0\linewidth}

\centering
\begin{AIbox}{Visual Grounding}
{\bf Prompt}: Locate the phrases in the caption: \{caption\}

\parbox[h]{0.5\textwidth}{
    \scriptsize \begin{alltt}
    \centering
        \includegraphics[width=0.45\textwidth]{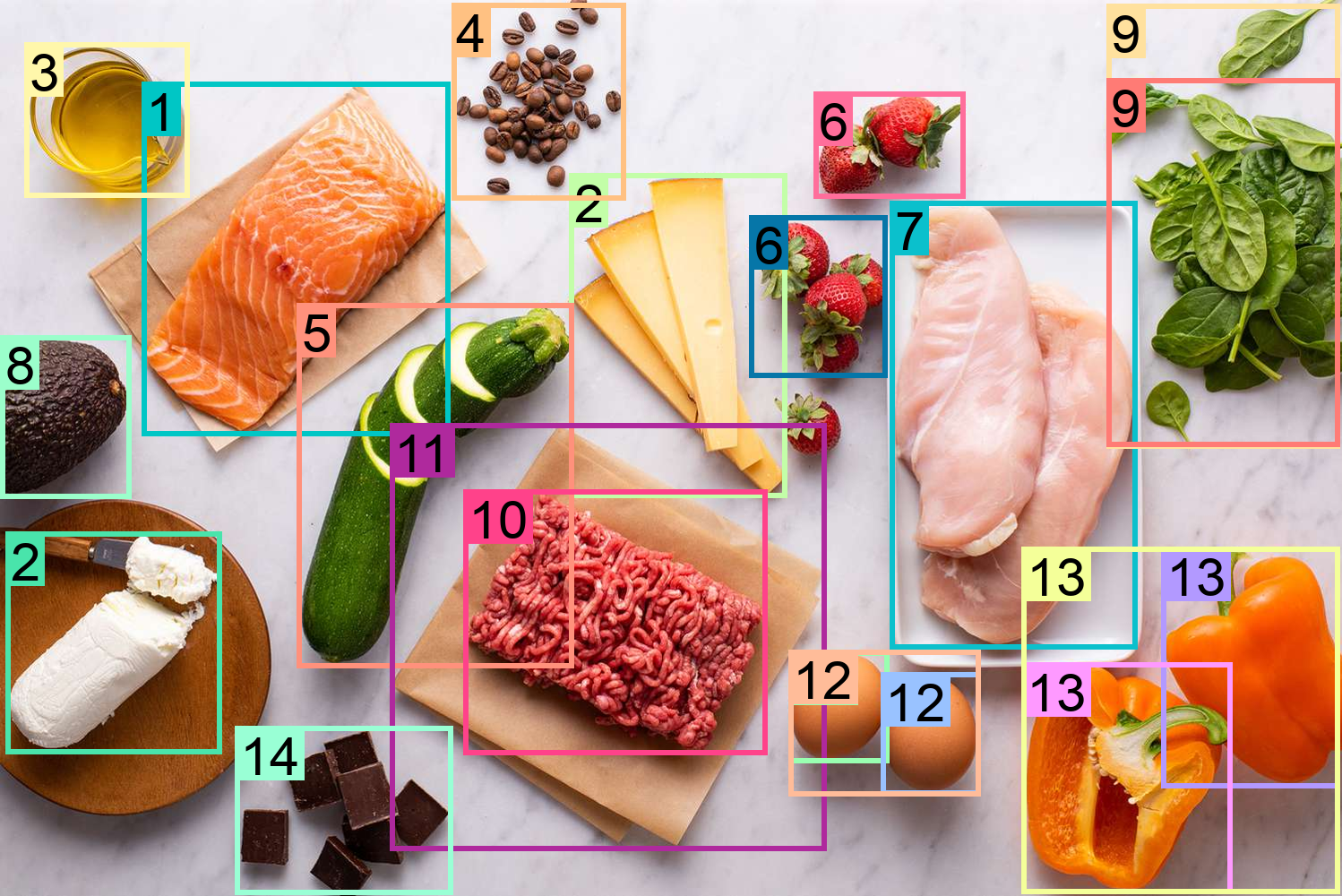}
    \end{alltt}
}
\hspace{10px}
\parbox[h]{0.50\textwidth}{ 
    \scriptsize \begin{alltt}
  \noindent The image is a flat lay of various food items arranged on a white marble countertop. On the left side of the image, there is \definecolor{customcolor}{rgb}{0.0,0.7764705882352941,0.7725490196078432}{\sethlcolor{customcolor}\textcolor{black}{\hl{a piece of salmon\textsuperscript{1}}}}. Next to it, there are \definecolor{customcolor}{rgb}{0.28627450980392155,0.8941176470588236,0.6705882352941176}{\sethlcolor{customcolor}\textcolor{black}{\hl{slices of cheese\textsuperscript{2}}}}, \definecolor{customcolor}{rgb}{1.0,0.9686274509803922,0.6705882352941176}{\sethlcolor{customcolor}\textcolor{black}{\hl{a glass of oil\textsuperscript{3}}}}, \definecolor{customcolor}{rgb}{1.0,0.7568627450980392,0.5098039215686274}{\sethlcolor{customcolor}\textcolor{black}{\hl{coffee beans\textsuperscript{4}}}}, \definecolor{customcolor}{rgb}{1.0,0.5686274509803921,0.48627450980392156}{\sethlcolor{customcolor}\textcolor{black}{\hl{a zucchini\textsuperscript{5}}}}, a bunch of \definecolor{customcolor}{rgb}{1.0,0.43137254901960786,0.6078431372549019}{\sethlcolor{customcolor}\textcolor{black}{\hl{strawberries\textsuperscript{6}}}}, two \definecolor{customcolor}{rgb}{0.043137254901960784,0.7529411764705882,0.8}{\sethlcolor{customcolor}\textcolor{black}{\hl{chicken breasts\textsuperscript{7}}}}, \definecolor{customcolor}{rgb}{0.6078431372549019,0.996078431372549,0.796078431372549}{\sethlcolor{customcolor}\textcolor{black}{\hl{a avocado\textsuperscript{8}}}} and \definecolor{customcolor}{rgb}{1.0,0.8823529411764706,0.6196078431372549}{\sethlcolor{customcolor}\textcolor{black}{\hl{a few whole spinach leaves\textsuperscript{9}}}}. In the center of the table, there appears to be  \definecolor{customcolor}{rgb}{1.0,0.25882352941176473,0.5411764705882353}{\sethlcolor{customcolor}\textcolor{black}{\hl{a pile of ground beef\textsuperscript{10}}}} on \definecolor{customcolor}{rgb}{0.6862745098039216,0.1607843137254902,0.615686274509804}{\sethlcolor{customcolor}\textcolor{black}{\hl{paper\textsuperscript{11}}}}, two \definecolor{customcolor}{rgb}{0.6,1.0,0.6941176470588235}{\sethlcolor{customcolor}\textcolor{black}{\hl{eggs\textsuperscript{12}}}}, two \definecolor{customcolor}{rgb}{0.6980392156862745,0.6,1.0}{\sethlcolor{customcolor}\textcolor{black}{\hl{orange bell peppers\textsuperscript{13}}}}, and \definecolor{customcolor}{rgb}{0.6,1.0,0.8352941176470589}{\sethlcolor{customcolor}\textcolor{black}{\hl{some dark chocolate bars\textsuperscript{14}}}}. The items are arranged in a way that suggests they are being prepared for a meal.
        
\end{alltt}
}
\tcbline
\parbox[h]{0.5\textwidth}{
    \scriptsize \begin{alltt}
        \centering
        \includegraphics[width=0.42\textwidth]{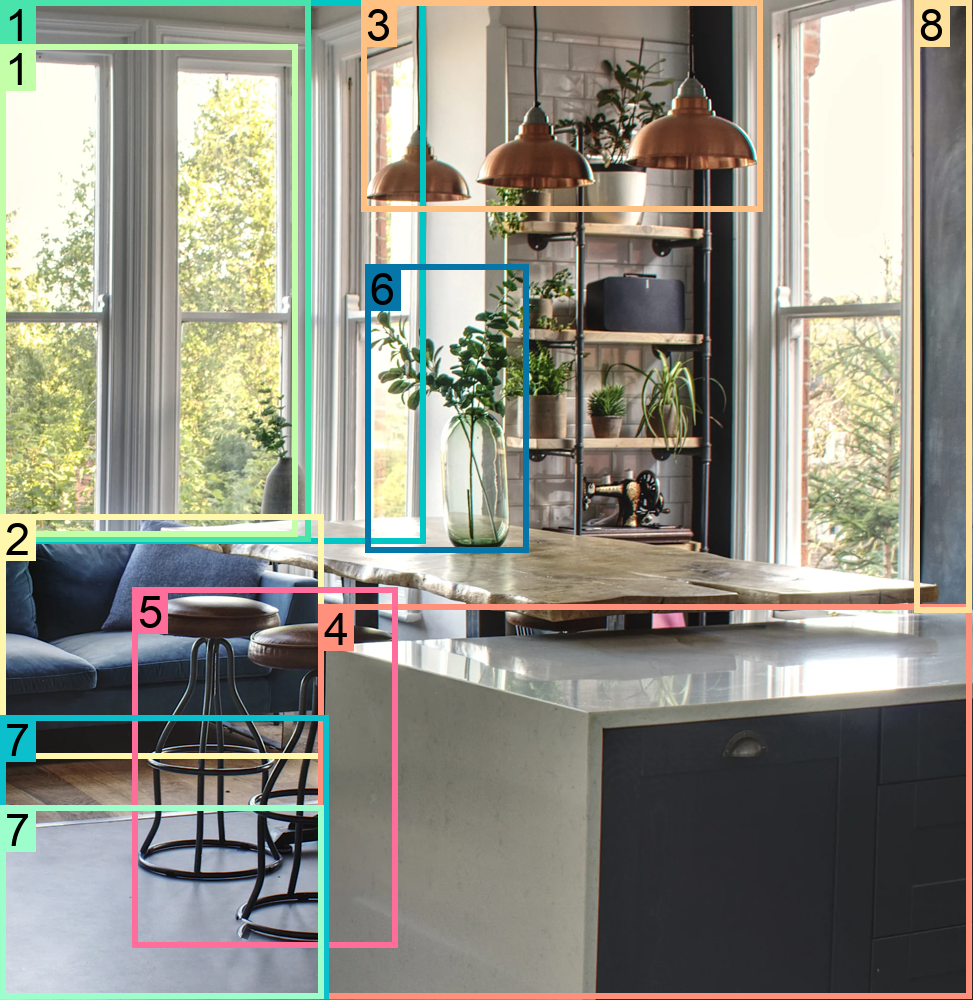}
    \end{alltt}
}
\hspace{10px}
\parbox[h]{0.50\textwidth}{ 
    \scriptsize \begin{alltt}
            \noindent The image shows a modern kitchen with a large window on the left side. \definecolor{customcolor}{rgb}{0.0,0.7764705882352941,0.7725490196078432}{\sethlcolor{customcolor}\textcolor{black}{\hl{The window\textsuperscript{1}}}} has a view of trees and greenery outside. On the left side of the image, there is \definecolor{customcolor}{rgb}{1.0,0.9686274509803922,0.6705882352941176}{\sethlcolor{customcolor}\textcolor{black}{\hl{a blue sofa\textsuperscript{2}}}} with a wooden coffee table in front of it. Above the table, there are \definecolor{customcolor}{rgb}{1.0,0.7568627450980392,0.5098039215686274}{\sethlcolor{customcolor}\textcolor{black}{\hl{three copper pendant lights\textsuperscript{3}}}} hanging from the ceiling. There is \definecolor{customcolor}{rgb}{1.0,0.5686274509803921,0.48627450980392156}{\sethlcolor{customcolor}\textcolor{black}{\hl{a large island\textsuperscript{4}}}} with a white countertop. There are \definecolor{customcolor}{rgb}{1.0,0.43137254901960786,0.6078431372549019}{\sethlcolor{customcolor}\textcolor{black}{\hl{two bar stools\textsuperscript{5}}}} next to the table. In the center of the kitchen, there is \definecolor{customcolor}{rgb}{0.023529411764705882,0.4627450980392157,0.6588235294117647}{\sethlcolor{customcolor}\textcolor{black}{\hl{a bottle green plants\textsuperscript{6}}}} on the table. \definecolor{customcolor}{rgb}{0.043137254901960784,0.7529411764705882,0.8}{\sethlcolor{customcolor}\textcolor{black}{\hl{The floor\textsuperscript{7}}}} is made of light{-}colored wood and \definecolor{customcolor}{rgb}{1.0,0.8823529411764706,0.6196078431372549}{\sethlcolor{customcolor}\textcolor{black}{\hl{the walls\textsuperscript{8}}}} are painted in a dark blue color.\newline%
\end{alltt}
}
\tcbline
\parbox[h]{0.5\textwidth}{
    \scriptsize \begin{alltt}
    \centering
        \includegraphics[width=0.45\textwidth]{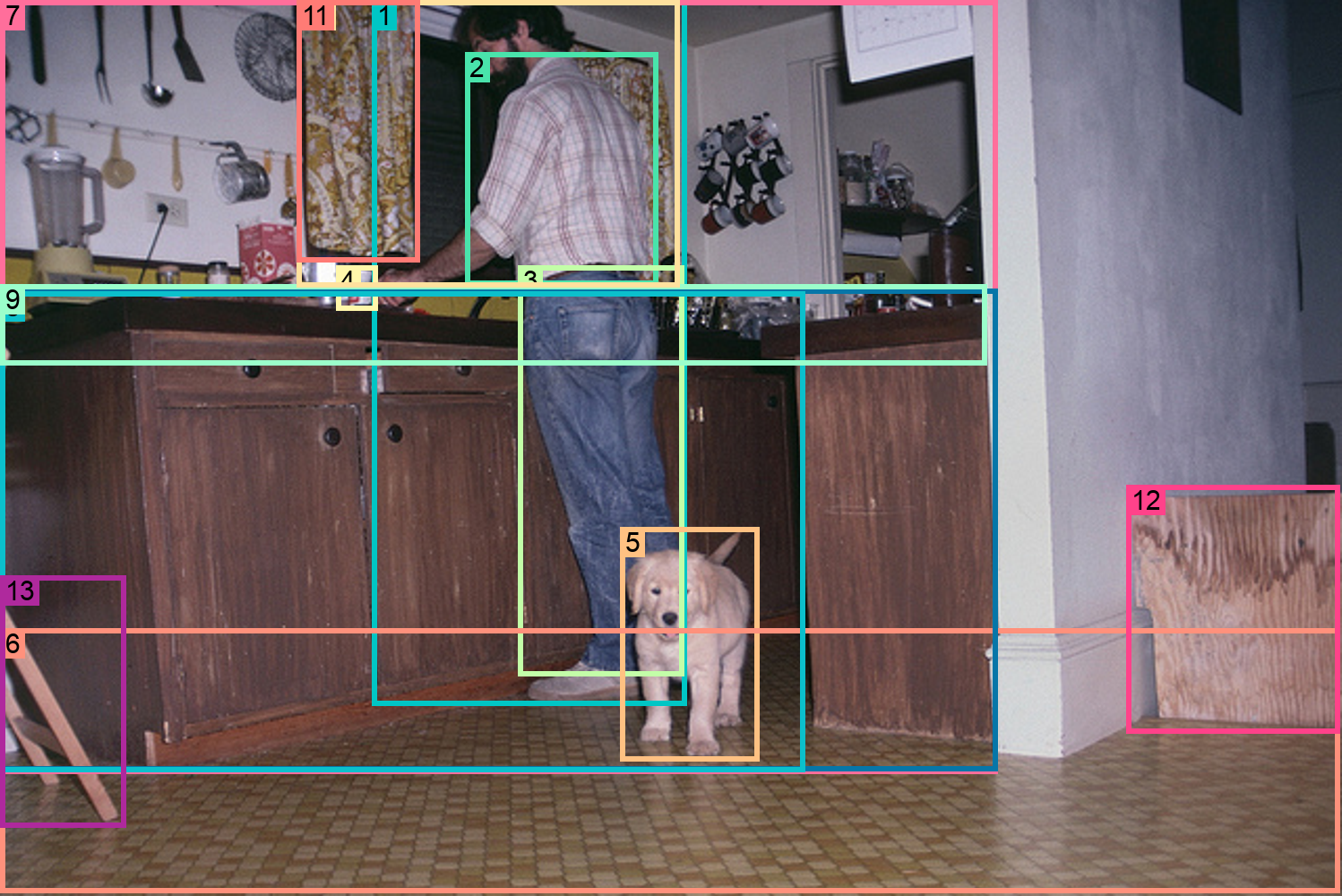}
    \end{alltt}
}
\hspace{10px}
\parbox[h]{0.50\textwidth}{ 
    \scriptsize \begin{alltt}
        \noindent The image shows a \definecolor{customcolor}{rgb}{0.0,0.7764705882352941,0.7725490196078432}{\sethlcolor{customcolor}\textcolor{black}{\hl{man\textsuperscript{1}}}} standing in a kitchen with a small dog. \definecolor{customcolor}{rgb}{0.0,0.7764705882352941,0.7725490196078432}{\sethlcolor{customcolor}\textcolor{black}{\hl{The man\textsuperscript{1}}}} is wearing a plaid \definecolor{customcolor}{rgb}{0.28627450980392155,0.8941176470588236,0.6705882352941176}{\sethlcolor{customcolor}\textcolor{black}{\hl{shirt\textsuperscript{2}}}} and \definecolor{customcolor}{rgb}{0.7647058823529411,0.996078431372549,0.6588235294117647}{\sethlcolor{customcolor}\textcolor{black}{\hl{jeans\textsuperscript{3}}}} and is holding a red \definecolor{customcolor}{rgb}{1.0,0.9686274509803922,0.6705882352941176}{\sethlcolor{customcolor}\textcolor{black}{\hl{cup\textsuperscript{4}}}} in his hand. \definecolor{customcolor}{rgb}{1.0,0.7568627450980392,0.5098039215686274}{\sethlcolor{customcolor}\textcolor{black}{\hl{The dog\textsuperscript{5}}}} is a light brown color and is standing on a tiled \definecolor{customcolor}{rgb}{1.0,0.5686274509803921,0.48627450980392156}{\sethlcolor{customcolor}\textcolor{black}{\hl{floor\textsuperscript{6}}}}. \definecolor{customcolor}{rgb}{1.0,0.43137254901960786,0.6078431372549019}{\sethlcolor{customcolor}\textcolor{black}{\hl{The kitchen\textsuperscript{7}}}} has wooden \definecolor{customcolor}{rgb}{0.023529411764705882,0.4627450980392157,0.6588235294117647}{\sethlcolor{customcolor}\textcolor{black}{\hl{cabinets\textsuperscript{8}}}} and a \definecolor{customcolor}{rgb}{0.6078431372549019,0.996078431372549,0.796078431372549}{\sethlcolor{customcolor}\textcolor{black}{\hl{countertop\textsuperscript{9}}}} with various kitchen utensils hanging on the wall. There is \definecolor{customcolor}{rgb}{1.0,0.8823529411764706,0.6196078431372549}{\sethlcolor{customcolor}\textcolor{black}{\hl{a window\textsuperscript{10}}}} with yellow \definecolor{customcolor}{rgb}{1.0,0.4823529411764706,0.4588235294117647}{\sethlcolor{customcolor}\textcolor{black}{\hl{curtains\textsuperscript{11}}}} in the background. On the right side of the image, there is \definecolor{customcolor}{rgb}{1.0,0.25882352941176473,0.5411764705882353}{\sethlcolor{customcolor}\textcolor{black}{\hl{a wooden cutting board\textsuperscript{12}}}} and a wooden \definecolor{customcolor}{rgb}{0.6862745098039216,0.1607843137254902,0.615686274509804}{\sethlcolor{customcolor}\textcolor{black}{\hl{stool\textsuperscript{13}}}}.
\end{alltt}
}

\end{AIbox}
\captionof{figure}{Visual grounding prediction results. (continued)}
\end{minipage}

\subsection{Dense Region Caption}
\begin{minipage}{1.0\linewidth}
\centering
\begin{AIbox}{Dense Region Caption}

\parbox[t]{0.45\textwidth}{
       \centering \includegraphics[width=0.45\textwidth]{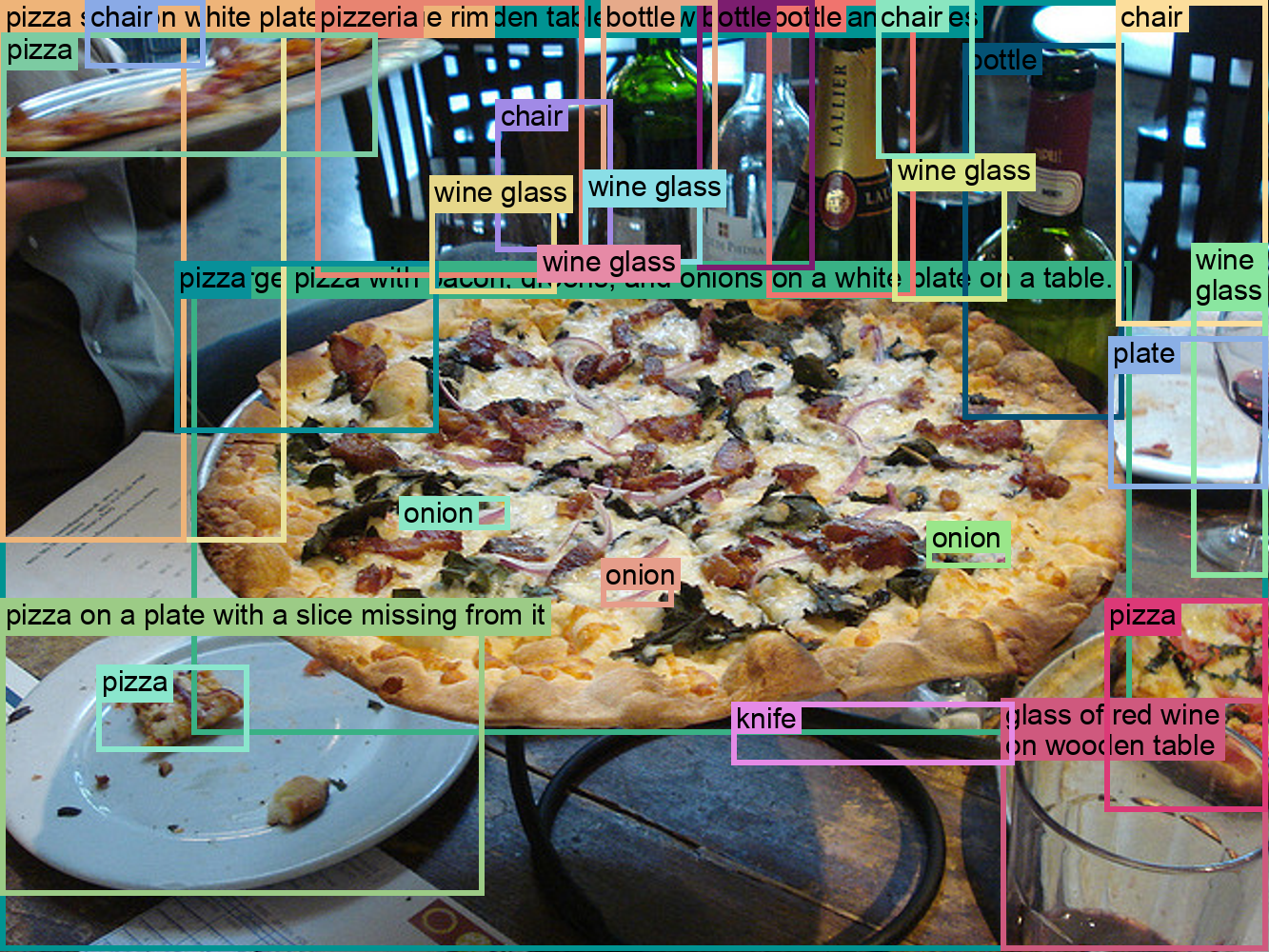}
}
\hspace{10px}
\parbox[t]{0.50\textwidth}{
       \centering \includegraphics[width=0.50\textwidth]{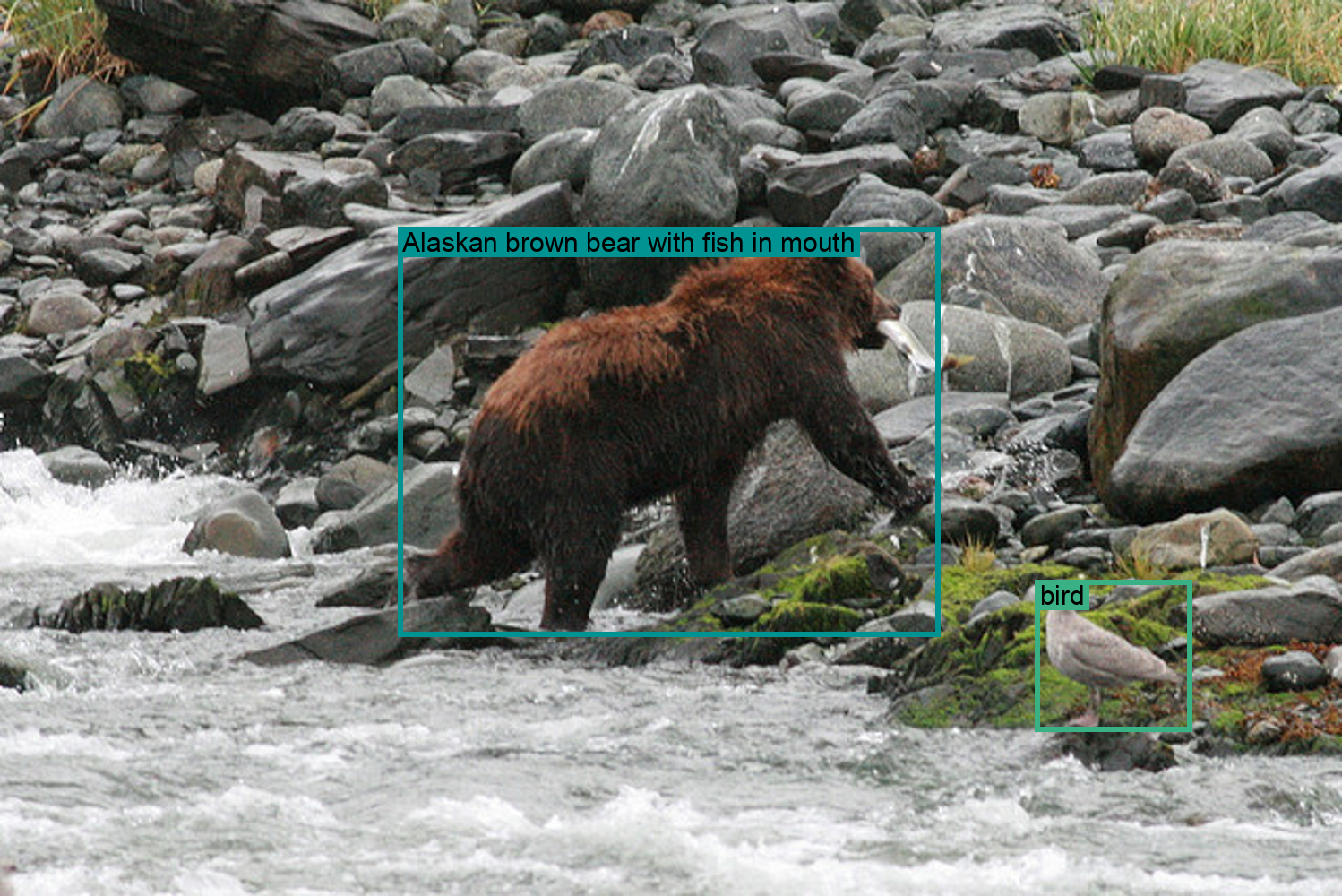}
}

\tcbline

\parbox[t]{0.45\textwidth}{
        
       \centering \includegraphics[width=0.40\textwidth]{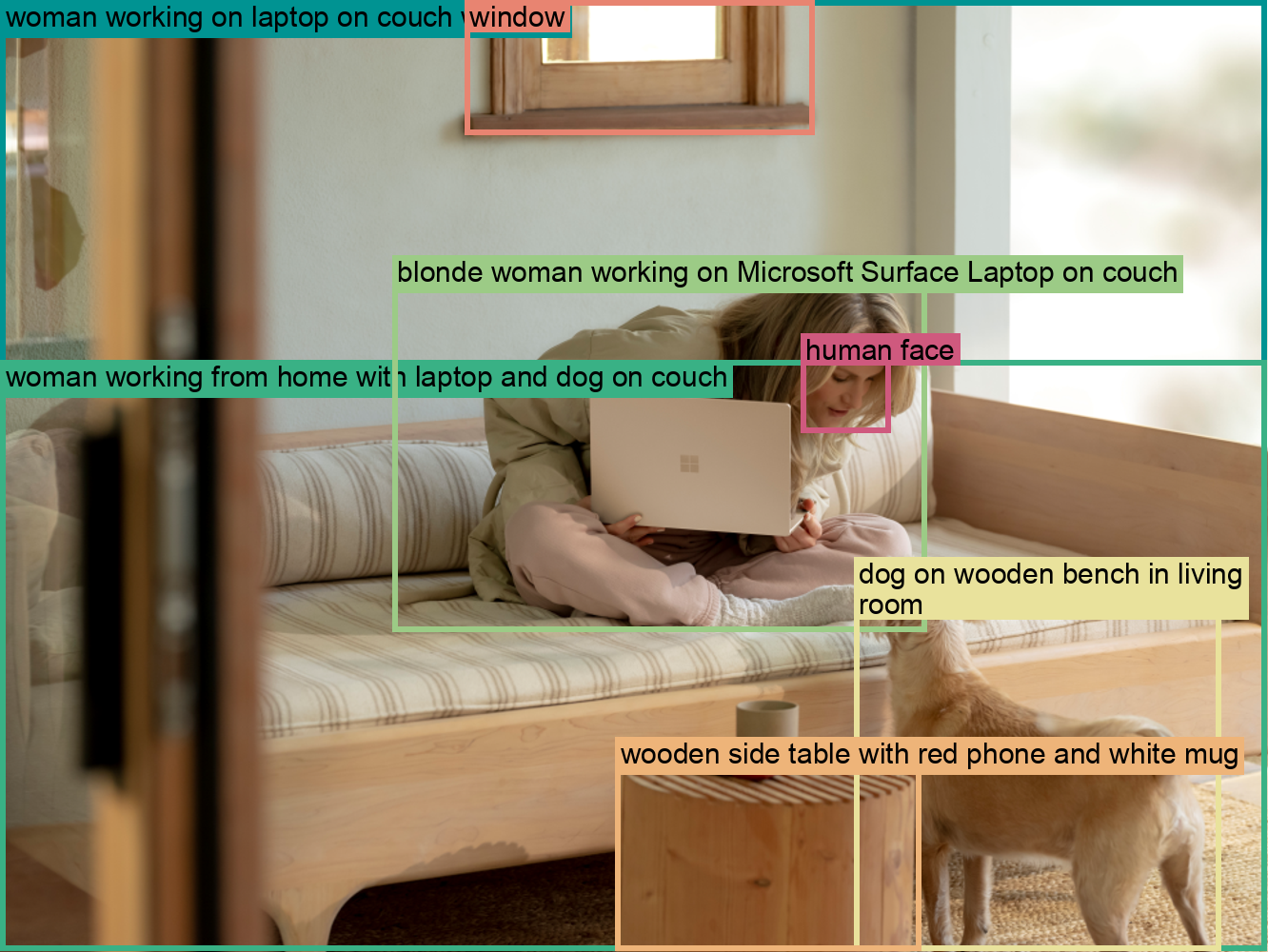}
}
\hspace{10px}
\parbox[t]{0.50\textwidth}{
        
      \centering  \includegraphics[width=0.38\textwidth]{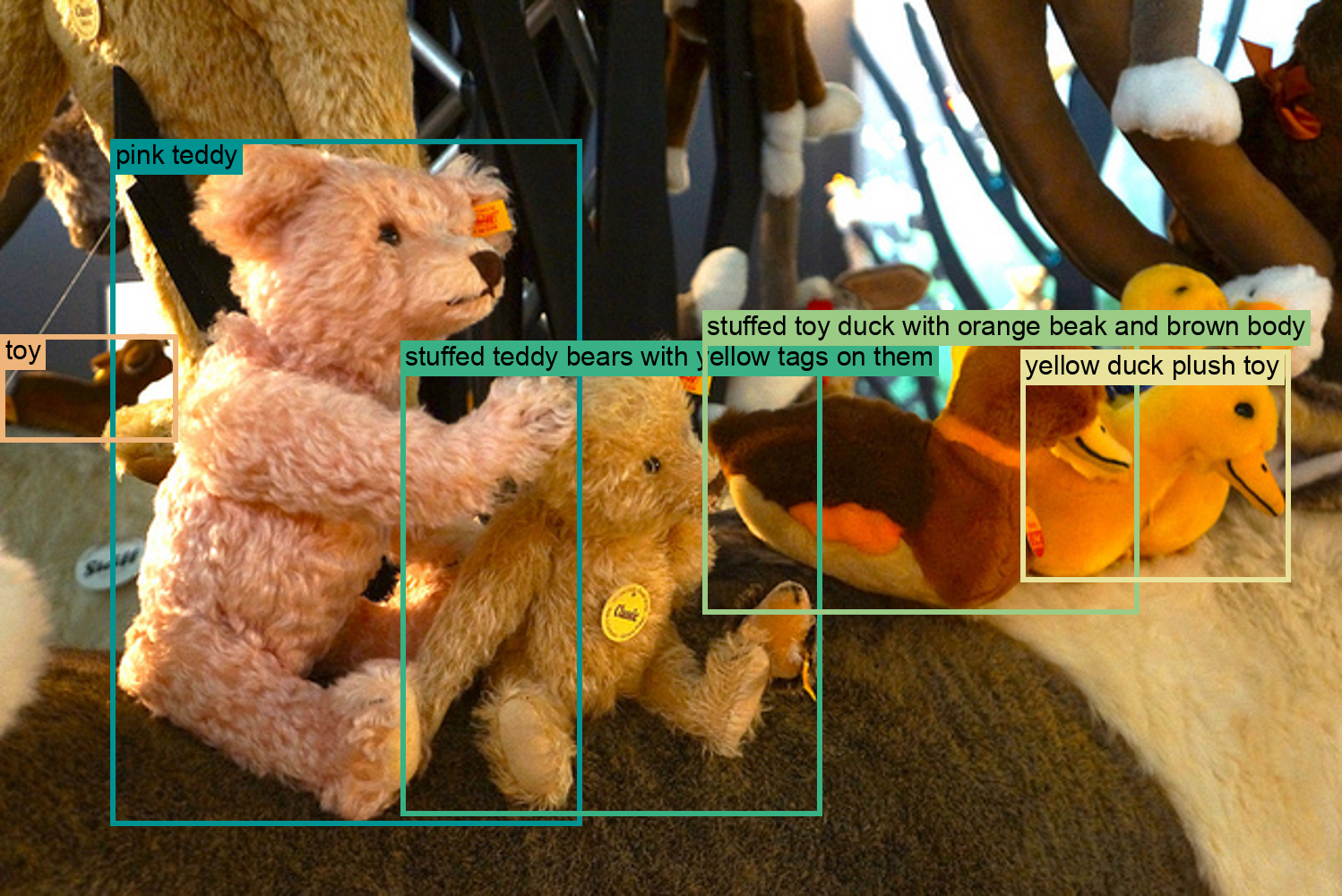}
}

\tcbline

\parbox[t]{0.45\textwidth}{
       \centering \includegraphics[width=0.35\textwidth]{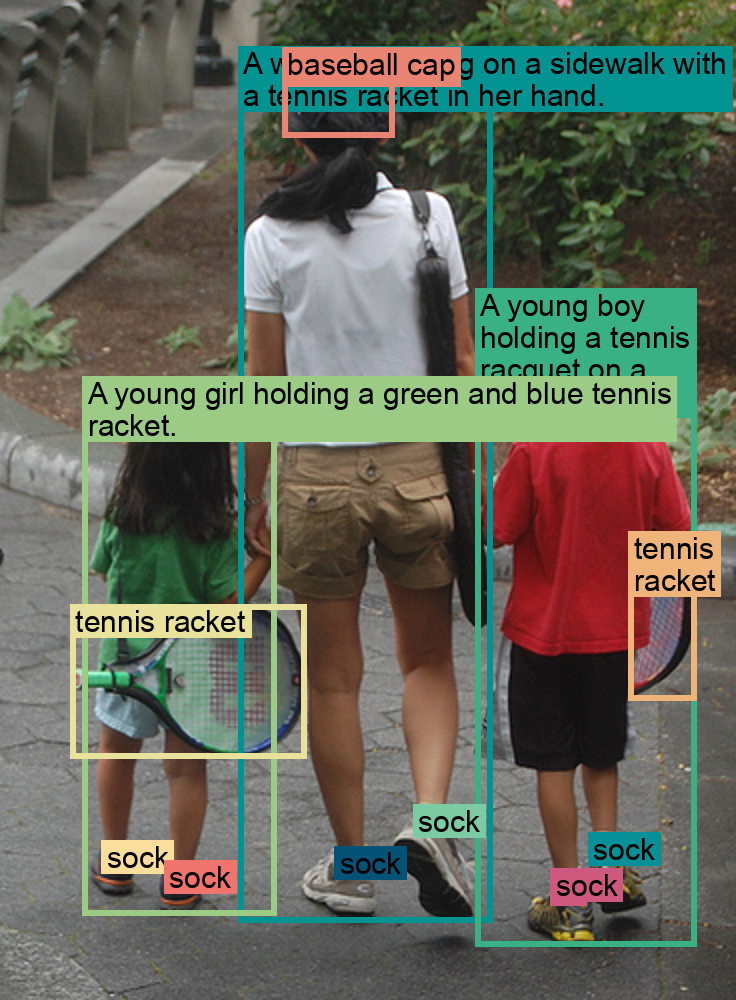}
}
\hspace{10px}
\parbox[t]{0.50\textwidth}{
       \centering \includegraphics[width=0.35\textwidth]{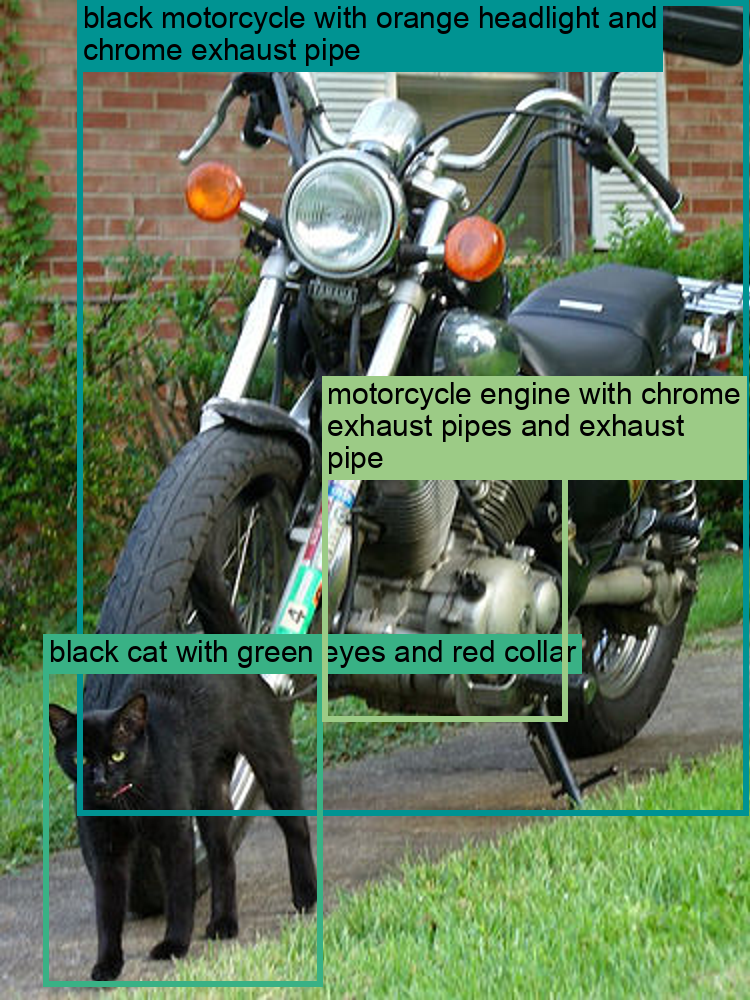}
}

\end{AIbox}
\captionof{figure}{Dense region caption prediction results.}
\end{minipage}

\subsection{Open Vocabulary Detection}
\begin{minipage}{1.0\linewidth}
\centering
\begin{AIbox}{Open Vocabulary Object Detection}

\parbox[t]{0.45\textwidth}{
        {\bf Prompt}: Locate \hl{Five Alive juice box}$\langle$and$\rangle$\hl{Colgate toothpaste} in the image.
        \\
        
       \centering \includegraphics[width=0.45\textwidth]{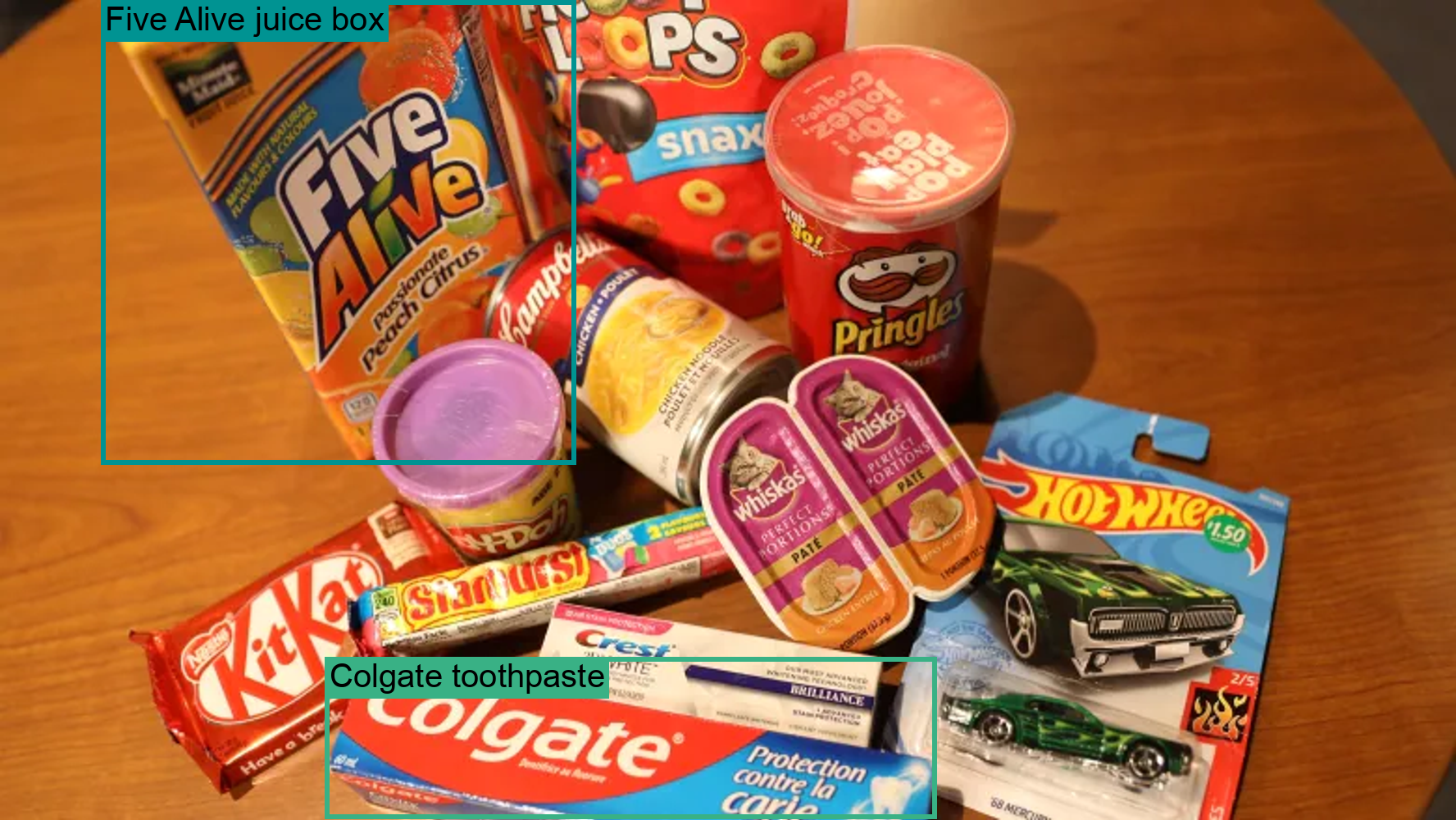}
}
\hspace{10px}
\parbox[t]{0.50\textwidth}{
        {\bf Prompt}: Locate \hl{Chewbacca} in the image.
        \\
        
       \centering \includegraphics[width=0.50\textwidth]{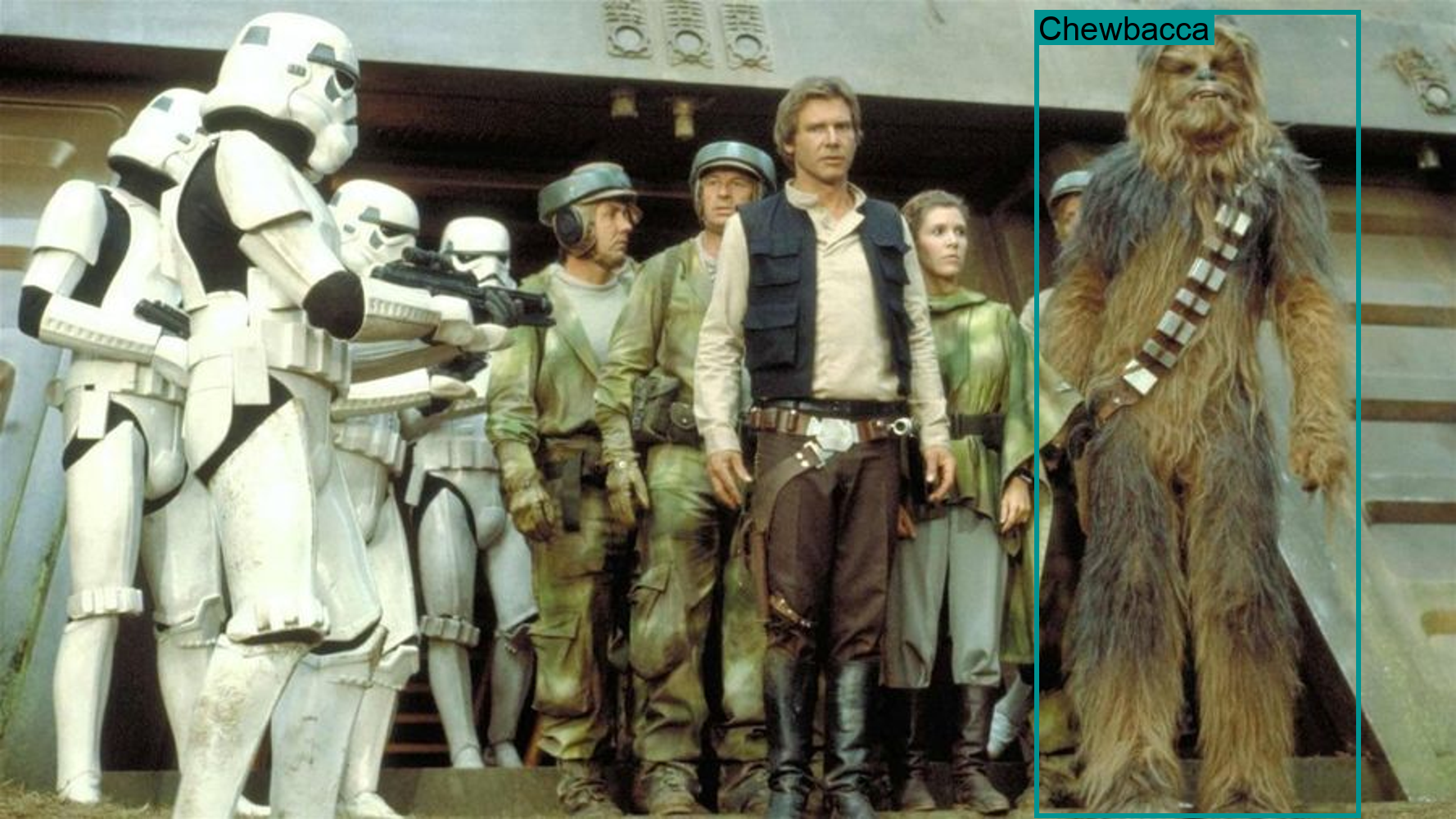}
}

\tcbline

\parbox[t]{0.45\textwidth}{
        {\bf Prompt}: 
        Locate \hl{giraffe} in the image.
        \\
        
       \centering \includegraphics[width=0.40\textwidth]{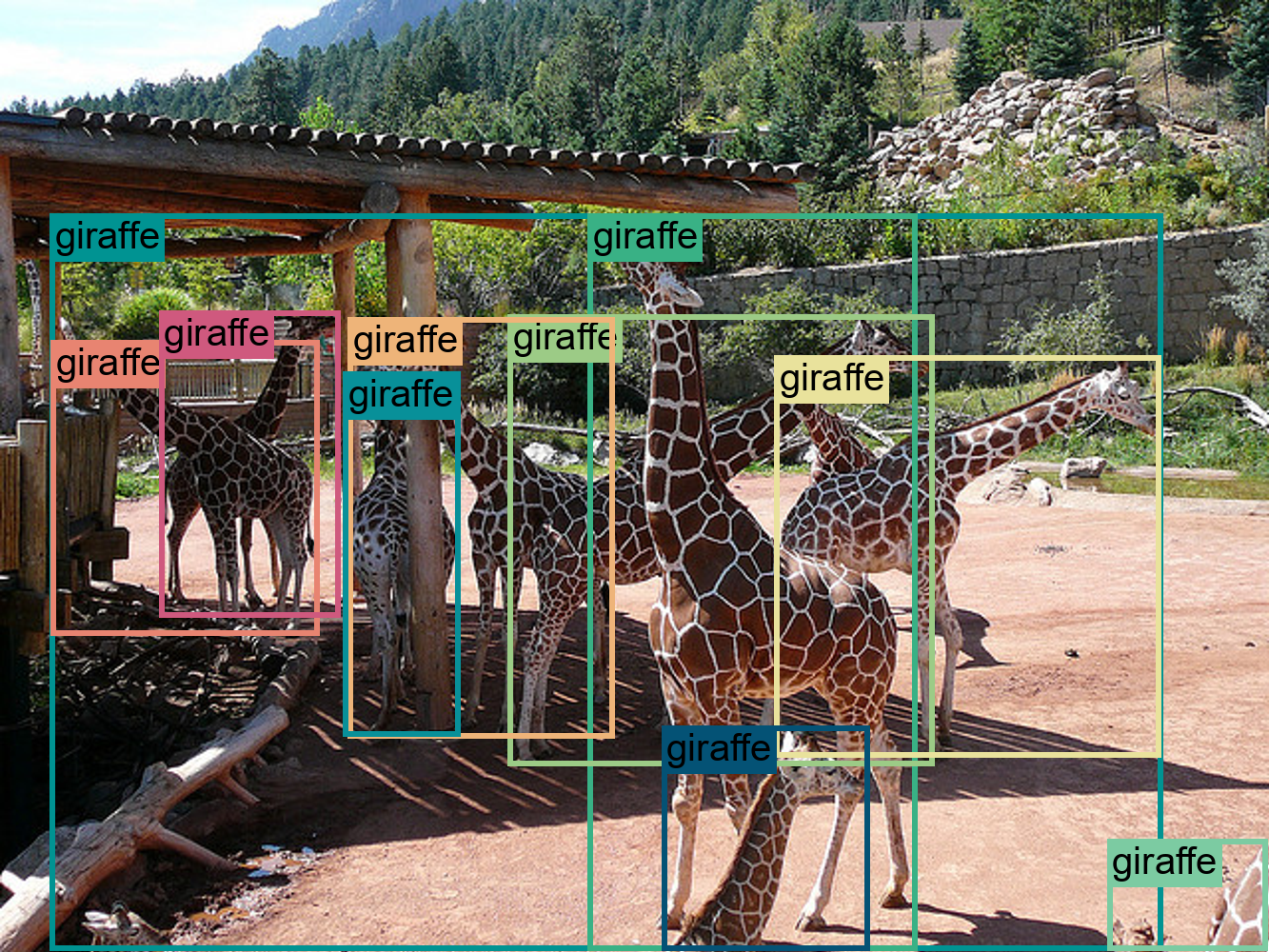}
}
\hspace{10px}
\parbox[t]{0.50\textwidth}{
        {\bf Prompt}: 
        Locate \hl{Mercedes-Benz}$\langle$and$\rangle$\hl{M2}$\langle$and$\rangle$\hl{Audi} in the image.
        \\
        
      \centering  \includegraphics[width=0.38\textwidth]{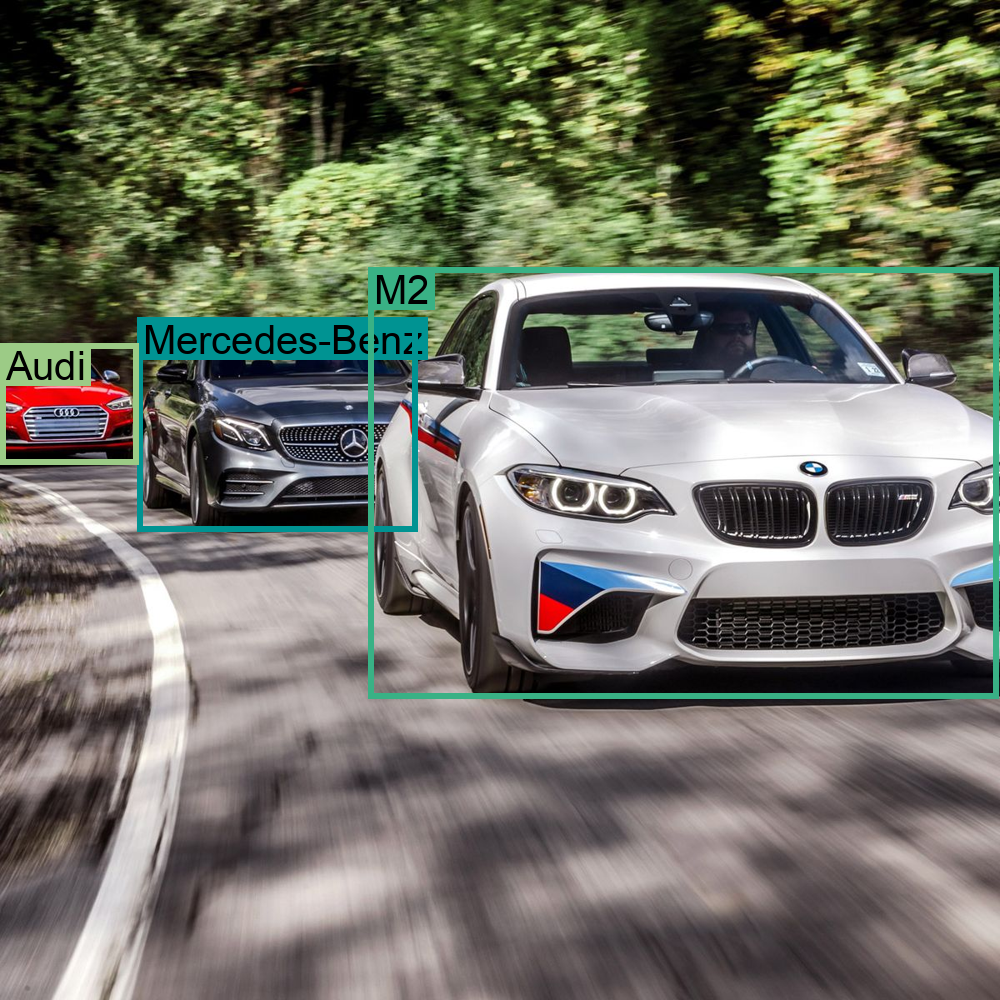}
}

\tcbline

\parbox[t]{0.45\textwidth}{
        {\bf Prompt}: Locate the \hl{objects with category name} in the image.
        \\
       \centering \includegraphics[width=0.45\textwidth]{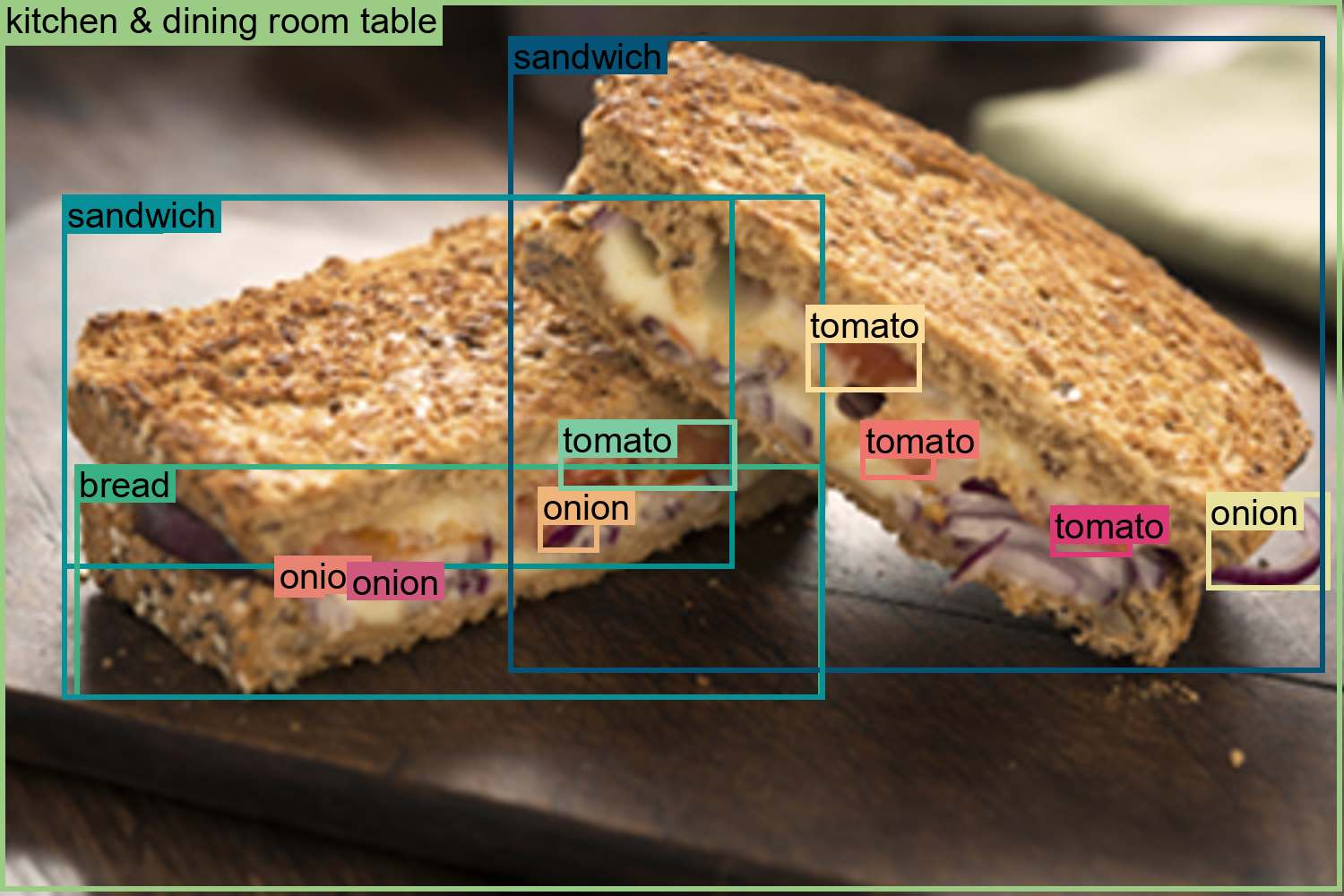}
}
\hspace{10px}
\parbox[t]{0.50\textwidth}{
        {\bf Prompt}: Locate the \hl{objects with category name} in the image.
        \\
       \centering \includegraphics[width=0.45\textwidth]{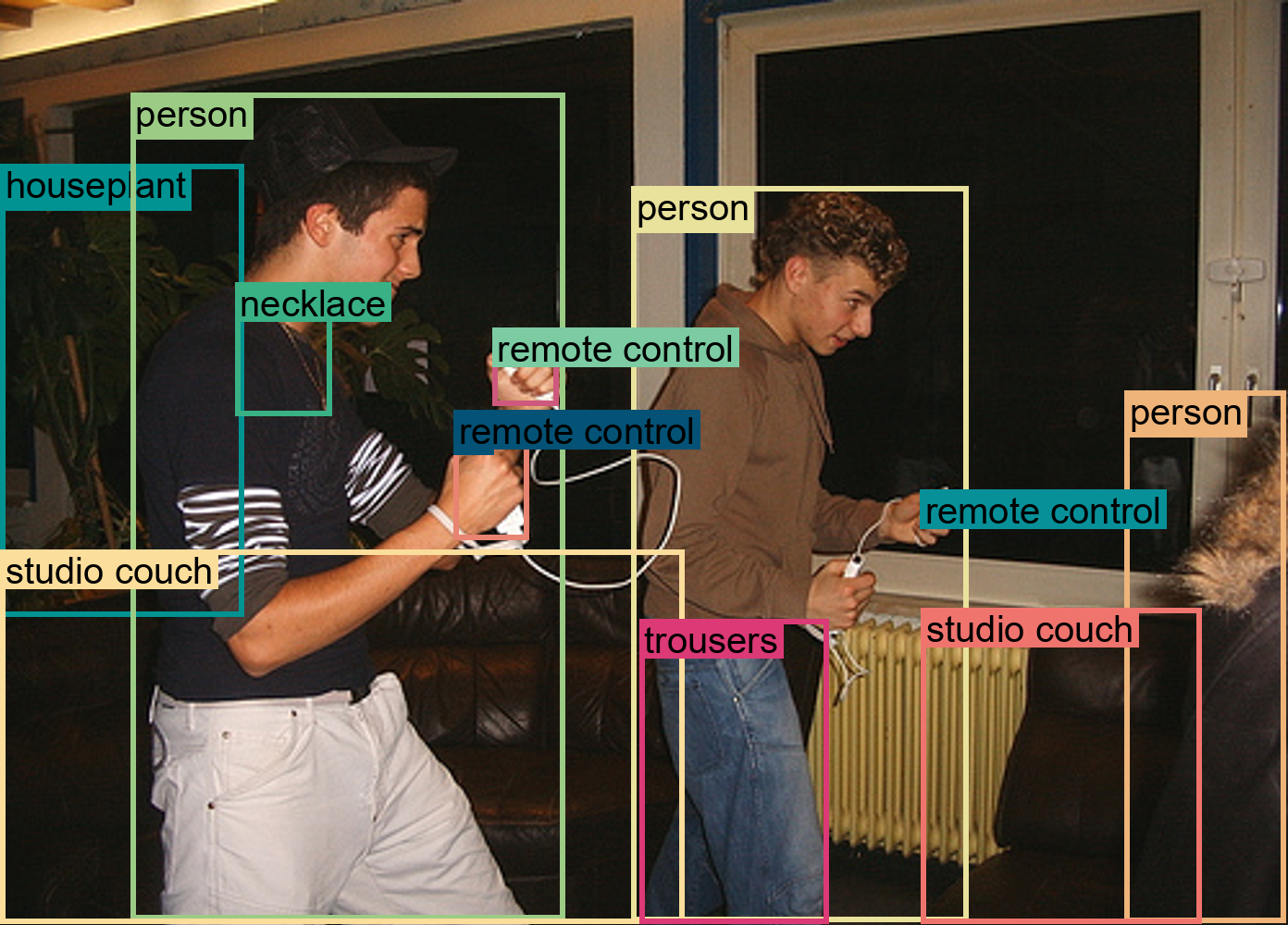}
}

\end{AIbox}

\captionof{figure}{Open vocabulary object detection prediction results.}
\end{minipage}

\subsection{OCR}

\begin{minipage}{1.0\linewidth}
    
\centering
\begin{AIbox}{Ocr with region}
{\bf Prompt}: What is the text in the image, with regions?

\parbox[h]{0.5\textwidth}{
    \centering
    \includegraphics[width=0.45\textwidth]{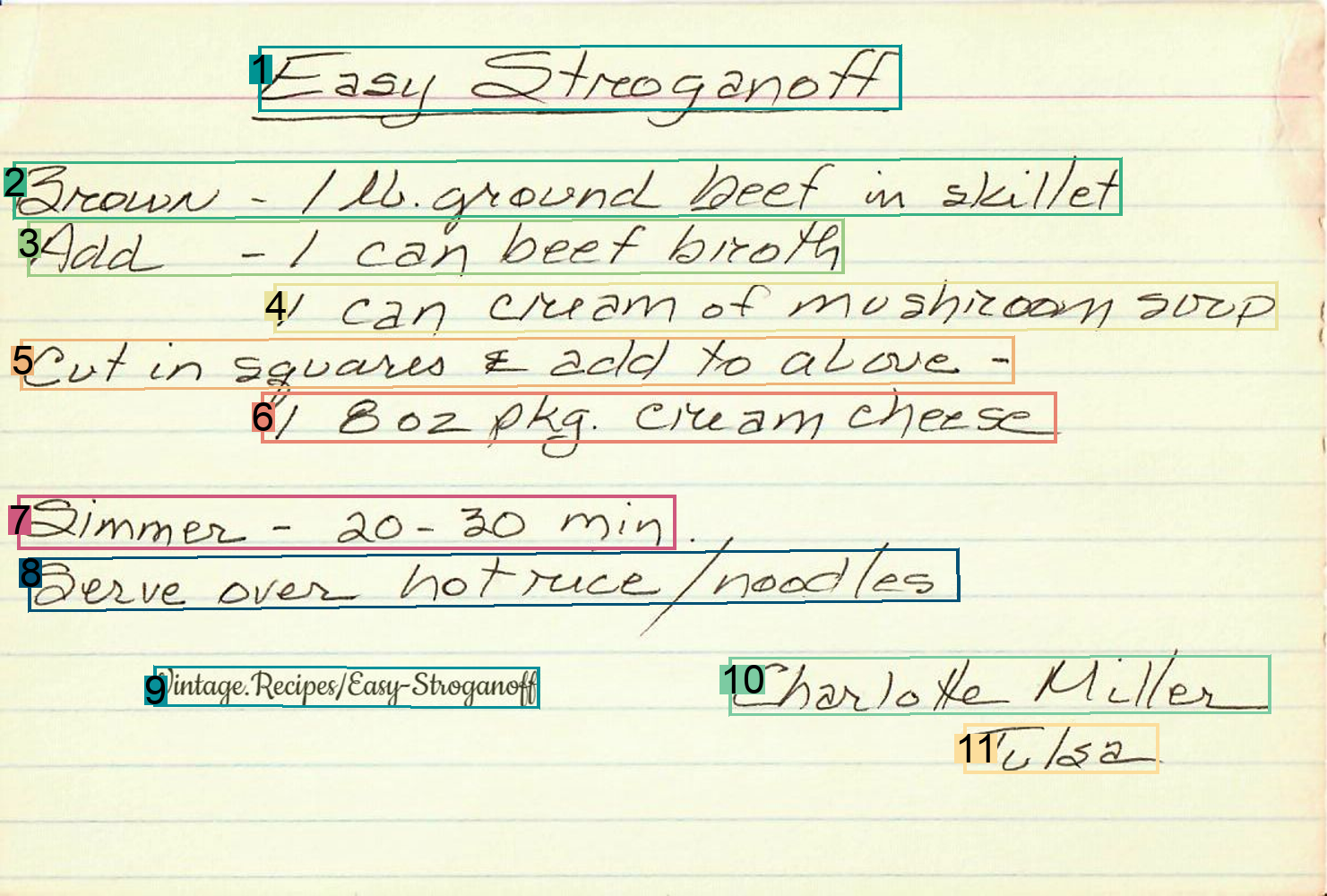}
}
\hspace{10px}
\parbox[h]{0.5\textwidth}{
    \scriptsize \begin{alltt}
     \definecolor{customcolor}{rgb}{0.0,0.7764705882352941,0.7725490196078432}{\sethlcolor{customcolor} \textcolor{black}
     {\hl{Easy Stroganoff\textsuperscript{1}}}}\\
     \definecolor{customcolor}{rgb}{0.28627450980392155,0.8941176470588236,0.6705882352941176}{\sethlcolor{customcolor} \textcolor{black}
     {\hl{Brown {-} 1 lb. ground beef in skillet\textsuperscript{2}}}}\\
     \definecolor{customcolor}{rgb}{0.7647058823529411,0.996078431372549,0.6588235294117647}{\sethlcolor{customcolor} \textcolor{black}
     {\hl{Add {-} 1 can beef broth\textsuperscript{3}}}}\\
     \definecolor{customcolor}{rgb}{1.0,0.9686274509803922,0.6705882352941176}{\sethlcolor{customcolor} \textcolor{black}
     {\hl{1 can cream of mushroom soup\textsuperscript{4}}}}\\
     \definecolor{customcolor}{rgb}{1.0,0.7568627450980392,0.5098039215686274}{\sethlcolor{customcolor} \textcolor{black}
     {\hl{Cut in squares \& 2dld to above {-}\textsuperscript{5}}}}\\
     \definecolor{customcolor}{rgb}{1.0,0.5686274509803921,0.48627450980392156}{\sethlcolor{customcolor} \textcolor{black}
     {\hl{1/ Boz pkg. cream cheese\textsuperscript{6}}}} \\
     \definecolor{customcolor}{rgb}{1.0,0.43137254901960786,0.6078431372549019}{\sethlcolor{customcolor} \textcolor{black}
     {\hl{Simmer {-} 20{-}30 min.\textsuperscript{7}}}}\\
     \definecolor{customcolor}{rgb}{0.023529411764705882,0.4627450980392157,0.6588235294117647}{\sethlcolor{customcolor} \textcolor{black}
     {\hl{Serve over hotrice /noodles.\textsuperscript{8}}}}\\
     \definecolor{customcolor}{rgb}{0.043137254901960784,0.7529411764705882,0.8}{\sethlcolor{customcolor} \textcolor{black}
     {\hl{Vintage. Recipes/Easy{-}Stroganof\textsuperscript{9}}}} 
     \definecolor{customcolor}{rgb}{0.6078431372549019,0.996078431372549,0.796078431372549}{\sethlcolor{customcolor} \textcolor{black}
     {\hl{Charlotte Miller\textsuperscript{10}}}}\\ 
     \definecolor{customcolor}{rgb}{1.0,0.8823529411764706,0.6196078431372549}{\sethlcolor{customcolor} \textcolor{black}
     {\hl{Tulsa\textsuperscript{11}}}}
    \end{alltt}
}
\tcbline
\parbox[h]{0.5\textwidth}{
    \centering
    \includegraphics[width=0.45\textwidth]{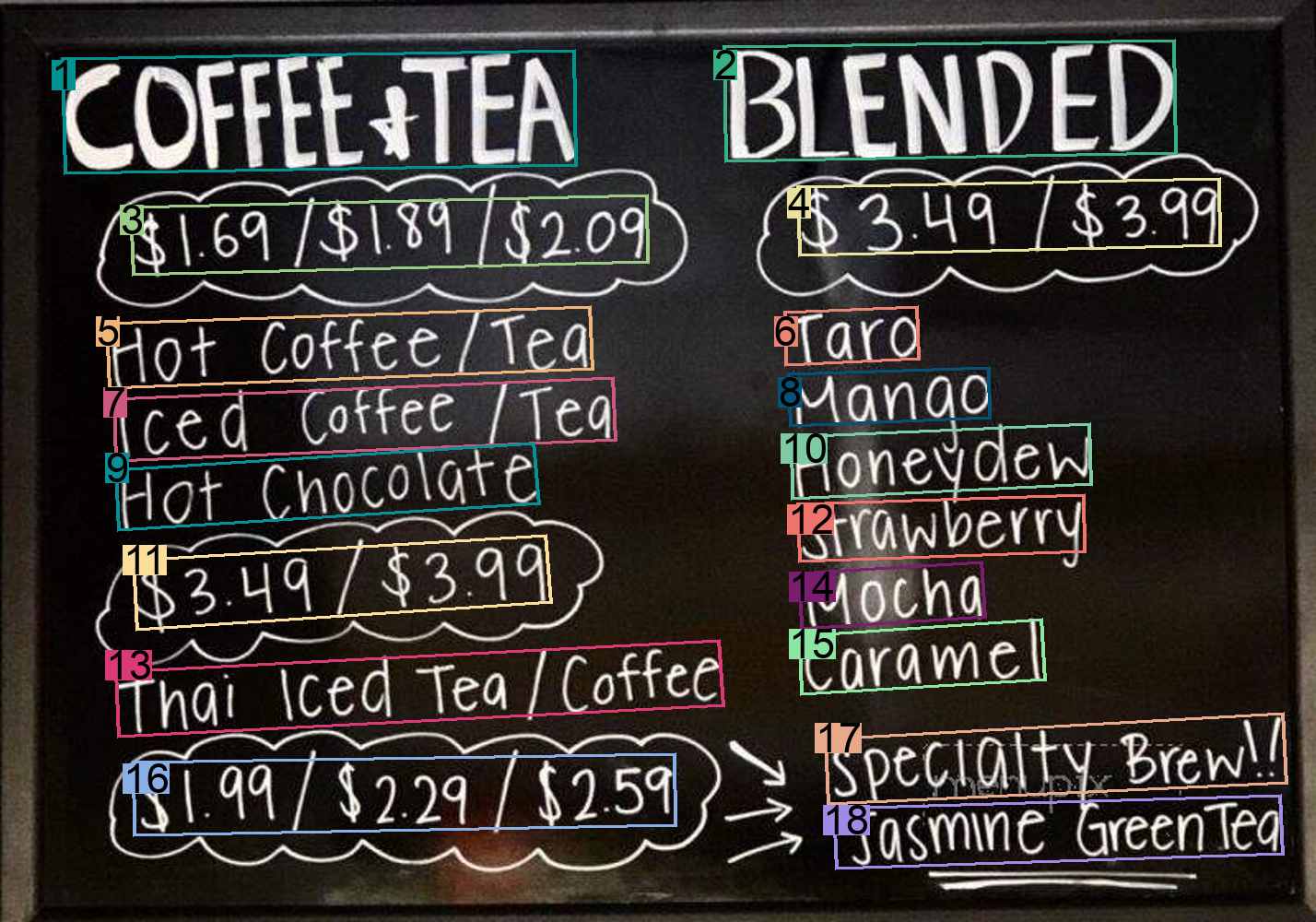}
}
\hspace{10px}
\parbox[h]{0.5\textwidth}{
    \scriptsize \begin{alltt}
      \definecolor{customcolor}{rgb}{0.0,0.7764705882352941,0.7725490196078432}{\sethlcolor{customcolor} \textcolor{black}{\hl{COFFEE+TEA\textsuperscript{1}}}} 
      \definecolor{customcolor}{rgb}{0.28627450980392155,0.8941176470588236,0.6705882352941176}{\sethlcolor{customcolor} \textcolor{black}{\hl{BLENDED\textsuperscript{2}}}}\\
      \definecolor{customcolor}{rgb}{0.7647058823529411,0.996078431372549,0.6588235294117647}{\sethlcolor{customcolor} \textcolor{black}{\hl{\$1.69/\$1.89/\$2.09\textsuperscript{3}}}} 
      \definecolor{customcolor}{rgb}{1.0,0.9686274509803922,0.6705882352941176}{\sethlcolor{customcolor} \textcolor{black}{\hl{\$3.49/\$3.99\textsuperscript{4}}}}\\
      \definecolor{customcolor}{rgb}{1.0,0.7568627450980392,0.5098039215686274}{\sethlcolor{customcolor} \textcolor{black}{\hl{Hot Coffee/Tea\textsuperscript{5}}}}
      \definecolor{customcolor}{rgb}{1.0,0.5686274509803921,0.48627450980392156}{\sethlcolor{customcolor} \textcolor{black}{\hl{Taro\textsuperscript{6}}}}\\
      \definecolor{customcolor}{rgb}{1.0,0.43137254901960786,0.6078431372549019}{\sethlcolor{customcolor} \textcolor{black}{\hl{Iced Coffee/ Tea\textsuperscript{7}}}} 
      \definecolor{customcolor}{rgb}{0.023529411764705882,0.4627450980392157,0.6588235294117647}{\sethlcolor{customcolor} \textcolor{black}{\hl{Mango\textsuperscript{8}}}}\\
      \definecolor{customcolor}{rgb}{0.043137254901960784,0.7529411764705882,0.8}{\sethlcolor{customcolor} \textcolor{black}{\hl{Hot Chocolate\textsuperscript{9}}}}
      \definecolor{customcolor}{rgb}{0.6078431372549019,0.996078431372549,0.796078431372549}{\sethlcolor{customcolor} \textcolor{black}{\hl{Honeydew\textsuperscript{10}}}} \\
      \definecolor{customcolor}{rgb}{1.0,0.8823529411764706,0.6196078431372549}{\sethlcolor{customcolor} \textcolor{black}{\hl{\$3,49/\$ 3.99\textsuperscript{11}}}}
      \definecolor{customcolor}{rgb}{1.0,0.4823529411764706,0.4588235294117647}{\sethlcolor{customcolor} \textcolor{black}{\hl{Strawberry\textsuperscript{12}}}} 
      \definecolor{customcolor}{rgb}{0.6862745098039216,0.1607843137254902,0.615686274509804}{\sethlcolor{customcolor} \textcolor{black}{\hl{Mocha\textsuperscript{14}}}}\\
      \definecolor{customcolor}{rgb}{1.0,0.25882352941176473,0.5411764705882353}{\sethlcolor{customcolor} \textcolor{black}{\hl{Thai Iced Tea / Coffee\textsuperscript{13}}}}
      \definecolor{customcolor}{rgb}{0.6,1.0,0.6941176470588235}{\sethlcolor{customcolor} \textcolor{black}
      {\hl{Caramel\textsuperscript{15}}}} \\
      \definecolor{customcolor}{rgb}{0.6,0.7607843137254902,1.0}{\sethlcolor{customcolor} \textcolor{black}{\hl{\$1,99/\$2,29/\$2:59\textsuperscript{16}}}}
      \definecolor{customcolor}{rgb}{1.0,0.7372549019607844,0.6}{\sethlcolor{customcolor} \textcolor{black}{\hl{SPECIALTY Brew !!\textsuperscript{17}}}}  \\
      \definecolor{customcolor}{rgb}{0.6980392156862745,0.6,1.0}{\sethlcolor{customcolor} \textcolor{black}{\hl{Jasmine GreenTea\textsuperscript{18}}}}
    \end{alltt}
}

\tcbline
\parbox[h]{0.50\textwidth}{
    \centering
    \includegraphics[width=0.36\textwidth]{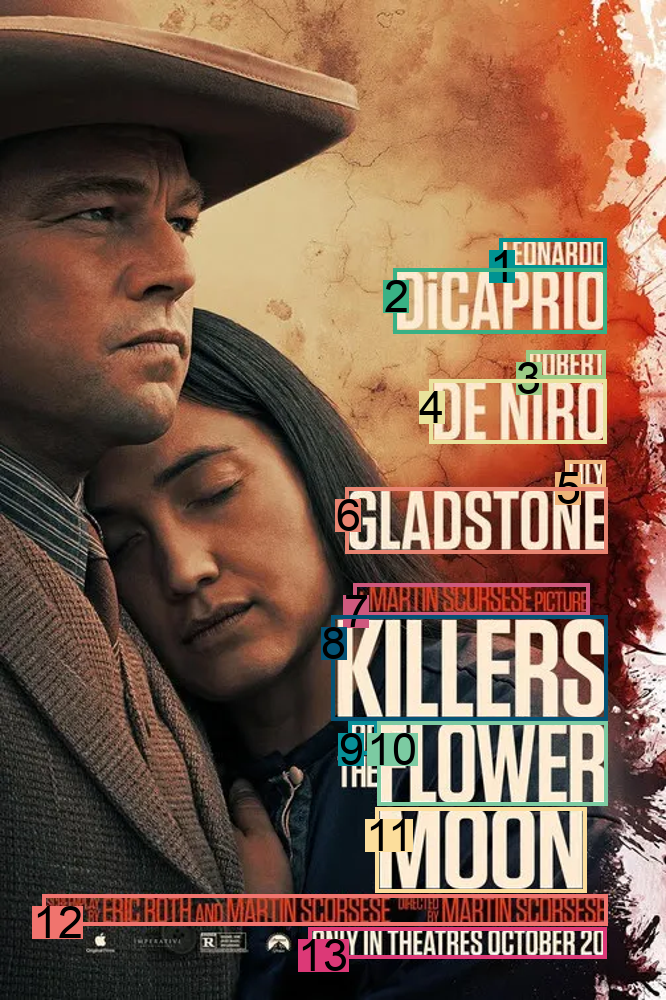}
}
\hspace{10px}
\parbox[h]{0.5\textwidth}{
\scriptsize \begin{alltt}
\definecolor{customcolor}{rgb}{0.0,0.7764705882352941,0.7725490196078432}{\sethlcolor{customcolor} \textcolor{black}
\noindent{\hl{LEONARDO\textsuperscript{1}}}}\\
\definecolor{customcolor}{rgb}{0.28627450980392155,0.8941176470588236,0.6705882352941176}{\sethlcolor{customcolor} \textcolor{black}{\hl{DiCAPRIO\textsuperscript{2}}}}\\
\definecolor{customcolor}{rgb}{0.7647058823529411,0.996078431372549,0.6588235294117647}{\sethlcolor{customcolor} \textcolor{black}{\hl{ROBERT\textsuperscript{3}}}}\\
\definecolor{customcolor}{rgb}{1.0,0.9686274509803922,0.6705882352941176}{\sethlcolor{customcolor} \textcolor{black}
{\hl{DE NIRO\textsuperscript{4}}}}\\
\definecolor{customcolor}{rgb}{1.0,0.7568627450980392,0.5098039215686274}{\sethlcolor{customcolor} \textcolor{black}
{\hl{LILY\textsuperscript{5}}}}\\
\definecolor{customcolor}{rgb}{1.0,0.5686274509803921,0.48627450980392156}{\sethlcolor{customcolor} \textcolor{black}
{\hl{GLADSTONE\textsuperscript{6}}}}\\
\definecolor{customcolor}{rgb}{1.0,0.43137254901960786,0.6078431372549019}{\sethlcolor{customcolor} \textcolor{black}{\hl{A MARTIN SCORSESE PICTURE\textsuperscript{7}}}}\\
\definecolor{customcolor}{rgb}{0.023529411764705882,0.4627450980392157,0.6588235294117647}{\sethlcolor{customcolor} \textcolor{black}{\hl{KILLERS\textsuperscript{8}}}}\\
\definecolor{customcolor}{rgb}{0.043137254901960784,0.7529411764705882,0.8}{\sethlcolor{customcolor} \textcolor{black}{\hl{OF\textsuperscript{9}}}} \definecolor{customcolor}{rgb}{0.6078431372549019,0.996078431372549,0.796078431372549}{\sethlcolor{customcolor} \textcolor{black}{\hl{FLOWER\textsuperscript{10}}}}\\
\definecolor{customcolor}{rgb}{1.0,0.8823529411764706,0.6196078431372549}{\sethlcolor{customcolor} \textcolor{black}
{\hl{MOON\textsuperscript{11}}}}\\
\definecolor{customcolor}{rgb}{1.0,0.4823529411764706,0.4588235294117647}{\sethlcolor{customcolor} \textcolor{black}{\hl{SCREENLY ERIC ROTH AND MARTIIN SCORSESE DIRECTED BYMARTIN SORSESE\textsuperscript{12}}}}\\
\definecolor{customcolor}{rgb}{1.0,0.25882352941176473,0.5411764705882353}{\sethlcolor{customcolor} \textcolor{black}
{\hl{ONLY IN THEATRES OCTOBER 20\textsuperscript{13}}}}
\end{alltt}
}
\hspace{10px}
\end{AIbox}
\captionof{figure}{OCR with region prediction results.}
\end{minipage}

\subsection{Region to segmentation}

\begin{minipage}{1.0\linewidth}
    
\centering
\begin{AIbox}{Region to Segmentation}

\parbox[t]{0.50\textwidth}{
        {\bf Prompt}: 
        What is the polygon mask of region
        $\langle$loc\_586$\rangle$$\langle$loc\_294$\rangle$$\langle$loc\_929$\rangle$$\langle$loc\_814$\rangle$
    \centering
        \includegraphics[width=0.50\textwidth]{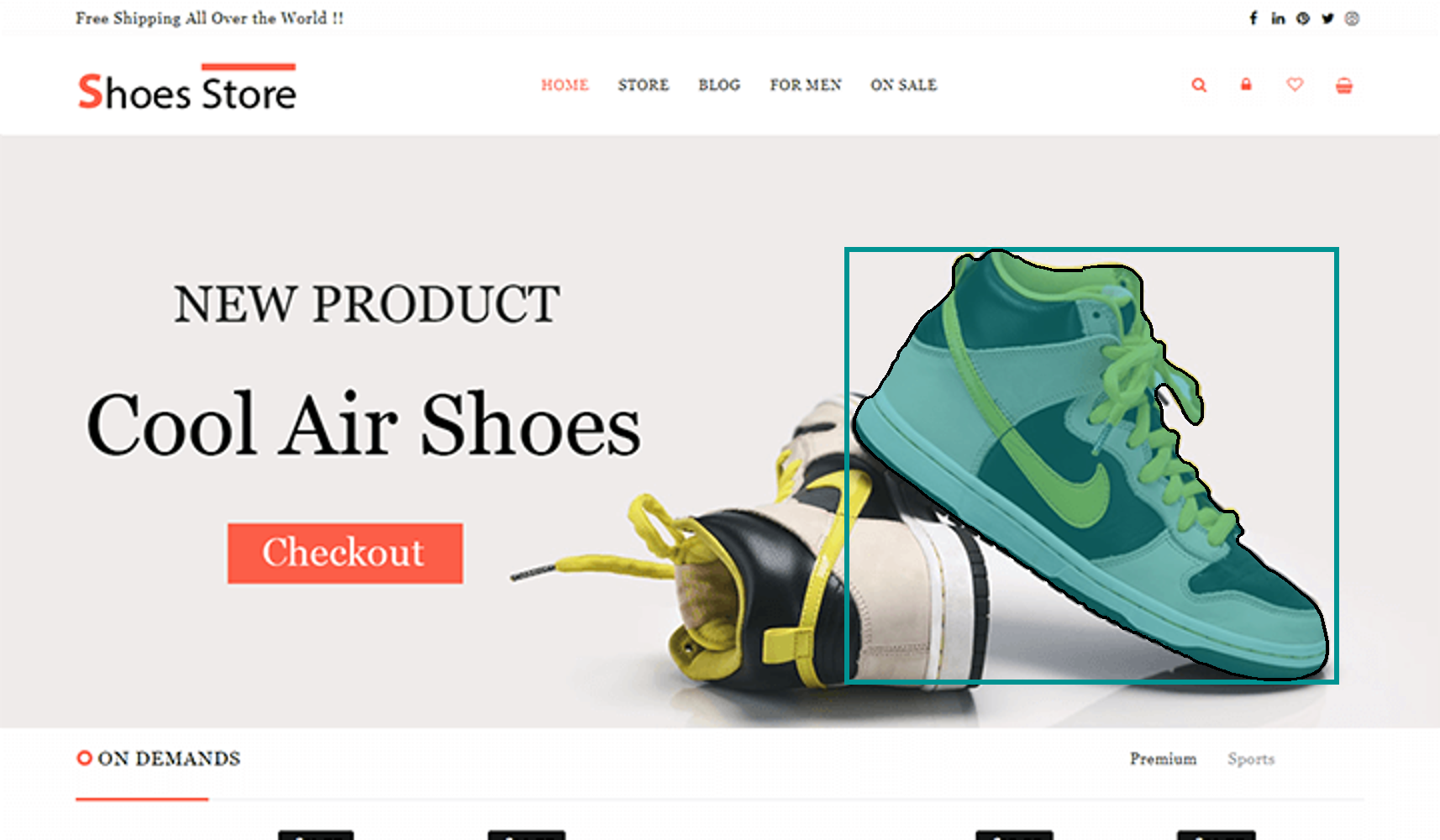}
}
\parbox[t]{0.50\textwidth}{
        {\bf Prompt}: 
        What is the polygon mask of region $\langle$loc\_317$\rangle$$\langle$loc\_314$\rangle$$\langle$loc\_893$\rangle$$\langle$loc\_904$\rangle$
    \centering
        \includegraphics[width=0.45\textwidth]{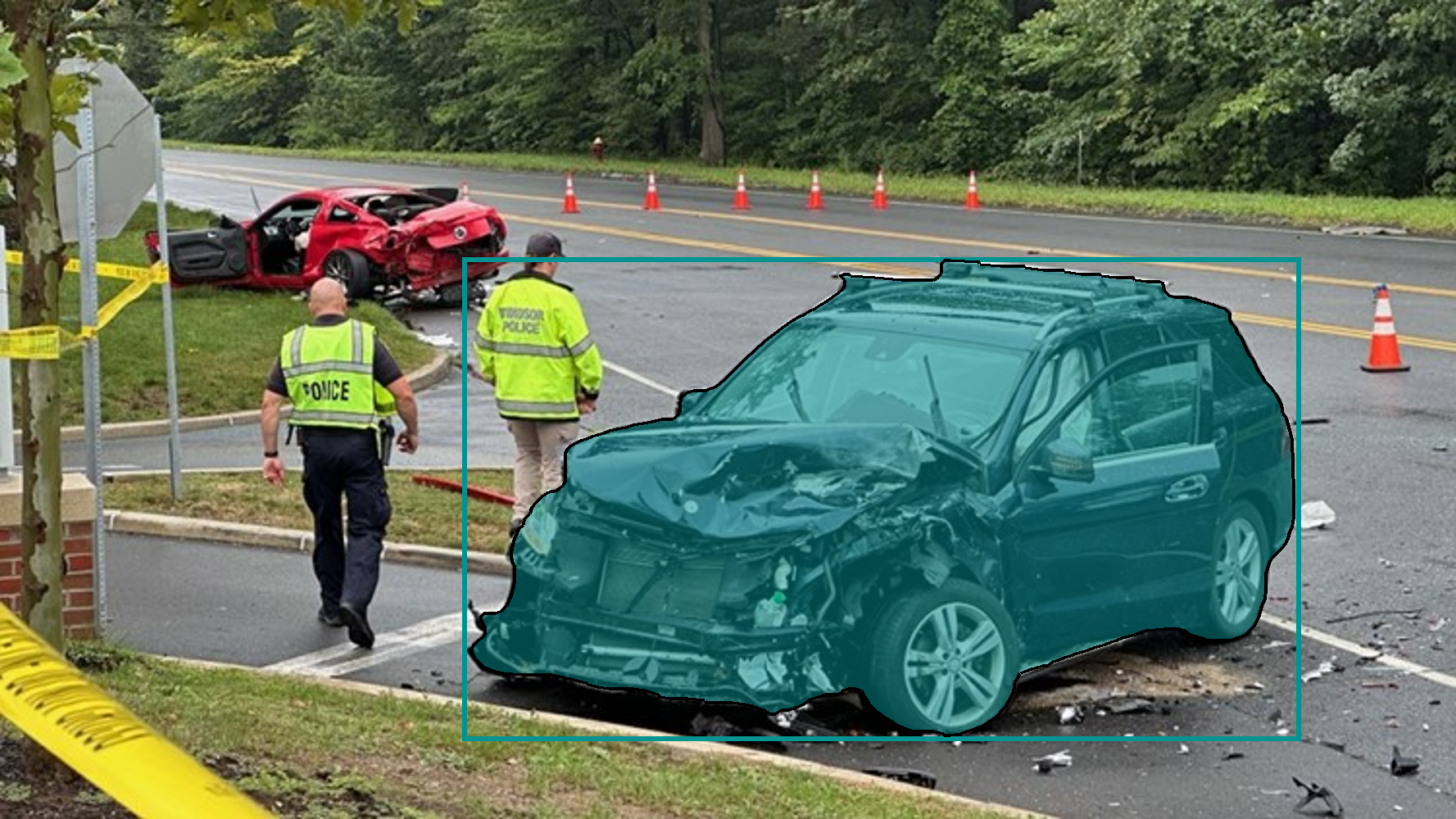}
}

\tcbline

\parbox[t]{0.50\textwidth}{
        {\bf Prompt}: 
        What is the polygon mask of region
        $\langle$loc\_541$\rangle$$\langle$loc\_266$\rangle$$\langle$loc\_692$\rangle$$\langle$loc\_627$\rangle$
    \centering
        \includegraphics[width=0.50\textwidth]{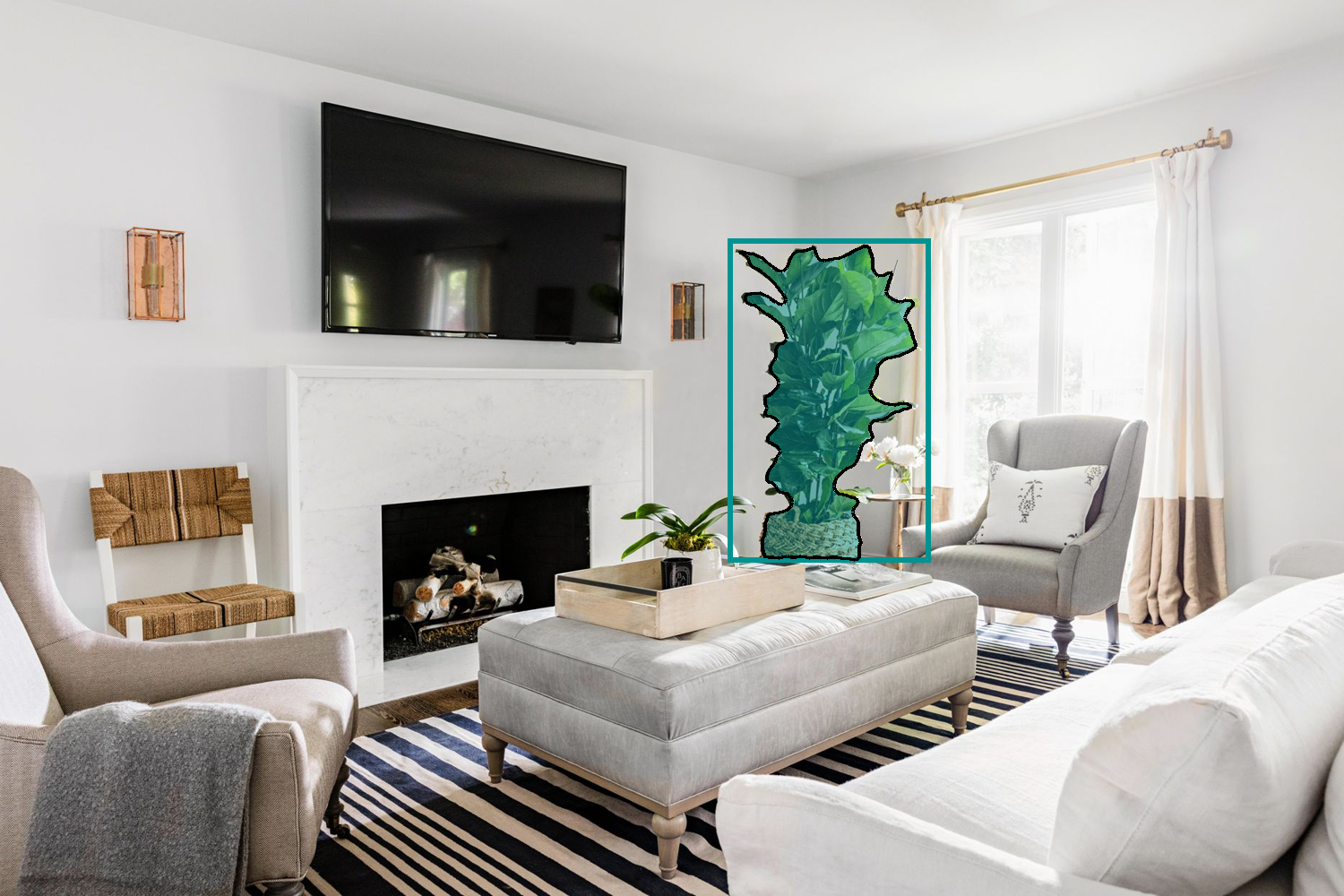}
}
\parbox[t]{0.50\textwidth}{
        {\bf Prompt}: 
        What is the polygon mask of region
        $\langle$loc\_583$\rangle$$\langle$loc\_66$\rangle$$\langle$loc\_794$\rangle$$\langle$loc\_331$\rangle$
    \centering
        \includegraphics[width=0.30\textwidth]{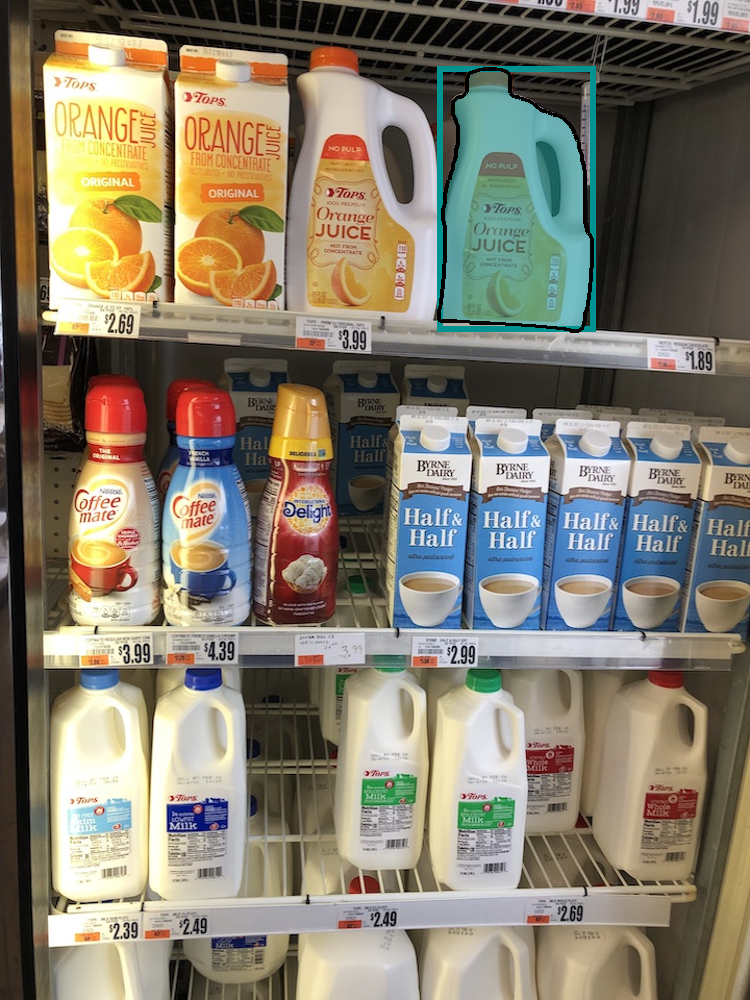}
}

\tcbline

\parbox[t]{0.50\textwidth}{
        {\bf Prompt}: 
        What is the polygon mask of region        $\langle$loc\_386$\rangle$$\langle$loc\_53$\rangle$$\langle$loc\_759$\rangle$$\langle$loc\_998$\rangle$
    \centering
        \includegraphics[width=0.45\textwidth]{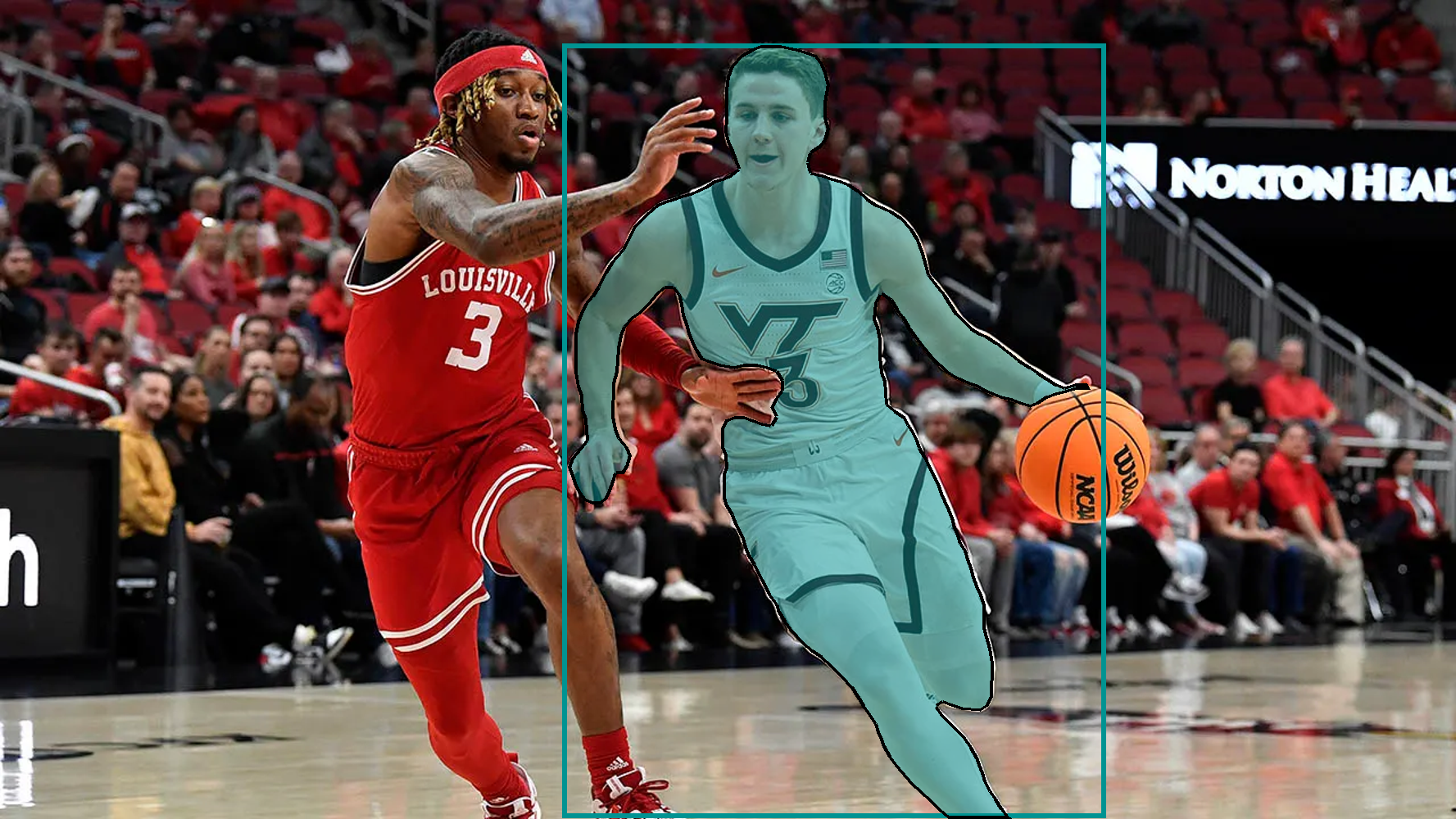}
}
\parbox[t]{0.50\textwidth}{
        {\bf Prompt}: 
        What is the polygon mask of region $\langle$loc\_102$\rangle$$\langle$loc\_7$\rangle$$\langle$loc\_375$\rangle$$\langle$loc\_648$\rangle$
    \centering
        \includegraphics[width=0.45\textwidth]{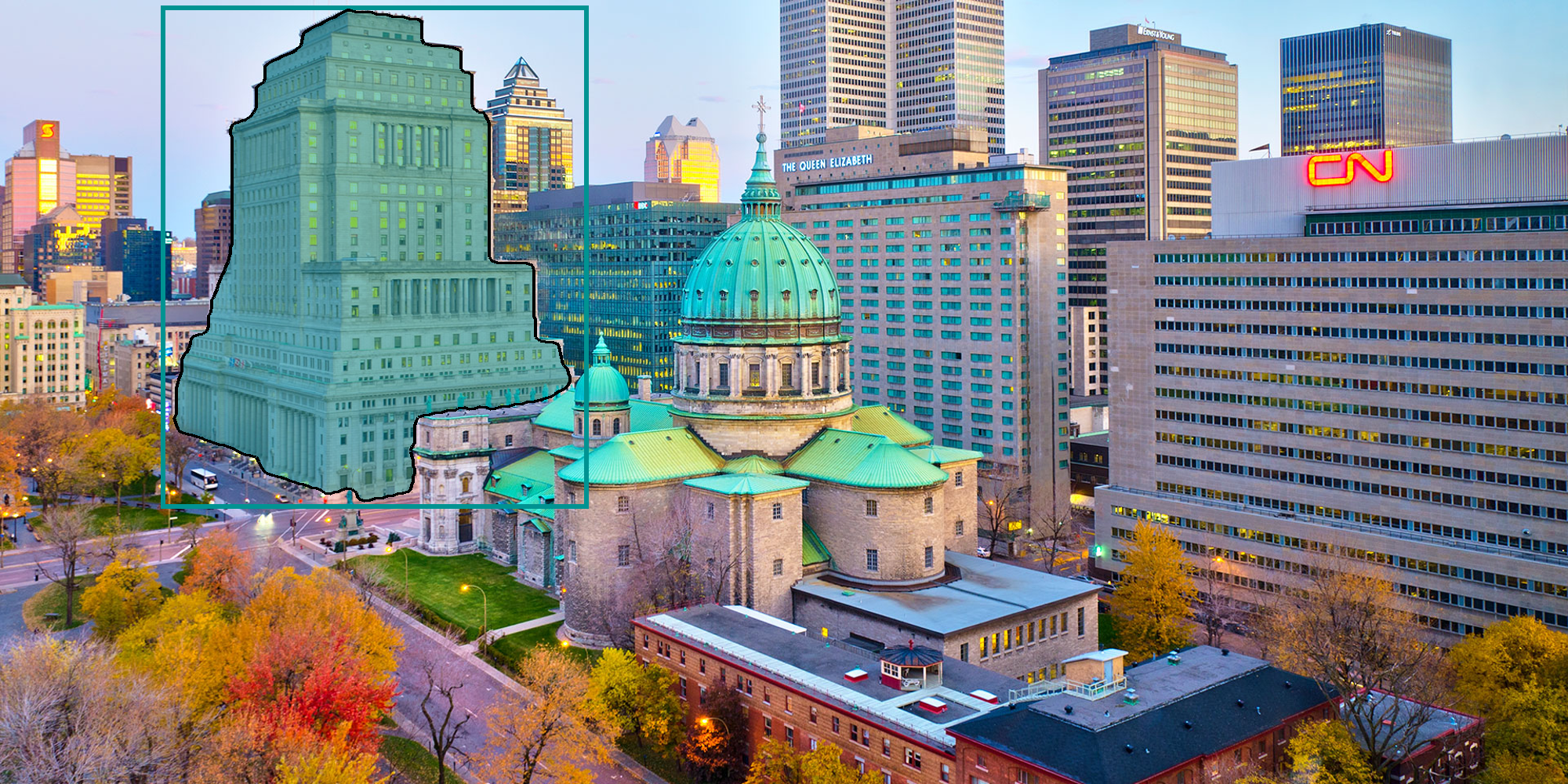}
}

\end{AIbox}
\captionof{figure}{Region to segmentation prediction results.}
\end{minipage}

\section{Comparision with LMMs on Detailed Caption Task}

\begin{minipage}{1.0\linewidth}
\centering
\begin{AIbox}{Comparison with LMMs on Detailed Image Caption Task}

\parbox[h]{1.0\textwidth}{
    \centering
    \includegraphics[angle=-90,width=0.48\textwidth]{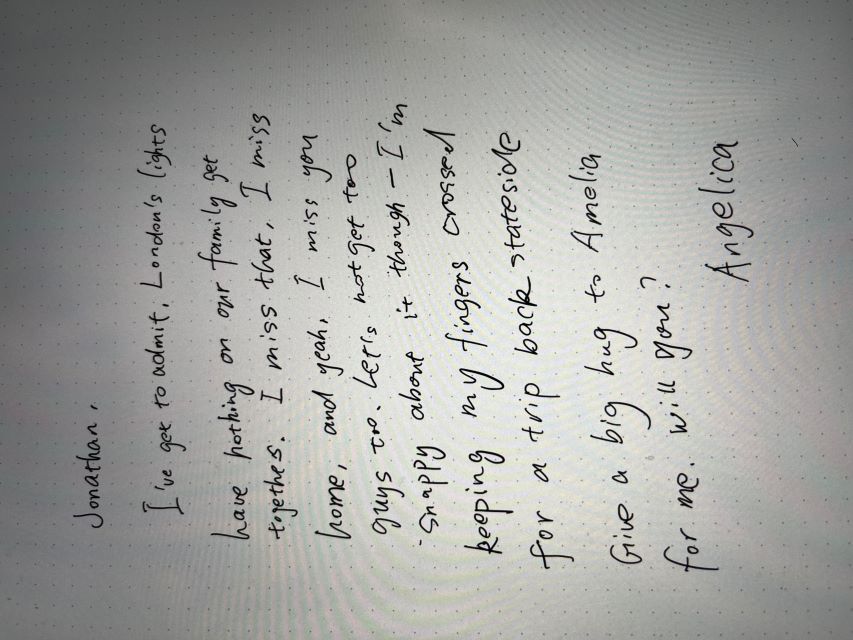}
}
\tcbline
\parbox[h]{1.0\textwidth}{
\scriptsize 
\begin{alltt}
\textbf{\flmodel}: The image is a photograph of a handwritten note from a person named Angelica. The note is written in black ink on a white background. The handwriting is neat and legible, and it appears to be a handwritten letter. The person's name, Jonathan, is written at the top of the note in a cursive font. Below the name, there is a note that reads, "I've got to admit, London's lights have nothing on our family get together. I miss that, I miss you guys too. Let's not get too happy about it though I'm keeping my fingers crossed for a trip back stateside. Give a big hug to Amelia for me. Will you?"
\end{alltt}

\begin{alltt}
\textbf{GPT-4V\footnote{\url{https://chat.openai.com}}}: The image shows a handwritten note addressed to Jonathan. The note conveys a personal message from the sender, Angelica, who admits to missing family gatherings and the feeling of home, contrasting it with the lights of London. Angelica expresses a longing to return stateside and asks Jonathan to give Amelia a big hug on her behalf. The tone of the note is casual and warm, with a sense of fondness and a hint of homesickness. The note is written informally, suggesting a close relationship between the sender and the recipients.
\end{alltt}

\begin{alltt}
\textbf{LLaVA\footnote{\url{https://llava.hliu.cc}}:} The image is a handwritten letter on a piece of paper, written by a person named Jonathan. The letter is addressed to someone named Angela, and it appears to be a heartfelt message expressing love and affection. The letter is written in cursive, giving it a personal and intimate touch.
\end{alltt}

\begin{alltt}
\textbf{MiniGPT-4\footnote{\url{https://huggingface.co/spaces/Vision-CAIR/minigpt4}}:} The image is a handwritten letter from an unknown person to someone named Angel. The letter is written in black ink on a piece of white paper. The letter is addressed to Angel and begins with the words, “Dear Angel,”. The letter goes on to describe the writer’s feelings about a recent event or situation, but the details are unclear. The letter ends with the words, “Sincerely, [Writer’s Name].”

\end{alltt}

}
\end{AIbox}
\captionof{figure}{Comparison with LMMs on detailed caption task.}
\end{minipage}

\begin{minipage}{1.0\linewidth}
\centering
\begin{AIbox}{Comparison with LMMs on Detailed Image Caption Task}

\parbox[h]{1.0\textwidth}{
    \centering
    \includegraphics[width=0.7\textwidth]{figures/appendix/comparison/com_5.jpg}
}
\tcbline
\parbox[h]{1.0\textwidth}{
\scriptsize 
\begin{alltt}
\textbf{\flmodel}: The image is a close-up of a red apple with a green label on it. The label has the text "FUJI 94131 USA" and a barcode on it, indicating that the apple is from the United States. The apple appears to be fresh and has a shiny texture. The background is white, making the apple stand out.
\end{alltt}

\vspace{2em}

\begin{alltt}
\textbf{GPT-4V\footnote{\url{https://chat.openai.com}}}: The image displays a Fuji apple with a sticker indicating it is organic, along with the identifying number 94131, and specifying that it is from the USA. The apple has a reddish-pink skin with some yellow and green coloration, which is characteristic of Fuji apples. The apple is placed on a white background, which appears to be a textured fabric.
\end{alltt}
\vspace{2em}

\begin{alltt}
\textbf{LLaVA\footnote{\url{https://llava.hliu.cc}}:} The image features a large, fresh, and ripe red apple sitting on a white surface. The apple has a sticker on it, which is likely a barcode or a label providing information about the product. The apple is prominently displayed, taking up a significant portion of the image.
\end{alltt}
\vspace{2em}

\begin{alltt}
\textbf{MiniGPT-4\footnote{\url{https://huggingface.co/spaces/Vision-CAIR/minigpt4}}:} The image shows a red apple with a label on it that reads “Organic” in white letters. The apple is sitting on a white surface.
\end{alltt}
\vspace{2em}

}
\end{AIbox}
\captionof{figure}{Comparison with LMMs on detailed caption task (continued).}
\end{minipage}

\begin{minipage}{1.0\linewidth}
\centering
\begin{AIbox}{Comparison with LMMs on Detailed Image Caption Task}

\parbox[h]{1.0\textwidth}{
    \centering
    \includegraphics[width=0.8\textwidth]{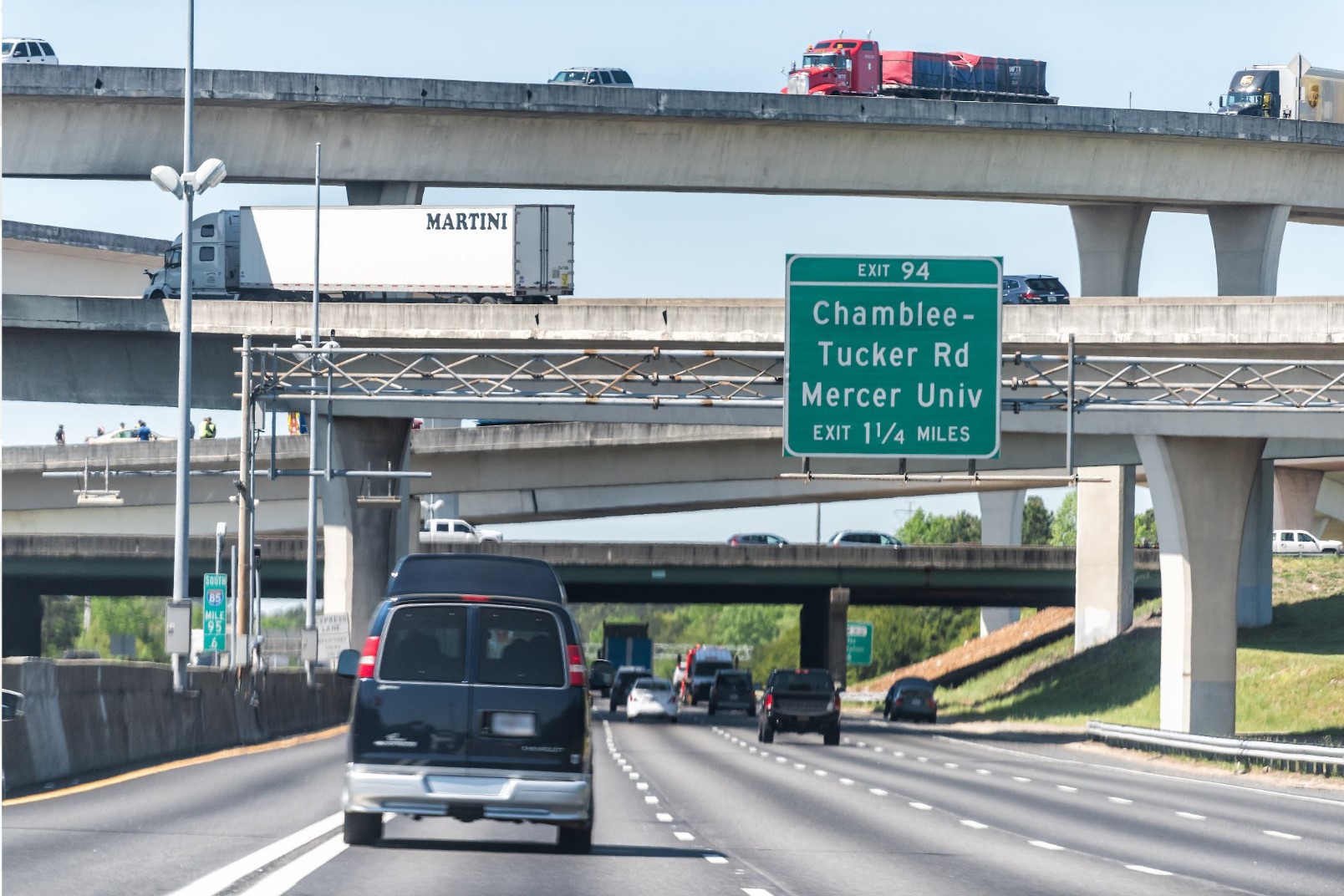}
}
\tcbline
\parbox[h]{1.0\textwidth}{
\scriptsize 
\begin{alltt}
\textbf{\flmodel}: The image shows a highway with multiple lanes. There are several vehicles on the road, including a black van and a red truck. On the left side of the image, there is a white semi-truck with the word "Martini" written on it. Above the highway, there are several bridges with cars and trucks on them. The sky is blue and there are trees and grass on the right side. In the center of the highway is a green sign that reads "Exit 94, Chamblee Tucker Rd, Mercer Univ."
\end{alltt}

\vspace{2em}

\begin{alltt}
\textbf{GPT-4V\footnote{\url{https://chat.openai.com}}}: The image shows a highway scene with several lanes of traffic. There are overpasses with cars and trucks traveling over them, and the highway itself appears to be busy with vehicles, including a prominent black van in the foreground.

A road sign indicates that Exit 94 for Chamblee-Tucker Rd and Mercer University is 1 1/4 miles ahead. The presence of the sign suggests that drivers in the right lanes will soon have the opportunity to take the exit to reach these destinations.

The infrastructure suggests a well-traveled interstate or highway system, possibly in an urban or suburban area given the overpass complexity and road signage for local roads and institutions. The vehicles on the road are diverse, indicating a mix of commercial and personal transport.
\end{alltt}
\vspace{2em}

\begin{alltt}
\textbf{LLaVA\footnote{\url{https://llava.hliu.cc}}:} The image depicts a busy highway with multiple lanes of traffic, including cars and trucks. There is a large truck driving on the highway, and several other trucks are also visible in the scene. The highway is filled with cars, and there is a traffic sign above the lanes, indicating the direction to Tucker Road. The scene captures the hustle and bustle of a typical day on a highway filled with vehicles.
\end{alltt}
\vspace{2em}

\begin{alltt}
\textbf{MiniGPT-4\footnote{\url{https://huggingface.co/spaces/Vision-CAIR/minigpt4}}:} The image shows a highway with multiple lanes of traffic moving in both directions. There are several large bridges spanning the highway, and in the distance, there are mountains and a blue sky. There are also several large trucks and cars on the highway, including a black van in the center of the image. The road surface is smooth and well-maintained.
\end{alltt}

}
\end{AIbox}
\captionof{figure}{Comparison with LMMs on detailed caption task (continued).}
\end{minipage}

\begin{minipage}{1.0\linewidth}
\centering
\begin{AIbox}{Comparison with LMMs on Detailed Image Caption Task}

\parbox[h]{1.0\textwidth}{
    \centering
    \includegraphics[width=0.44\textwidth]{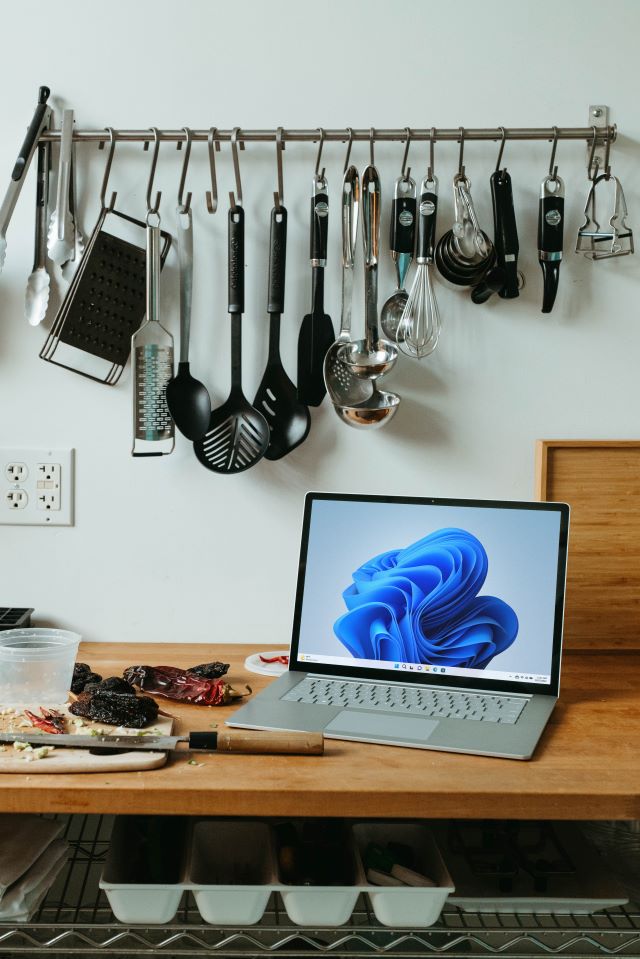}
}
\tcbline
\vspace{-1em}
\parbox[h]{1.0\textwidth}{
\scriptsize 
\begin{alltt}
\textbf{\flmodel}: The image shows a kitchen countertop with a laptop on it. The laptop is open and the screen is displaying a blue abstract design. On the countertop, there are various kitchen utensils hanging on a metal rack. The rack is attached to the wall above the sink and has a variety of kitchen tools hanging on it, including a grater, a spatula, a ladle, a spoon, and a whisk. There are also a few other kitchen items scattered around the counter. The countertop is made of wood and there is a wire rack at the bottom of the image.
\end{alltt}

\begin{alltt}
\textbf{GPT-4V\footnote{\url{https://chat.openai.com}}}: The image shows a modern kitchen setting where culinary tools are organized neatly on a rack against the wall. These tools include a variety of utensils such as a grater, tongs, a whisk, ladles, and spatulas. On the counter, there is an open laptop with a vibrant blue wallpaper on the screen, which suggests that someone might be using it for a recipe or as entertainment while cooking.
In front of the laptop, there appears to be a cutting board with some chopped ingredients and a knife, indicating that food preparation is underway or has recently taken place. There are also some dried chilies and a plastic container that might contain other ingredients or spices, further suggesting cooking activities.
The counter also has an electrical outlet and some covered sockets, and below the main counter, there are storage bins that likely hold additional kitchen supplies. The environment looks lived-in and practical, with a mix of technology and traditional cooking tools.  
\end{alltt}

\begin{alltt}
\textbf{LLaVA\footnote{\url{https://llava.hliu.cc}}:} The image features a wooden kitchen counter with a laptop computer sitting on top of it. The laptop is open and displaying a picture of a cake. The counter is also cluttered with various kitchen utensils, including multiple spoons and knives, as well as a bowl. The scene suggests that someone might be using the laptop for cooking or baking purposes, possibly following a recipe or watching a tutorial.
\end{alltt}

\begin{alltt}
\textbf{MiniGPT-4\footnote{\url{https://huggingface.co/spaces/Vision-CAIR/minigpt4}}:} The image shows a kitchen counter with a laptop computer on it, a set of cooking utensils hanging from a rack above it, and a knife block on the counter. The laptop appears to be open, with a blue screen displaying an image of a kitchen. The utensils hanging from the rack include a spatula, a ladle, a whisk, and a cutting board. The knife block contains several knives, including a chef’s knife, a paring knife, and a bread knife.
\end{alltt}

}
\end{AIbox}
\captionof{figure}{Comparison with LMMs on detailed caption task (continued).}
\end{minipage}

\section{Comarison with Kosmos-2 on Detailed Caption and Grounding Tasks}

\begin{minipage}{1.0\linewidth}

\centering
\begin{AIbox}{Comparison on detailed caption and grounding tasks.}

\parbox[h]{0.5\textwidth}{
    \scriptsize \begin{alltt}
    \centering
        \includegraphics[width=0.45\textwidth]{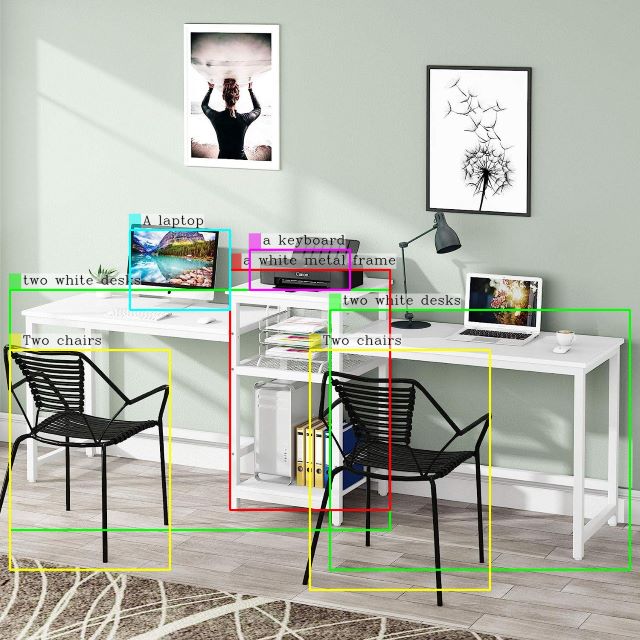}
    \end{alltt}
}
\hspace{10px}
\parbox[h]{0.50\textwidth}{ 
    \scriptsize \begin{alltt} The image features a home office with \definecolor{customcolor}{HTML}{69ff85}{\sethlcolor{customcolor}\textcolor{black}{\hl{two white desks}}}, one on the left and the other on the right. The desks are connected by  \definecolor{customcolor}{HTML}{d0544f}{\sethlcolor{customcolor}\textcolor{black}{\hl{a white metal frame}}}, creating a functional and stylish workspace. \definecolor{customcolor}{HTML}{73f7f3}{\sethlcolor{customcolor}\textcolor{black}{\hl{A laptop}}}  is placed on the desk on the far left, and   \definecolor{customcolor}{HTML}{e86ce7}{\sethlcolor{customcolor}\textcolor{black}{\hl{a keyboard}}}  is visible on the other desk.   \definecolor{customcolor}{HTML}{dee35f}{\sethlcolor{customcolor}\textcolor{black}{\hl{Two chairs}}}  are placed in front of the desks, one closer to the left desk and the second one on its right side.
\end{alltt}
}
\vspace{1em}
\begin{center}
(a) Result from Kosmos-2.
\end{center}
\vspace{1em}

\parbox[h]{0.5\textwidth}{
    \scriptsize \begin{alltt}
    \centering
        \includegraphics[width=0.45\textwidth]{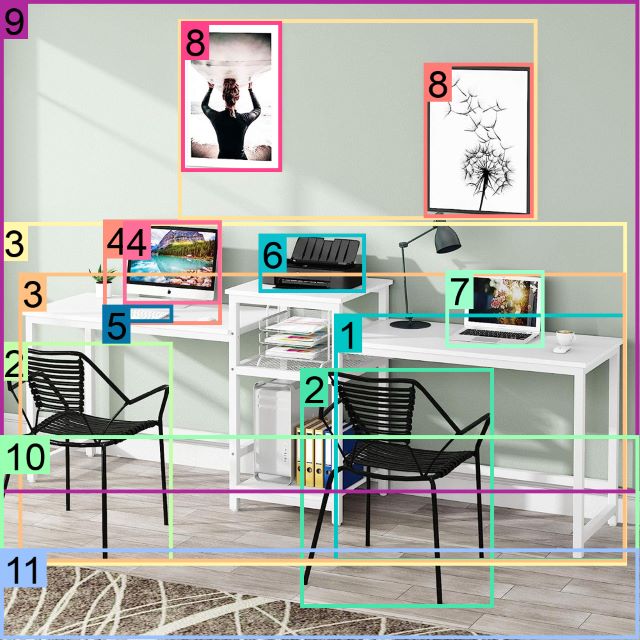}
    \end{alltt}
}
\hspace{10px}
\parbox[h]{0.45\textwidth}{ 
    \scriptsize \begin{alltt} 
     \noindent The image shows a modern home office setup with two white \definecolor{customcolor}{rgb}{0.0,0.7764705882352941,0.7725490196078432}{\sethlcolor{customcolor}\textcolor{black}{\hl{desks\textsuperscript{1}}}} and \definecolor{customcolor}{rgb}{0.28627450980392155,0.8941176470588236,0.6705882352941176}{\sethlcolor{customcolor}\textcolor{black}{\hl{two black chairs\textsuperscript{2}}}}. \definecolor{customcolor}{rgb}{1.0,0.9686274509803922,0.6705882352941176}{\sethlcolor{customcolor}\textcolor{black}{\hl{The desks\textsuperscript{3}}}} are arranged in a corner of the room with a large window on the left side. On the right side of the desk, there is \definecolor{customcolor}{rgb}{1.0,0.5686274509803921,0.48627450980392156}{\sethlcolor{customcolor}\textcolor{black}{\hl{a computer monitor\textsuperscript{4}}}}, a \definecolor{customcolor}{rgb}{0.023529411764705882,0.4627450980392157,0.6588235294117647}{\sethlcolor{customcolor}\textcolor{black}{\hl{keyboard\textsuperscript{5}}}}, a mouse, \definecolor{customcolor}{rgb}{0.043137254901960784,0.7529411764705882,0.8}{\sethlcolor{customcolor}\textcolor{black}{\hl{a printer\textsuperscript{6}}}}, and a \definecolor{customcolor}{rgb}{0.6078431372549019,0.996078431372549,0.796078431372549}{\sethlcolor{customcolor}\textcolor{black}{\hl{laptop\textsuperscript{7}}}}. Above the computer monitor and keyboard, there are \definecolor{customcolor}{rgb}{1.0,0.8823529411764706,0.6196078431372549}{\sethlcolor{customcolor}\textcolor{black}{\hl{two framed pictures\textsuperscript{8}}}} hanging on the wall. \definecolor{customcolor}{rgb}{0.6862745098039216,0.1607843137254902,0.615686274509804}{\sethlcolor{customcolor}\textcolor{black}{\hl{The walls\textsuperscript{9}}}} are painted in a light green color and \definecolor{customcolor}{rgb}{0.6,1.0,0.6941176470588235}{\sethlcolor{customcolor}\textcolor{black}{\hl{the floor\textsuperscript{10}}}} is made of light{-}colored wood. \definecolor{customcolor}{rgb}{0.6,0.7607843137254902,1.0}{\sethlcolor{customcolor}\textcolor{black}{\hl{The floor\textsuperscript{11}}}} is covered with a beige area rug with a geometric pattern. The overall style of the space is minimal and contemporary.
\end{alltt}
}
\vspace{1em}
\begin{center}
(b) Result from \flmodel.
\end{center}
\vspace{1em}

\end{AIbox}
\captionof{figure}{Systematic comparison with Kosmos-2~\cite{peng2023kosmos} on detailed caption and grounding tasks. The models generate both the detailed caption and grounding results. The results of Kosmos-2 are from \url{https://huggingface.co/spaces/ydshieh/Kosmos-2}.}
\end{minipage}

\begin{minipage}{1.0\linewidth}

\centering
\begin{AIbox}{Comparison on detailed caption and grounding tasks.}

\parbox[h]{0.5\textwidth}{
    \scriptsize \begin{alltt}
    \centering
        \includegraphics[width=0.4\textwidth]{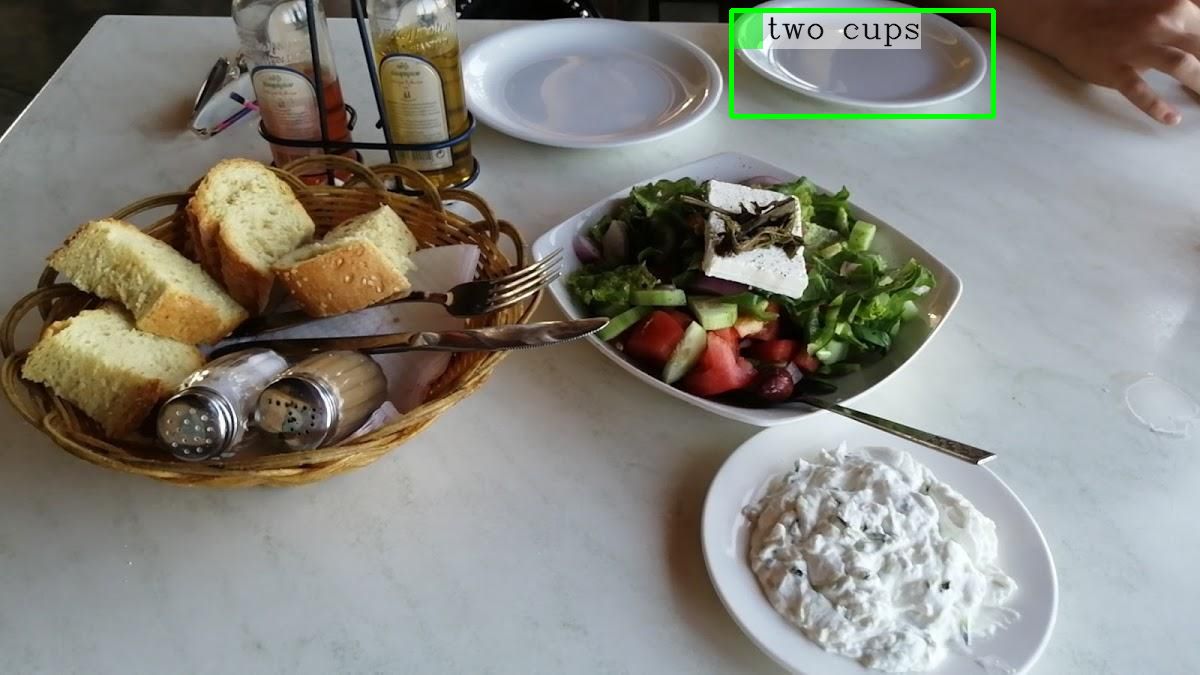}
    \end{alltt}
}
\hspace{10px}
\parbox[h]{0.45\textwidth}{ 
    \scriptsize \begin{alltt} The image features a white dining table with a variety of food items on it. On the table, there is a bowl of bread, a bowl with a salad, and a plate with a side of yogurt. In addition to the food, there are   \definecolor{customcolor}{HTML}{69ff85}{\sethlcolor{customcolor}\textcolor{black}{\hl{two cups}}}  placed on the table. One cup is located near the left side of the table and the other cup is situated on the right side. The table is set with a fork and a knife, ready for a meal.
\end{alltt}
}
\begin{center}
\vspace{-1em}
(c) Result from Kosmos-2.
\vspace{-1em}
\end{center}

\parbox[h]{0.5\textwidth}{
\vspace{-1em}
    \scriptsize \begin{alltt}
    \centering
        \includegraphics[width=0.4\textwidth]{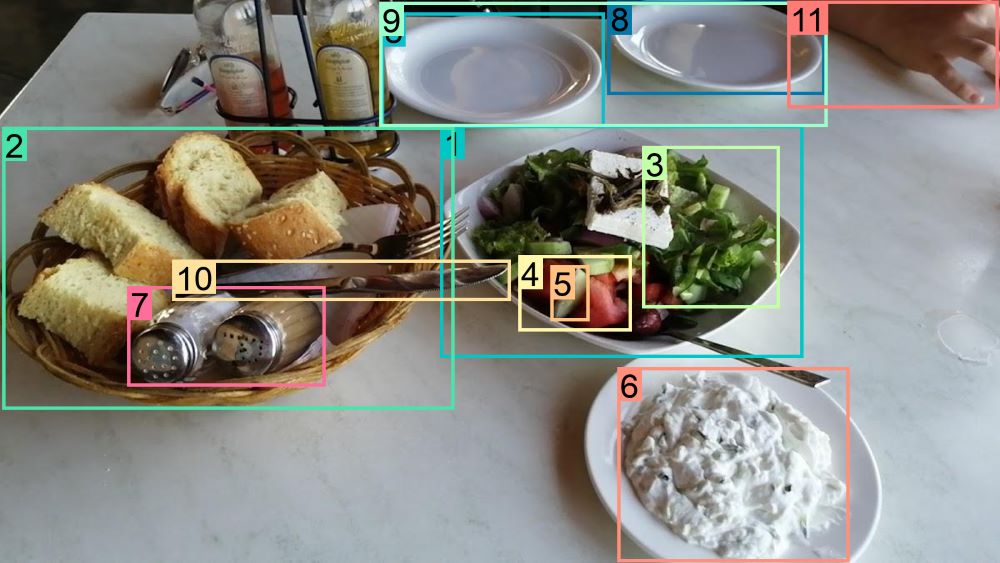}
    \end{alltt}
}
\hspace{10px}
\parbox[h]{0.45\textwidth}{
\begin{alltt}
\scriptsize 
 \noindent The image shows a table with a basket of bread and a plate of \definecolor{customcolor}{rgb}{0.0,0.7764705882352941,0.7725490196078432}{\sethlcolor{customcolor}\textcolor{black}{\hl{salad\textsuperscript{1}}}}. \definecolor{customcolor}{rgb}{0.28627450980392155,0.8941176470588236,0.6705882352941176}{\sethlcolor{customcolor}\textcolor{black}{\hl{The basket\textsuperscript{2}}}} is made of woven straw and has several slices of bread in it. Next to the basket, there is \definecolor{customcolor}{rgb}{0.0,0.7764705882352941,0.7725490196078432}{\sethlcolor{customcolor}\textcolor{black}{\hl{a small bowl\textsuperscript{1}}}} of salad with a variety of vegetables, including \definecolor{customcolor}{rgb}{0.7647058823529411,0.996078431372549,0.6588235294117647}{\sethlcolor{customcolor}\textcolor{black}{\hl{lettuce\textsuperscript{3}}}}, \definecolor{customcolor}{rgb}{1.0,0.9686274509803922,0.6705882352941176}{\sethlcolor{customcolor}\textcolor{black}{\hl{tomatoes\textsuperscript{4}}}}, \definecolor{customcolor}{rgb}{1.0,0.7568627450980392,0.5098039215686274}{\sethlcolor{customcolor}\textcolor{black}{\hl{cucumbers\textsuperscript{5}}}}, and \definecolor{customcolor}{rgb}{1.0,0.5686274509803921,0.48627450980392156}{\sethlcolor{customcolor}\textcolor{black}{\hl{feta cheese\textsuperscript{6}}}}. There are also \definecolor{customcolor}{rgb}{1.0,0.43137254901960786,0.6078431372549019}{\sethlcolor{customcolor}\textcolor{black}{\hl{two salt\textsuperscript{7}}}} and pepper \definecolor{customcolor}{rgb}{1.0,0.43137254901960786,0.6078431372549019}{\sethlcolor{customcolor}\textcolor{black}{\hl{shakers\textsuperscript{7}}}} on the table. On the right side of the table, there are \definecolor{customcolor}{rgb}{0.023529411764705882,0.4627450980392157,0.6588235294117647}{\sethlcolor{customcolor}\textcolor{black}{\hl{two white plates\textsuperscript{8}}}}\definecolor{customcolor}{rgb}{0.6078431372549019,0.996078431372549,0.796078431372549}{\sethlcolor{customcolor}\textcolor{black}{\hl{\textsuperscript{9}}}} with a dollop of white sauce on them. The table is covered with a white tablecloth and there are \definecolor{customcolor}{rgb}{0.6078431372549019,0.996078431372549,0.796078431372549}{\sethlcolor{customcolor}\textcolor{black}{\hl{a few other dishes\textsuperscript{9}}}} and \definecolor{customcolor}{rgb}{1.0,0.8823529411764706,0.6196078431372549}{\sethlcolor{customcolor}\textcolor{black}{\hl{utensils\textsuperscript{10}}}} scattered around. \definecolor{customcolor}{rgb}{1.0,0.4823529411764706,0.4588235294117647}{\sethlcolor{customcolor}\textcolor{black}{\hl{A person"s hand\textsuperscript{11}}}} can be seen in the top right corner of the image.
\end{alltt}
}
\begin{center}
\vspace{-1em}
(d) Result from \flmodel.
\vspace{-1em}
\end{center}

\tcbline

\parbox[h]{0.5\textwidth}{
    \scriptsize \begin{alltt}
    \centering
        \includegraphics[width=0.4\textwidth]{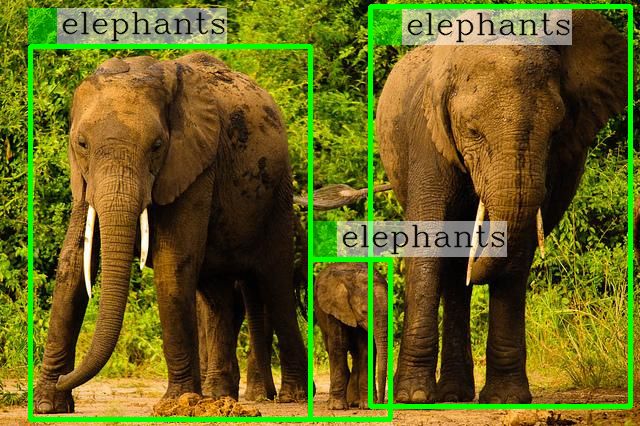}
    \end{alltt}
}
\hspace{10px}
\parbox[h]{0.50\textwidth}{ 
    \scriptsize \begin{alltt}  The image features a family of   \definecolor{customcolor}{HTML}{69ff85}{\sethlcolor{customcolor}\textcolor{black}{\hl{elephants}}}  walking together in a lush green forest. The  \definecolor{customcolor}{HTML}{69ff85}{\sethlcolor{customcolor}\textcolor{black}{\hl{elephants}}} are walking in a line, with one of them walking in front of the rest of the family. The family consists of a mother and her two calves, with the mother and the two calves walking side by side. The forest is filled with green trees, adding to the natural beauty of the scene.
\end{alltt}
}
\begin{center}
\vspace{-1em}
(e) Result from Kosmos-2.
\vspace{-1em}
\end{center}

\parbox[h]{0.5\textwidth}{
    \scriptsize \begin{alltt}
    \centering
        \includegraphics[width=0.4\textwidth]{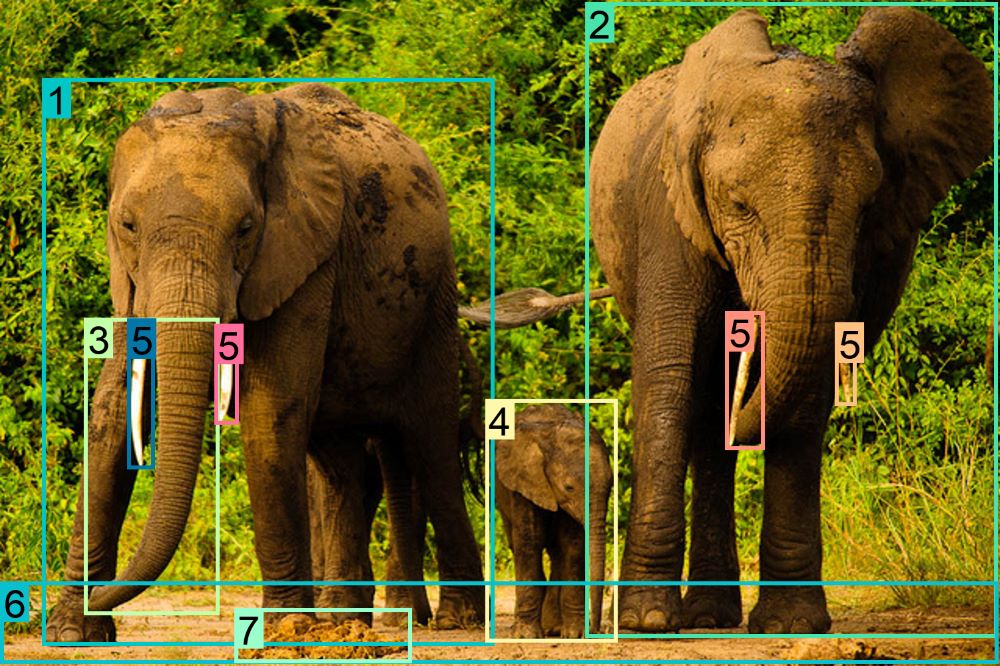}
    \end{alltt}
}
\hspace{10px}
\parbox[h]{0.45\textwidth}{ 
\begin{alltt}
\scriptsize
 \noindent The image shows a group of three elephants standing in a dirt field with trees and bushes in the background. \definecolor{customcolor}{rgb}{0.0,0.7764705882352941,0.7725490196078432}{\sethlcolor{customcolor}\textcolor{black}{\hl{The elephants\textsuperscript{1}}}} are standing close together, with the largest elephant in the center and two smaller ones on either side. \definecolor{customcolor}{rgb}{0.0,0.7764705882352941,0.7725490196078432}{\sethlcolor{customcolor}\textcolor{black}{\hl{The largest elephant\textsuperscript{1}}}}\definecolor{customcolor}{rgb}{0.28627450980392155,0.8941176470588236,0.6705882352941176}{\sethlcolor{customcolor}\textcolor{black}{\hl{\textsuperscript{2}}}}  on the left is standing with its \definecolor{customcolor}{rgb}{0.7647058823529411,0.996078431372549,0.6588235294117647}{\sethlcolor{customcolor}\textcolor{black}{\hl{trunk\textsuperscript{3}}}} extended, while \definecolor{customcolor}{rgb}{1.0,0.9686274509803922,0.6705882352941176}{\sethlcolor{customcolor}\textcolor{black}{\hl{the smaller one\textsuperscript{4}}}} is standing next to it. \definecolor{customcolor}{rgb}{0.0,0.7764705882352941,0.7725490196078432}{\sethlcolor{customcolor}\textcolor{black}{\hl{All three elephants\textsuperscript{1}}}} have \definecolor{customcolor}{rgb}{1.0,0.7568627450980392,0.5098039215686274}{\sethlcolor{customcolor}\textcolor{black}{\hl{tusks\textsuperscript{5}}}} and appear to be in their natural habitat. \definecolor{customcolor}{rgb}{0.043137254901960784,0.7529411764705882,0.8}{\sethlcolor{customcolor}\textcolor{black}{\hl{The ground\textsuperscript{6}}}} is covered in dirt and there is \definecolor{customcolor}{rgb}{0.6078431372549019,0.996078431372549,0.796078431372549}{\sethlcolor{customcolor}\textcolor{black}{\hl{a small pile of dirt\textsuperscript{7}}}} in front of them. The overall mood of the image is peaceful and serene.
\end{alltt}
}

\begin{center}
\vspace{-1em}
(f) Result from \flmodel.
\end{center}

\end{AIbox}
\captionof{figure}{Systematic comparison with Kosmos-2~\cite{peng2023kosmos} on detailed caption and grounding tasks. The models generate both the detailed caption and grounding results. The results of Kosmos-2 are from \url{https://huggingface.co/spaces/ydshieh/Kosmos-2}. (continued)}
\end{minipage}

\end{document}